\documentclass{article} 
\usepackage{iclr2025_conference,times}

\usepackage{amsmath,amsfonts,bm}

\def\eqref#1{equation~\ref{#1}}

\def\1{\bm{1}}

\DeclareMathAlphabet{\mathsfit}{\encodingdefault}{\sfdefault}{m}{sl}
\SetMathAlphabet{\mathsfit}{bold}{\encodingdefault}{\sfdefault}{bx}{n}

\usepackage{amsmath,amssymb}
\usepackage{booktabs}
\usepackage{graphicx}
\usepackage{subcaption}
\usepackage{hyperref}
\usepackage{url}
\usepackage{subcaption}
\usepackage{booktabs}
\usepackage{multirow}
\usepackage{graphicx}
\usepackage{xcolor}
\usepackage{algorithm}
\usepackage{algpseudocode}
\definecolor{KleinBlue}{RGB}{45, 75, 155}
\definecolor{lightKleinBlue}{RGB}{92, 126, 210}

\definecolor{lightMagenta}{RGB}{210, 140, 180}
\definecolor{deepMagenta}{RGB}{170, 35, 125}
\definecolor{lightPurple}{RGB}{176, 140, 220}

\definecolor{deepPurple}{RGB}{92, 45, 145}

\definecolor{lightGreen}{RGB}{46,160,67}

\definecolor{deepMagenta}{HTML}{A0006D}
\definecolor{DeepSeaGreen}{HTML}{00796B}

\usepackage{tikz}
\usepackage{xcolor}
\usepackage[table]{xcolor}
\usepackage{pgf}

\newcommand{\heatcell}[2]{%
  \cellcolor{black!#1}#2%
}

\usepackage{booktabs}
\usepackage{wrapfig}
\usepackage{amsmath,amssymb}
\usepackage{xcolor}
\usepackage{tikz}

\definecolor{RbwViolet}{HTML}{7650A0}
\definecolor{RbwBlue}{HTML}{4780B8}
\definecolor{RbwCyan}{HTML}{5AA7B3}
\definecolor{RbwGreen}{HTML}{5AA07B}
\definecolor{RbwYellow}{HTML}{C0A956}
\definecolor{RbwOrange}{HTML}{C57F50}
\definecolor{RbwRed}{HTML}{B55872}
\newsavebox{\rainbowbox}

\newcommand{\rainbowmath}[2][84]{%
  \sbox{\rainbowbox}{$\displaystyle #2$}%
  \begin{tikzpicture}[baseline=(base.base)]

    \node[
      anchor=base west,
      inner sep=0pt,
      outer sep=0pt,
      text opacity=0
    ] (base) at (0,0) {$\displaystyle #2$};

    \pgfmathtruncatemacro{\laststripe}{#1-1}%

    \foreach \i in {0,...,\laststripe}{%

      \pgfmathsetmacro{\u}{6*\i/\laststripe}%
      \pgfmathtruncatemacro{\seg}{min(5,floor(\u))}%
      \pgfmathtruncatemacro{\mix}
        {round(100*(\u-\seg))}%

      \ifcase\seg
        \colorlet{stripecolor}
          {RbwBlue!\mix!RbwViolet}%
      \or
        \colorlet{stripecolor}
          {RbwCyan!\mix!RbwBlue}%
      \or
        \colorlet{stripecolor}
          {RbwGreen!\mix!RbwCyan}%
      \or
        \colorlet{stripecolor}
          {RbwYellow!\mix!RbwGreen}%
      \or
        \colorlet{stripecolor}
          {RbwOrange!\mix!RbwYellow}%
      \or
        \colorlet{stripecolor}
          {RbwRed!\mix!RbwOrange}%
      \fi

      \pgfmathsetlengthmacro{\xleft}
        {\i/#1*\wd\rainbowbox}%
      \pgfmathsetlengthmacro{\xright}
        {(\i+1)/#1*\wd\rainbowbox}%

      \begin{scope}
        \clip
          (\xleft,-\dp\rainbowbox)
          rectangle
          (\xright,\ht\rainbowbox);

        \node[
          anchor=base west,
          inner sep=0pt,
          outer sep=0pt,
          text=stripecolor
        ] at (0,0) {$\displaystyle #2$};
      \end{scope}%
    }%
  \end{tikzpicture}%
}

\usepackage{booktabs}

\definecolor{lightKleinBlue}{RGB}{92,126,210}

\definecolor{TiffanyBlue}{RGB}{10,186,181}

\definecolor{deepTiffanyBlue}{RGB}{0, 105, 102}
\definecolor{DeepKleinBlue}{RGB}{0,32,128}
\usepackage{xcolor}
\usepackage{tikz}

\usepackage{xcolor}
\usepackage{tikz}
\usetikzlibrary{fadings}

\usepackage{caption}
\usepackage{amsmath}
\usepackage{amssymb}
\usepackage{amsthm}

\theoremstyle{plain}
\newtheorem{theorem}{Theorem}[section]
\newtheorem{lemma}[theorem]{Lemma}
\newtheorem{proposition}[theorem]{Proposition}
\newtheorem{corollary}[theorem]{Corollary}

\theoremstyle{definition}
\newtheorem{definition}[theorem]{Definition}

\theoremstyle{remark}

\usepackage{xcolor}
\definecolor{citegray}{gray}{0.5}

\let\oldcitep\citep
\renewcommand{\citep}[1]{%
  \textcolor{citegray}{\oldcitep{#1}}%
}

\usepackage{xcolor}
\definecolor{citegray}{RGB}{120,120,120}

  \hypersetup{
    colorlinks=true,
    citecolor=citegray,
    linkcolor=citegray,
    urlcolor=citegray,
    filecolor=citegray
  }

\newcommand{\proofnote}[1]{\textcolor{citegray}{#1}}

\definecolor{NAgray}{gray}{0.75}

\newcommand{\yescell}{\textcolor{DeepSeaGreen}{\textbf{Y}}}
\newcommand{\nocell}{\textcolor{deepMagenta}{\textbf{N}}}
\newcommand{\nacell}{\textcolor{NAgray}{N/A}}

\title{$\mathbb{SL}(n)$ Representation Learning:\\
An Intrinsic Mixed-Curvature Space with \\Higher Curvature Capacities and \\Deeper Order-Aware Composition}

\author{
Xingrun Li$^{1}$ \quad
Yusuke Mukuta$^{1}$ \quad
Xin Yang$^{1}$ \quad
Yinyu Ye$^{2,\ddagger}$ \quad
Tatsuya Harada$^{1,\dagger}$ \\[1.5mm]
$^{1}$The University of Tokyo
\qquad
$^{2}$Stanford University \\[0.8mm]
$^{\dagger}$\texttt{harada@mi.t.u-tokyo.ac.jp}
\qquad
$^{\ddagger}$\texttt{yinyu-ye@stanford.edu}
\qquad
}

\begin{document}

\maketitle

\vspace{-10pt}
\begin{abstract}
Mixed-curvature representation learning seeks to capture rich geometric
structures that cannot be adequately modeled by a single curvature regime.
Existing approaches largely rely on product manifolds, which require manually
specifying how different curvature spaces are combined and separate their
curvature contributions across factors. We introduce the $\mathbb{SL}(n)$
space, a representation geometry defined by the simple $\det(A)=1$ constraint
and a left invariant Schatten-$p$ Finsler structure. Despite this minimal
construction, $\mathbb{SL}(n)$ exhibits pointwise negative, zero, and positive
flag curvature around a common flagpole, while its mixed-curvature and
curvature-coupling capacities are asymptotically maximal relative to the
intrinsic geometric upper bound. Beyond geometry, its noncommutative group
structure provides inherent order sensitivity, and its non-nilpotent Lie
algebra admits nonzero nested Lie brackets at arbitrary depth, enabling deep
order-aware composition. Empirically, $\mathbb{SL}(n)$ consistently
outperforms a broad range of representation manifold baselines across graph
benchmarks at different scales. It reduces average distortion over the
strongest baselines by $44.3\%$ on KEGG and $40.5\%$ on HumanCyc, and improves
Hits@20 by $42.8\%$ on OGBL-PPA. Experiments on Flickr30k-Order further
support its ability to capture higher order dependencies from ordered
composition. Together, these results show how a seemingly simple structural constraint
can yield unexpectedly rich geometry, capacity, and composition within a
unified representation space.
\end{abstract}

\vspace{-10pt}
\section{Introduction}
Finding informative representations of data has long been a central problem in machine
learning, from dimensionality reduction by principal component analysis
\citep{hotelling1933analysis}, through learning compact representations with neural networks
\citep{hinton2006reducing} and distributed representations of words through embeddings
\citep{mikolov2013distributed}, to modern non-Euclidean representations on manifolds
\citep{bronstein2021geometric,diepeveen2025score}.
This progression naturally raises a key question:
\emph{what geometry should the representation space be endowed with?}
The manifold hypothesis provides a fundamental starting point, suggesting that
high dimensional observations often concentrate near manifolds of substantially lower
intrinsic dimension
\citep{tenenbaum2000global,fefferman2016testing}.
Yet it does not determine the geometry of this latent manifold, so the chosen geometry acts as an inductive bias toward particular structural patterns.
For instance, classical Euclidean geometry $\mathbb{E}^n$ (curvature $K=0$) is suited to approximately flat or grid-like structures
\citep{mikolov2013distributed},
hyperbolic geometry $\mathbb{H}^n$ ($K<0$) naturally accommodates hierarchical or tree-like structures
\citep{nickel2017poincare},
whereas spherical geometry $\mathbb{S}^n$ ($K>0$) provides a compact geometry suited to cyclic or clique-like structures
\citep{bachmann2020constant,sun2022self}.

Real-world structures, however, rarely conform to a simple, single-curvature regime.
For example, even a single graph, relational system, or biological network may simultaneously contain
hierarchical, cyclic, densely interconnected, and approximately flat substructures
\citep{mcneela2024product}.
This observation has motivated \textbf{mixed-curvature representation learning}. A straightforward idea is to combine manifolds of different curvatures into a single representation space.
This leads to the widely used \emph{mixed-curvature product manifold}
\citep{gu2019mixed}, defined as the Cartesian product
\begin{equation}
\mathcal{P}
=
\mathbb{E}^{d_0}
\times
\prod_{i=1}^{m_-}\mathbb{H}_{K_i}^{d_i}
\times
\prod_{j=1}^{m_+}\mathbb{S}_{K_j}^{d_j},
\end{equation}
where $d_0,d_i,d_j$ denote the factor dimensions, while
$K_i<0$ and $K_j>0$ denote the corresponding hyperbolic and spherical curvatures. Despite its simplicity, this construction has been successfully extended to generative modeling
and applied across a variety of downstream domains
\citep{skopek2020mixed,bachmann2020constant,wang2021mixed,sun2022self,wang2023mixed}.
However, product mixed-curvature spaces inherit two fundamental limitations.

First, product mixed-curvature spaces require the combination of spaces with different curvatures to be \emph{specified in advance}, while the appropriate combination can vary substantially across data objects.
This strong inductive bias makes a suitable product geometry difficult to determine. For example,
even for a fixed total dimension
$d_0+\sum_{i=1}^{m_-} d_i+\sum_{j=1}^{m_+} d_j=64$ and fixed curvature magnitudes, there are 
$11{,}555{,}651{,}398$\footnote{\proofnote{See Appendix~[\ref{app:product_count}] for how this number is calculated.}} different candidates! 
(note that $\mathbb{S}^{3}\times\mathbb{H}^{1}\neq
\mathbb{S}^{2}\times\mathbb{H}^{2}$ and
$\mathbb{H}^{2}\times\mathbb{H}^{2}\neq\mathbb{H}^{4}$). Thus, related work often requires searching over many product spaces
\citep{mcneela2024product}.

Second, and more importantly, product spaces with mixed curvature have an inherent
\emph{representation limitation}. Let $\mathcal{M}=\prod_i\mathcal{M}_i$ be a product of
component manifolds and $T_x\mathcal{M}$ its tangent space at $x$.
The product geometry satisfies
$T_x\mathcal{M}=\bigoplus_iT_{x_i}\mathcal{M}_i$, while its curvature is generated
independently within each factor, with no curvature interaction across factors
\citep{saez2026expressive}.
Thus, positive and negative curvature coexist through separate factor components rather
than through their intrinsic interaction. This limits the representation of coupled structures, where hierarchical,
cyclic, and other patterns share common features instead of decomposing into independent parts.

These two severe limitations motivate the central question of this work:
\emph{can we find a single geometry that intrinsically represents coupled
mixed-curvature structures, without manually allocating different curvatures
to separate factors?}
Some promising alternatives have emerged from previous works. 
SPD and higher-rank Siegel manifolds provide intrinsic $\{-,0\}$ curvature structures for representation learning
\citep{lopez2021symmetric,zhao2023modeling},
while Grassmann manifolds provide intrinsic $\{+,0\}$ curvature
\citep{bendokat2024grassmann}.
This leaves a pressing open problem in representation learning of realizing
intrinsically coupled $\{-,0,+\}$ curvature within a single latent manifold.

To address this problem, we propose the \textbf{$\mathbf{\mathbb{SL}(n)}$ space} (we omit $p$ when no confusion arises), an intrinsic mixed-curvature space constructed from the special linear group and defined as
\begin{equation}
\mathbb{SL}_p(n):=\bigl(\mathrm{SL}(n),F_p,\circ\bigr),
\quad
\mathrm{SL}(n):=\{A\in\mathbb{R}^{n\times n}\mid \det(A)=1\},
\end{equation}
where $F_p(A,V)=\|A^{-1}V\|_{S_p}$ is the globally defined
Schatten-$p$ tangent norm, introducing direction dependent sensitivity to
matrix variations beyond a quadratic Riemannian metric, and $\circ$ denotes
group composition, naturally given by matrix multiplication. Importantly, $F_p$ and the induced length geometry are
defined for all tangent vectors, while full rank regularity is required only
for smooth curvature analysis and excludes a measure zero set of directions.
The induced Finsler geometry has directional flag curvature $K_F(Y,\Pi)$,
allowing different tangent directions to exhibit different local curvature.
We prove that positive, zero, and negative flag curvatures coexist at every
point, with different signs interacting through shared tangent directions
rather than separate factors. We quantify these two properties by the
mixed-curvature capacity $\mathcal C_{\mathbb{SL}}^{\mathrm{mix}}$ and curvature-coupling
capacity $\mathcal C_{\mathbb{SL}}^{\mathrm{cpl}}$, respectively, capturing balanced
coexistence and genuine coupling. Both are asymptotically maximal relative
to the intrinsic geometric upper bound. Experiments on real-world
graphs of substantially different scales further show strong improvements
over a broad range of geometric baselines.

Beyond geometry, the Lie group structure of $\mathrm{SL}(n)$ provides an
intrinsic mechanism for composition. Its group operation is matrix
multiplication, while its Lie algebra $\mathfrak{sl}(n)$ carries the Lie
bracket $[X,Y]=XY-YX$. Noncommutativity makes sequential composition inherently
order-aware, while the non-nilpotent structure of $\mathfrak{sl}(n)$
permits nonzero nested Lie brackets
$[X_k,[\cdots,[X_2,X_1]\cdots]]$ at arbitrary depth $k$. This enables
progressively deeper ordered composition beyond pairwise noncommutativity,
which we further evaluate empirically.

Together, these properties unify mixed-curvature geometry and order-aware algebraic structure within a single matrix representation space. And due to page limitations, beyond the related work discussed above, a more detailed Related Work section is provided in Appendix~[\ref{app:related_work}].

\section{The $\mathbb{SL}$ Space}
\label{sec:sl_space}

\subsection{Structures of $\mathbb{SL}$ Space}
\label{sec:sl_geometry}

\begin{figure}[h]
 \vspace{-5pt}
    \centering
    \includegraphics[width=1.0\linewidth]{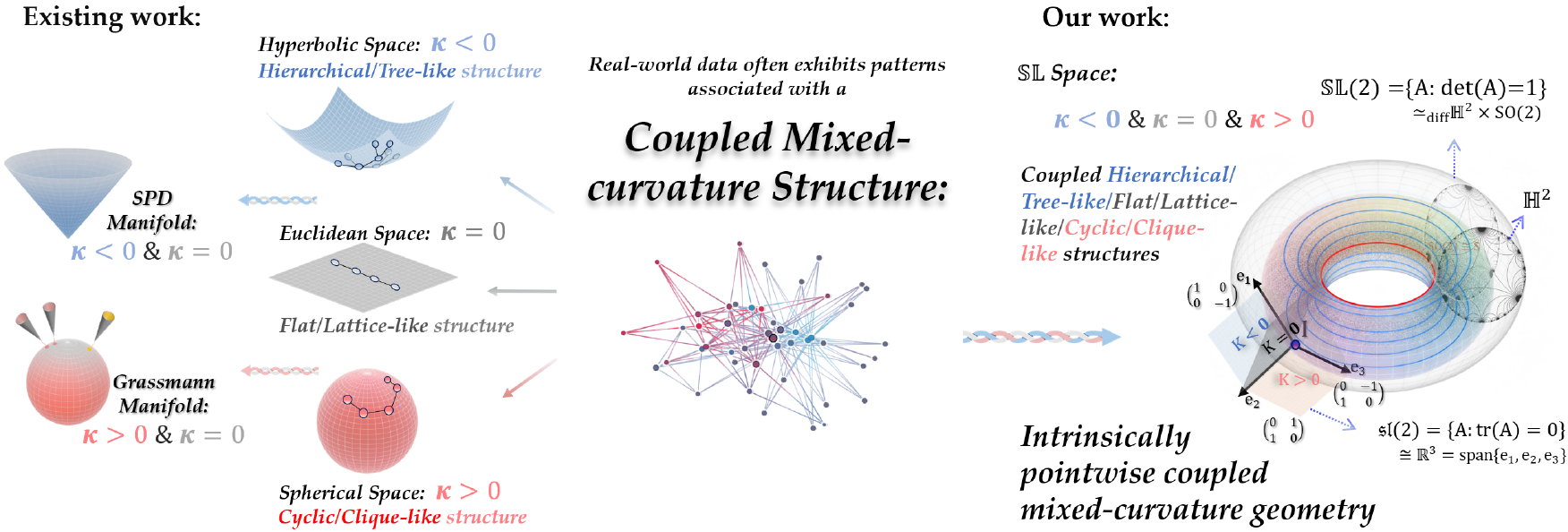}
    \caption{
Existing manifolds capture only restricted curvature regimes, whereas real-world
data may contain complex coupled structures. The $\mathbb{SL}$ space intrinsically
couples $\{-,0,+\}$ curvature within a single geometry. The right side depicts the
global topology and tangent Lie algebra
$\mathfrak{sl}(2)$.
}
    \label{fig:overview}
     \vspace{-2pt}
\end{figure}

As illustrated in Fig.~\ref{fig:overview}, the $\mathbb{SL}(n)$ space provides a 
matrix representation that combines a smooth manifold structure, a flexible
Finsler geometry, and an intrinsic Lie group structure, while supporting coupled
mixed-curvature within a single space. We introduce these structures in turn below.

\subsubsection{Smooth Manifold Structure}
\label{sec:sl_manifold}

A smooth manifold is a space that locally resembles Euclidean space and admits
smooth coordinate systems for differential operations~\citep{lee2012smooth}. Begin with the matrix set
\begin{equation}
\mathrm{SL}(n)
:=
\left\{
A\in\mathbb{R}^{n\times n}:\det(A)=1
\right\}.
\end{equation}
For the smooth map
$\det:\mathbb{R}^{n\times n}\rightarrow\mathbb{R}$,
writing $D\det_A[V]$ for the directional differential of $\det$ at
$A\in\mathrm{SL}(n)$ along $V\in\mathbb{R}^{n\times n}$ in the ambient
Euclidean space,
$D\det_A[V]=\operatorname{tr}(A^{-1}V)$.
Since $D\det_A\neq0$ on $\det^{-1}(1)$, $1$ is a regular value of $\det$.
By the regular level set theorem,
$\mathrm{SL}(n)=\det^{-1}(1)$ is a \textbf{smooth embedded manifold} of
dimension $n^2-1$. Its tangent space is therefore
\begin{equation}
T_A\mathrm{SL}(n)
=
\ker(D\det_A)
=
\left\{
V\in\mathbb{R}^{n\times n}:
\operatorname{tr}(A^{-1}V)=0
\right\}. \vspace{-4pt}
\end{equation}
\subsubsection{Schatten-$p$ Finsler Geometric Structure}
\label{sec:sl_finsler}

A Riemannian structure assigns a smoothly varying inner product $g_x$ to each
tangent space $T_xM$, inducing the norm
$\|V\|_x=\sqrt{g_x(V,V)}$ for tangent vectors. A Finsler structure generalizes
this construction by allowing a smoothly varying tangent norm $F(x,V)$ that
need not arise from an inner product~\citep{bao2000riemannfinsler}. This lets local geometry depend more richly on tangent directions.

For $1\leq p<\infty$, the Schatten-$p$ norm is
\(
\|X\|_{S_p}
:=
\left(\sum_{i=1}^n \sigma_i(X)^p\right)^{1/p},
\)
where $\sigma_i(X)$ is the $i$-th singular value of $X$.
We equip $\mathrm{SL}(n)$ with the \textbf{Schatten-$p$ tangent norm}
\begin{equation}
F_p(A,V):=\|A^{-1}V\|_{S_p},
\quad
A\in\mathrm{SL}(n),\quad
V\in T_A\mathrm{SL}(n).
\end{equation}
We focus on $1<p<\infty$. Importantly, $F_p$ is well defined on the entire
tangent bundle and therefore induces a global length structure on
$\mathrm{SL}(n)$. For $p\neq2$, the smooth differential geometry required for
flag curvature is considered on full rank tangent directions, which form an
open dense set with measure zero complement. The case $p=2$ is globally
Riemannian, while $p=1$ is excluded due to non-smoothness. For a piecewise
smooth curve $\gamma:[0,1]\to\mathrm{SL}(n)$, we define
\(
L_p(\gamma)=\int_0^1F_p(\gamma(t),\dot{\gamma}(t))\,dt
\)
and
\(
d_p(A,B)=\inf_{\gamma:A\to B}L_p(\gamma).
\)

Finsler curvature is described by flag curvature, which generalizes
Riemannian sectional curvature~\citep{bao2000riemannfinsler}. We call a nonzero tangent direction
$Y\in T_A\mathrm{SL}(n)$ regular when $A^{-1}Y$ is full rank.
For a regular tangent direction $Y$ and a two-dimensional plane
$\Pi=\operatorname{span}\{Y,U\}\subset T_A\mathrm{SL}(n)$ containing $Y$,
the pair $(Y,\Pi)$ is called a flag and $Y$ its flagpole.
The fundamental tensor at $Y$ is
\(
g_Y(U,V)
:=
\frac{1}{2}
\left.
\frac{\partial^2}{\partial s\,\partial t}
F_p^2(A,Y+sU+tV)
\right|_{s=t=0}.
\)
Let $\boldsymbol{\mathcal R}^{Y}(U,V)$ denote the Chern curvature
operator with reference direction $Y$, and define the Jacobi operator by
$R_YU:=\boldsymbol{\mathcal R}^{Y}(U,Y)Y$. Their explicit expressions are
deferred to the curvature analysis in Appendix~[\ref{app:proofs}]. The flag
curvature is then defined as
\begin{equation}
K_F(Y,\Pi)
:=
\frac{g_Y(R_YU,U)}
{g_Y(Y,Y)g_Y(U,U)-g_Y(Y,U)^2}.
\end{equation}
Intuitively, flag curvature $K_F(Y,\Pi)$ depends on direction, meaning that
the curvature varies with the flag $\Pi=\operatorname{span}\{Y,U\}$. The Jacobi operator $R_Y$ is self adjoint with respect to $g_Y$, meaning
\(
g_Y(R_YU,V)=g_Y(U,R_YV).
\)
Therefore, all its eigenvalues are real, with positive eigenspace
\begin{equation}
  \label{eigen}  
E_+(Y):=\bigoplus_{\lambda>0}\ker(R_Y-\lambda I),
\end{equation}
and negative eigenspace $E_-(Y)$ defined analogously over $\lambda<0$.

\subsubsection{Lie Group Algebraic Structure}
\label{sec:sl_group}

 Moreover, the underlying manifold carries a natural algebraic group structure.
A group $(G,\circ)$ is a set equipped with an associative composition $\circ$,
an identity element $e$, and an inverse $a^{-1}$ for every $a\in G$.
For $\mathrm{SL}(n)$, matrix multiplication $A\circ B:=AB$, the identity
$e=I$, and matrix inversion $A\mapsto A^{-1}$ define the group structure. 
A Lie group is simultaneously a smooth manifold and a group, with
smooth composition and inversion~\citep{hall2015lie}. Since matrix multiplication and inversion
are smooth on $\mathrm{SL}(n)$, it forms a \textbf{Lie group}. The associated Lie algebra $\mathfrak{sl}(n)$ is the tangent space at the identity, endowed with the Lie bracket $[\cdot,\cdot]$. For $\mathrm{SL}(n)$,
\begin{equation}
\mathfrak{sl}(n)
=
T_I\mathrm{SL}(n)
=
\{\Xi\in\mathbb{R}^{n\times n}:\operatorname{tr}(\Xi)=0\},
\quad
[X,Y]=XY-YX.
\end{equation}
Left translation identifies every tangent space with the same Lie algebra,
giving
$T_A\mathrm{SL}(n)=A\mathfrak{sl}(n)$.
Consequently, every $V\in T_A\mathrm{SL}(n)$ admits the unique
left-trivialized coordinate $\Xi=A^{-1}V\in\mathfrak{sl}(n)$. 
An arbitrary matrix $X\in\mathbb{R}^{n\times n}$ can be mapped to the Lie algebra by $\Pi_{\mathfrak{sl}}(X)$:
\begin{equation}
\Pi_{\mathfrak{sl}}(X)
=
X-\operatorname{tr}(X)I/n
\in\mathfrak{sl}(n).
\end{equation}
The matrix exponential and logarithm then provide natural local mappings
between the tangent space at $A$ and the manifold. From a tangent vector
$V\in T_A\mathrm{SL}(n)$, the exponential gives the constraint preserving
retraction map $R_A$:
\begin{equation}
R_A:T_A\mathrm{SL}(n)\rightarrow\mathrm{SL}(n),
\quad
R_A(V)
=
A\exp(A^{-1}V),
\end{equation}
where the matrix exponential is defined by
$\exp(X)=\sum_{k=0}^{\infty}X^k/k!$.
Since $\det(\exp\Xi)$$=\exp(\operatorname{tr}\Xi)=1$ for
$\Xi\in\mathfrak{sl}(n)$, the retraction remains in $\mathrm{SL}(n)$.

\begin{wrapfigure}{r}{0.27\textwidth}
\vspace{-16pt}
  \centering
\includegraphics[width=0.27\textwidth]{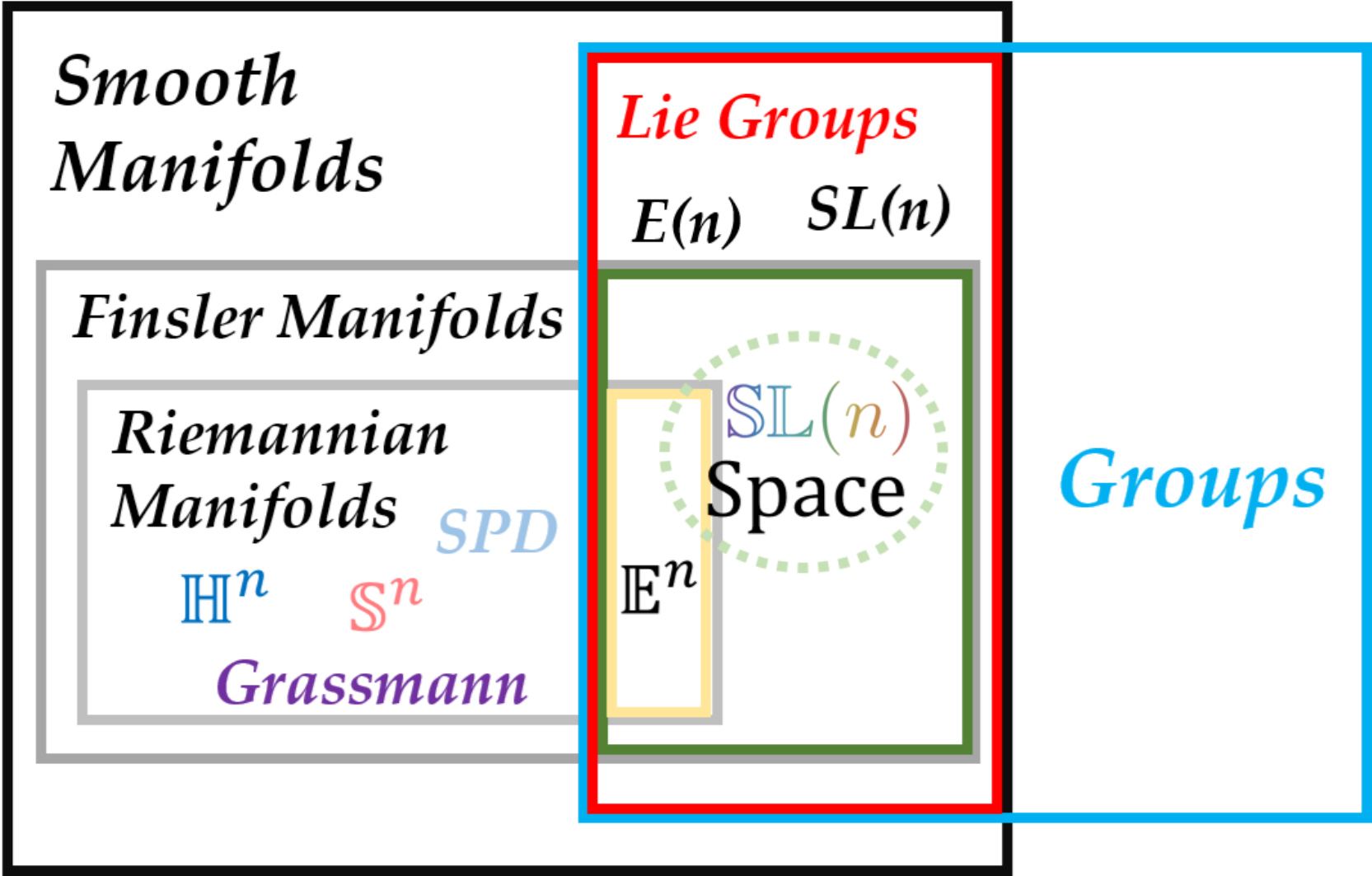}
  \caption{The $\mathbb{SL}(n)$ space combines Finsler geometry with Lie group structure, reducing to the Riemannian case at $p=2$.}
  \label{fig:boundary}
\end{wrapfigure}
Conversely, on a neighborhood of $A$, the retraction admits a local inverse.
The principal matrix logarithm
$\operatorname{log}(M)$ is the unique matrix satisfying
$\exp(\operatorname{log}M)=M$ whose eigenvalues have imaginary parts in
$(-\pi,\pi)$. Hence, locally around $A$,
\begin{equation}
R_A^{-1}:\mathrm{SL}(n)\rightarrow T_A\mathrm{SL}(n),
\quad
R_A^{-1}(B)
=
A\,\operatorname{log}(A^{-1}B),
\end{equation}
We can also define the closed form
Schatten semidistance
\begin{equation}
D_{\mathbb{SL}}(A,B)
:=
\|\operatorname{log}(A^{-1}B)\|_{S_p}.
\end{equation}
It is nonnegative, symmetric, and point separating, but need not satisfy the
triangle inequality or require tangent regularity. We use
$D_{\mathbb{SL}}$ as the pairwise dissimilarity for representation learning.

The Lie bracket on $\mathfrak{sl}(n)$ is the matrix commutator
$[X,Y]=XY-YX$, which quantifies the change induced by reversing the order of
two infinitesimal transformations and provides an intrinsic mechanism for
order-aware composition.
Locally, successive group transformations are related to nested Lie brackets
through the BCH expansion,
\(
\operatorname{log}\!\left(\exp X\exp Y\right)
=
X+Y+\frac{1}{2}[X,Y]
+\frac{1}{12}[X,[X,Y]]
+\cdots .
\)
Overall, these structures combine smooth manifold, Schatten-$p$ Finsler geometric,
and Lie group algebraic structures, supporting both intrinsic geometry and ordered composition. Finally, we define the unified
$\mathbb{SL}$ representation space as
\begin{equation}
\mathbf{\mathbb{SL}_p(n)
:=
\bigl(\mathrm{SL}(n),F_p,\circ\bigr)}.
\end{equation}

\subsection{Intrinsic Coupled Mixed Curvature}
\label{sec:mixed_curvature}

\subsubsection{Pointwise Intrinsic Coupled Mixed Flag Curvature}

Having defined the $\mathbb{SL}$ space, we now characterize its intrinsic
curvature structure. Hereafter, we write $A\in\mathbb{SL}_p(n)$ and
$T_A\mathbb{SL}_p(n):=T_A\mathrm{SL}(n)$ (with $p$ omitted when unambiguous)
when referring to the resulting representation space. We call mixed-curvature intrinsic when different curvature regimes arise within a single tangent geometry and are not attributable to separate factors of a metric product.
$\mathbb{SL}_p(n)$ exhibits such intrinsic mixed-curvature with different
curvature signs within the same tangent geometry.

\begin{theorem}[Pointwise mixed flag curvature]
\label{thm:sl_mixed}
For every $n\geq2$ and \textbf{every} $p\in(1,\infty)$, for \textbf{every}
$A\in\mathbb{SL}_p(n)$, there exist a regular full-rank matrix (flagpole)
$Y_A\in T_A\mathbb{SL}_p(n)$ and three tangent matrices
$U_A^+,U_A^0,U_A^-\in T_A\mathbb{SL}_p(n)$, each linearly independent of $Y_A$,
such that
\[
K_F(Y_A,\operatorname{span}\{Y_A,U_A^+\})>0,\quad
K_F(Y_A,\operatorname{span}\{Y_A,U_A^0\})=0,\quad
K_F(Y_A,\operatorname{span}\{Y_A,U_A^-\})<0.
\]
\end{theorem}
\begin{proof}
\proofnote{The full and detailed proof is in
Appendix~[\ref{app:rootwise_curvature}].}
\end{proof}
Hence, $\mathbb{SL}(n)$ exhibits pointwise ${-,0,+}$ mixed flag curvature around a common flagpole. We next distinguish genuine curvature coupling from mere coexistence.
\begin{definition}[Mixed-curvature and coupling capacities]
\label{def:coupling_capacity}
For a Finsler manifold $(\mathcal M,F)$ and a regular flagpole
$Y\in T_x\mathcal M$, let $E_+(Y)$ and $E_-(Y)$ be the positive and negative
eigenspaces of $R_Y$, respectively, as defined in Eq.~\ref{eigen}. Define the mixed-curvature capacity as
\begin{equation}
\mathcal C_{\mathrm{mix}}(Y)
:=
\min\{\dim E_+(Y),\dim E_-(Y)\}.
\end{equation}
Define the uncoupled subspaces
$\mathcal N_+(Y):=
\{U\in E_+(Y):\boldsymbol{\mathcal R}^{Y}(U,V)=0,\ \forall V\in E_-(Y)\}$
and $\mathcal N_-(Y)$ analogously. The curvature coupling capacity is
\begin{equation}
\mathcal C_{\mathrm{cpl}}(Y)
:=
\min\{
\dim E_+(Y)-\dim\mathcal N_+(Y),\,
\dim E_-(Y)-\dim\mathcal N_-(Y)
\}.
\end{equation}
At $x\in\mathcal M$, a flagpole level capacity $\mathcal C(Y)$ induces
$\mathcal C(x):=\max_{Y\in T_x\mathcal M\ {\rm regular}}\mathcal C(Y)$ and
$\mathcal C_{\mathcal M}(F):=\min_{x\in\mathcal M}\mathcal C(x)$.
We use this convention for both $\mathcal C_{\mathrm{mix}}$ and
$\mathcal C_{\mathrm{cpl}}$.
\end{definition}
Intuitively, $\mathcal C_{\mathrm{mix}}$ measures the balanced number of
positive and negative curvature modes coexisting around a common flagpole,
whereas $\mathcal C_{\mathrm{cpl}}$, like rank in linear algebra after excluding null directions,
counts only those modes that interact across curvature signs. Hence
$\mathcal C_{\mathrm{cpl}}\leq\mathcal C_{\mathrm{mix}}$.
Zero curvature is not counted separately, as it follows between positive and
negative flag curvatures by continuity.
\begin{corollary}[Asymptotically maximal mixed curvature and coupling]
\label{thm:asymptotic_coupling}
For every $n\geq2$ and $p\in(1,\infty)$, since the transverse tangent space
has dimension $n^2-2$, define
\(
\mathcal C_{\max}(n):=\lfloor(n^2-2)/2\rfloor.
\)
Then
\begin{equation}
\begin{aligned}
\frac{(n-1)(n-2)}{2}
&\leq
\mathcal C_{\mathbb{SL}}^{\mathrm{cpl}}(n,p)
\leq
\mathcal C_{\mathbb{SL}}^{\mathrm{mix}}(n,p)
\leq
\mathcal C_{\max}(n),
\\
\frac{\mathcal C_{\mathbb{SL}}^{\mathrm{cpl}}(n,p)}
{\mathcal C_{\max}(n)}
&\longrightarrow1,
\qquad
\frac{\mathcal C_{\mathbb{SL}}^{\mathrm{mix}}(n,p)}
{\mathcal C_{\max}(n)}
\longrightarrow1
\quad\text{as }n\rightarrow\infty.
\end{aligned}
\end{equation}
\end{corollary}
\begin{proof}
\proofnote{The full and detailed proof is in
Appendix~[\ref{proof:2.3}].}
\end{proof}
Accordingly, $\mathbb{SL}(n)$ supports asymptotically maximal coexistence of
positive and negative curvature modes while intrinsically coupling an
asymptotically maximal number of these modes.

\subsection{Deep Order-Aware Composition}
\label{sec:order_composition}

Beyond its geometric structure, $\mathbb{SL}(n)$ provides an intrinsic
mechanism for representing ordered interactions through noncommutative group
composition. For $A,B\in\mathbb{SL}(n)$, generally $AB\neq BA$, so reversing
their order changes the composition. Locally, this difference is captured by
the Lie bracket $[X,Y]=XY-YX$ through the BCH expansion introduced above.

\begin{figure}[t]
    \centering
    \includegraphics[width=1.0\linewidth]{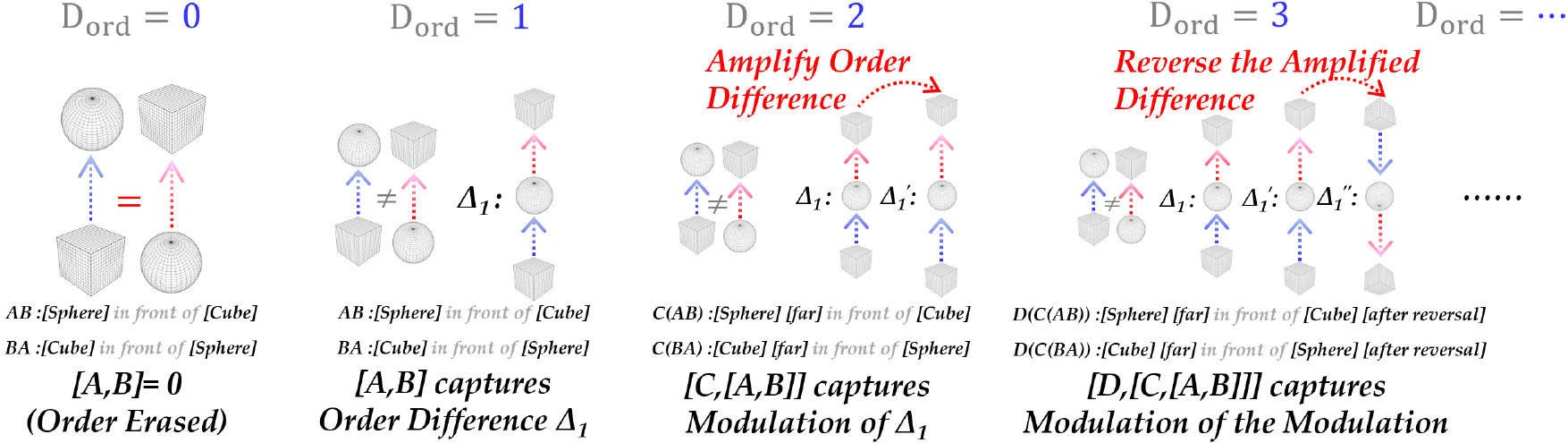}
    \caption{Illustration of order depth. Increasing $D_{\rm ord}$ enables progressively
deeper order-dependent interactions, from pairwise order sensitivity to
higher level modulation through nested Lie brackets.}
    \label{fig:order}
    \vspace{-10pt}
\end{figure}

Pairwise noncommutativity, however, captures only first order interactions.
Successive compositions may further modulate existing order differences through
nested Lie brackets, as illustrated in Fig.~\ref{fig:order}. We therefore
quantify the depth of such interactions by the following notion.
\begin{lemma}[Order depth of $\mathbb{SL}$]
\label{thm:sl_order_depth}
For a Lie group representation space $\mathcal{M}$, define its order depth as
\begin{equation}
D_{\mathrm{ord}}(\mathcal{M})
:=
\sup\left\{
k\geq1:
\exists\,X_0,\ldots,X_k\in T_I\mathcal{M}
\text{ such that }
[X_k,[\cdots,[X_1,X_0]\cdots]]\neq0
\right\}.
\end{equation}
We set $D_{\mathrm{ord}}(\mathcal{M})=0$ when all brackets vanish and
$D_{\mathrm{ord}}(\mathcal{M})=\infty$ when nonzero nested brackets exist
at arbitrary depth. Then, for every $n\geq2$ and every $p\in(1,\infty)$,
\begin{equation}
D_{\mathrm{ord}}\!\left(\mathbb{SL}_p(n)\right)=\infty .
\end{equation}
\end{lemma}
\begin{proof}
\textcolor{citegray}{The full and detailed proof is in
Appendix~[\ref{proof:2.4}].}
\end{proof}

Thus, $D_{\mathrm{ord}}$ characterizes the depth at which nested
order-dependent interactions can remain nonzero. A nilpotent Lie algebra of
class $c$ has $D_{\mathrm{ord}}=c-1$, whereas $\mathbb{SL}(n)$ has
$D_{\mathrm{ord}}=\infty$. This provides algebraic capacity for order-aware
composition beyond pairwise noncommutativity. We examine the empirical
relevance of this property on Flickr30k-Order in Sec.~\ref{sec:flickr}.
\begin{table}[h]
\centering
\caption{
Comparison of geometric representation spaces and their structural properties.
}
\label{tab:space_properties}

\resizebox{0.80\linewidth}{!}{%
\begin{tabular}{lcccccccc}
\toprule
& \multicolumn{4}{c}{Mixed curvature}
& \multicolumn{2}{c}{Curvature coupling}
& \multicolumn{2}{c}{Native group composition} \\
\cmidrule(lr){2-5}
\cmidrule(lr){6-7}
\cmidrule(lr){8-9}
Manifold Space
& Mixed
& Signs
& Intrinsic
& Capacity $\mathcal C_{\mathcal M}^{\mathrm{mix}}$
& Coupled
& Capacity $\mathcal C_{\mathcal M}^{\mathrm{cpl}}$
& Noncommutative
& $D_{\mathrm{ord}}$ \\
\midrule

$\mathbb{S}^{d}$
& \nocell
& $\{+\}$
& \nacell
& \nacell
& \nacell
& \nacell
& \nacell
& \nacell \\

$\mathbb{H}^{d}$
& \nocell
& $\{-\}$
& \nacell
& \nacell
& \nacell
& \nacell
& \nacell
& \nacell \\

$\mathbb{E}^{d}$
& \nocell
& $\{0\}$
& \nacell
& \nacell
& \nacell
& \nacell
& \nocell
& $\mathbf{0}$ \\

$\mathrm{Grassmann}(k,n)$
& \yescell
& $\{0,+\}$
& \yescell
& \nacell
& \nacell
& \nacell
& \nacell
& \nacell \\

$\mathrm{SPD}(n)$
& \yescell
& $\{-,0\}$
& \yescell
& \nacell
& \nacell
& \nacell
& \nacell
& \nacell \\

$\mathrm{Siegel}(n)$
& \yescell
& $\{-,0\}$
& \yescell
& \nacell
& \nacell
& \nacell
& \nacell
& \nacell \\

$\mathbb{S}^{d_S}_{\kappa_1}
\times
\mathbb{H}^{d_H}_{\kappa_2}
\times
\mathbb{E}^{d_E}$
& \yescell
& $\{-,0,+\}$
& \nocell
& $\min\{d_S-1,d_H-1\}$
& \nocell
& $\mathbf{0}$
& \nacell
& \nacell \\

Heisenberg-$H^{2m+1}$
& \yescell
& $\{-,0,+\}$
& \yescell
& $\mathbf{1}$
& \yescell
& $\mathbf{1}$
& \yescell
& $\mathbf{1}$ \\

$\mathbb{SL}(n)$
& \yescell
& $\{-,0,+\}$
& \yescell
& $\boldsymbol{\geq}\binom{\mathbf{n-1}}{\mathbf{2}}$
& \yescell
& $\boldsymbol{\geq}\binom{\mathbf{n-1}}{\mathbf{2}}$
& \yescell
& $\boldsymbol{\infty}$ \\

\bottomrule
\end{tabular}%
}
\end{table}
\vspace{-8pt}

Overall, Table~\ref{tab:space_properties} shows that $\mathbb{SL}(n)$ uniquely
combines intrinsic $\{-,0,+\}$ mixed-curvature, high curvature-coupling
capacity, and unbounded order depth within a single representation space.
\vspace{-5pt}

\section{Experiments}
\label{sec:experiments}

We first examine the practical training ability of $\mathbb{SL}(n)$. The intrinsic
Finsler distance has no simple closed form and requires costly path optimization.
We therefore use the Schatten semidistance
$D_{\mathbb{SL}}(A,B)=\|\log(A^{-1}B)\|_{S_p}$, which locally approximates
the intrinsic distance to second order while being substantially cheaper to
compute. Optimization presents another challenge, as general Finsler manifolds
lack mature adaptive optimization methods such as AdamW. We therefore use an
exponential parameterized AdamW scheme for $\mathbb{SL}(n)$, where the
Schatten-$p$ geometry enters through the representation objective, and compare
it against a Riemannian AdamW control. Full analyses and algorithms are
provided in Appendix~[\ref{app:semidistance_geodesic}][\ref{sec:sl_optimization}].

For experiments, we evaluate two central properties of $\mathbb{SL}(n)$:
its ability to represent mixed curvature structures and its order aware
composition induced by noncommutative group multiplication. We first assess
geometric representation through graph reconstruction on biological networks
of increasing scale, and further evaluate downstream utility through large
scale link prediction on OGBL-PPA. We then isolate compositional ability on
Flickr30k-Order. Full implementation details, geometric parameterizations,
distance functions, and additional results are provided in
Appendix~[\ref{app:experimental_details}].

\subsection{Mixed-Curvature Graph Representation}
\begin{figure}[h]
    \centering

    \begin{subfigure}[t]{0.32\linewidth}
        \centering
        \includegraphics[width=\linewidth]{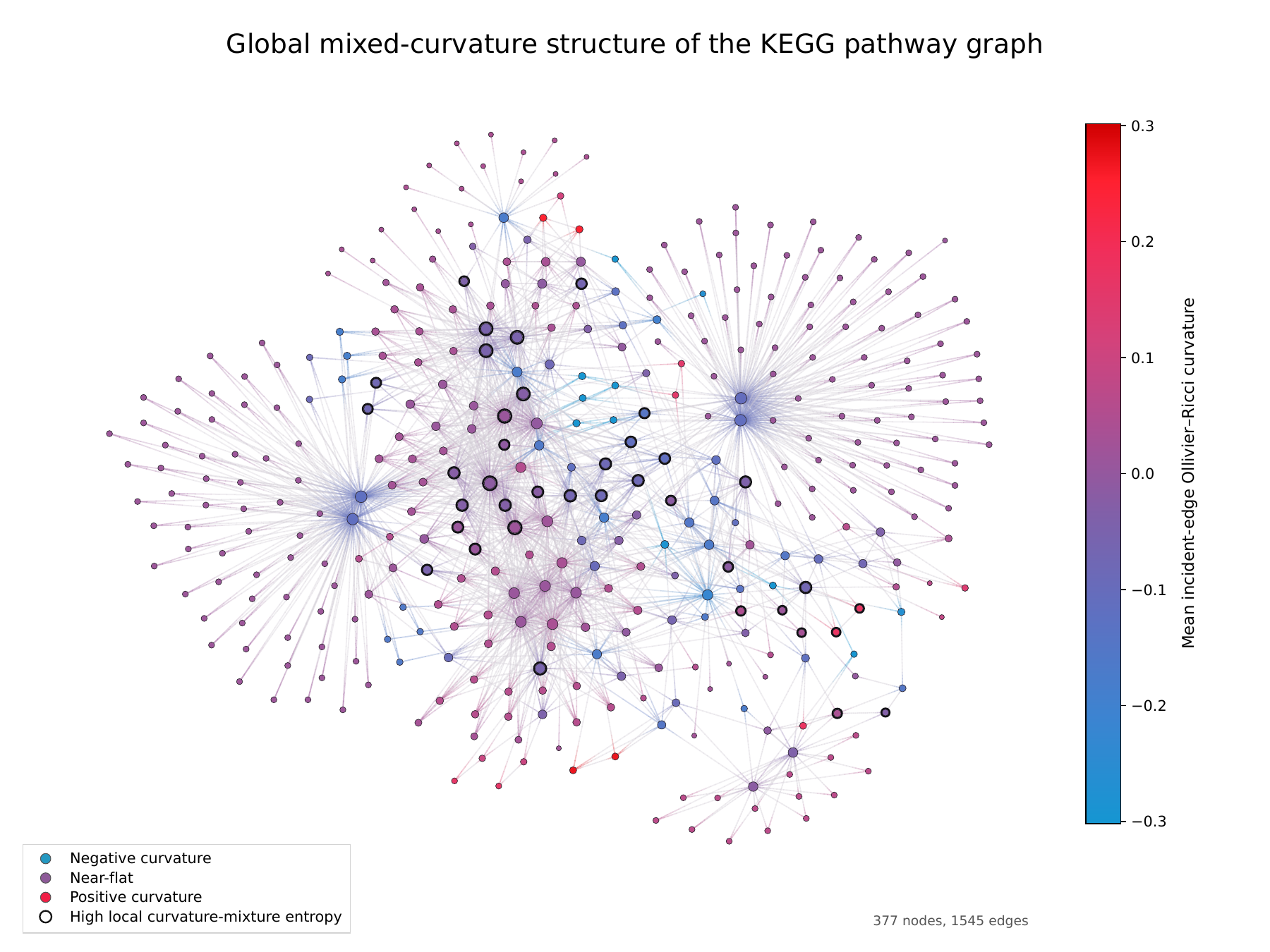}
        \caption{
\begin{tabular}[t]{@{}c@{}}
$\text{KEGG:}~|V| = 377,$\\
$|E| = 1{,}545$.
\end{tabular}
}
        \label{fig:subfig_a}
    \end{subfigure}
    \hfill
    \begin{subfigure}[t]{0.32\linewidth}
        \centering
        \includegraphics[width=\linewidth]{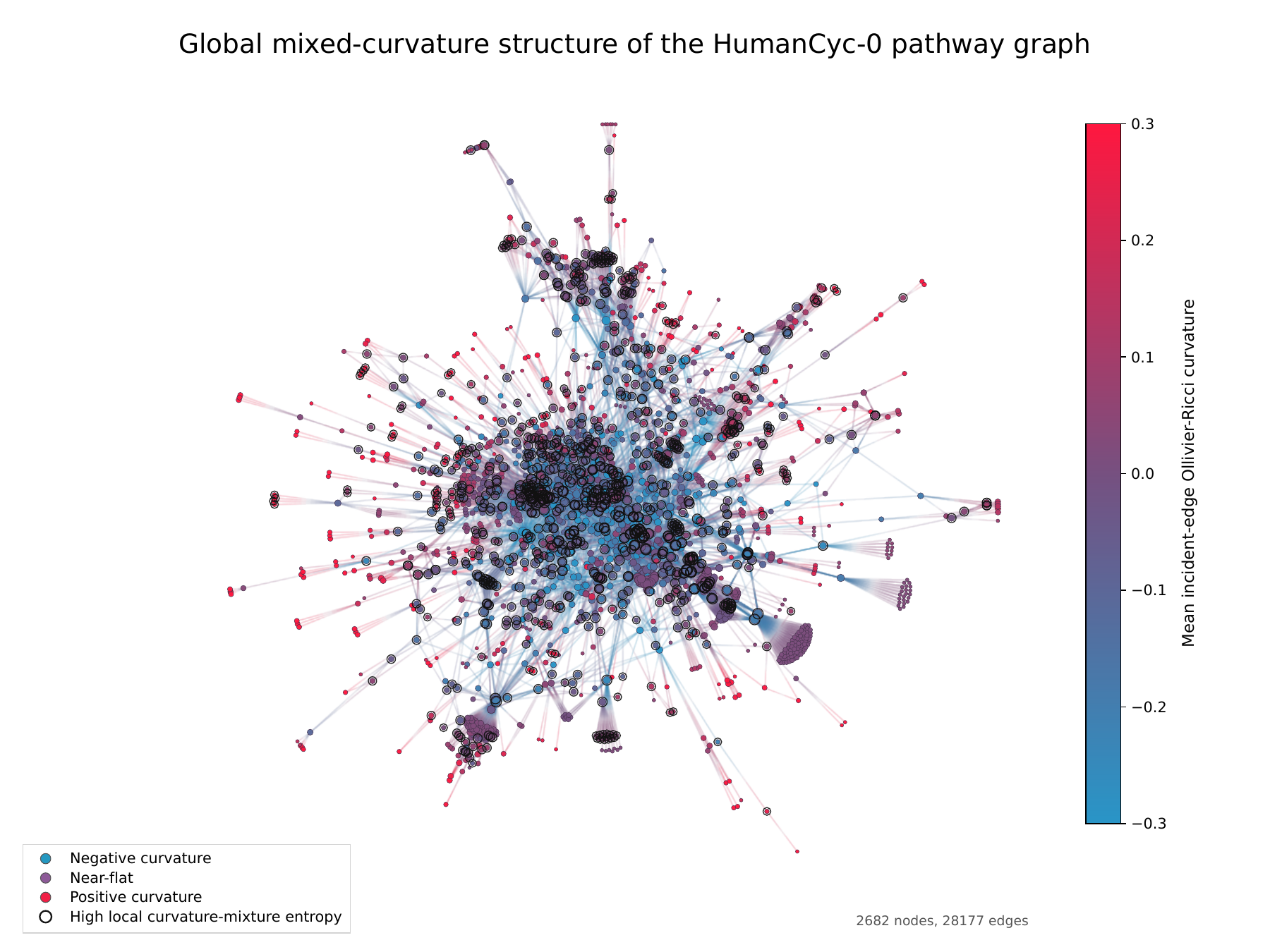}
        \caption{ \begin{tabular}[t]{@{}c@{}} $\text{HumanCyc:}~|V| = 2{,}682,$\\ $|E| = 28{,}177$. \end{tabular} }
        \label{fig:subfig_b}
    \end{subfigure}
   \hfill
    \begin{subfigure}[t]{0.32\linewidth}
        \centering
        \includegraphics[width=\linewidth]{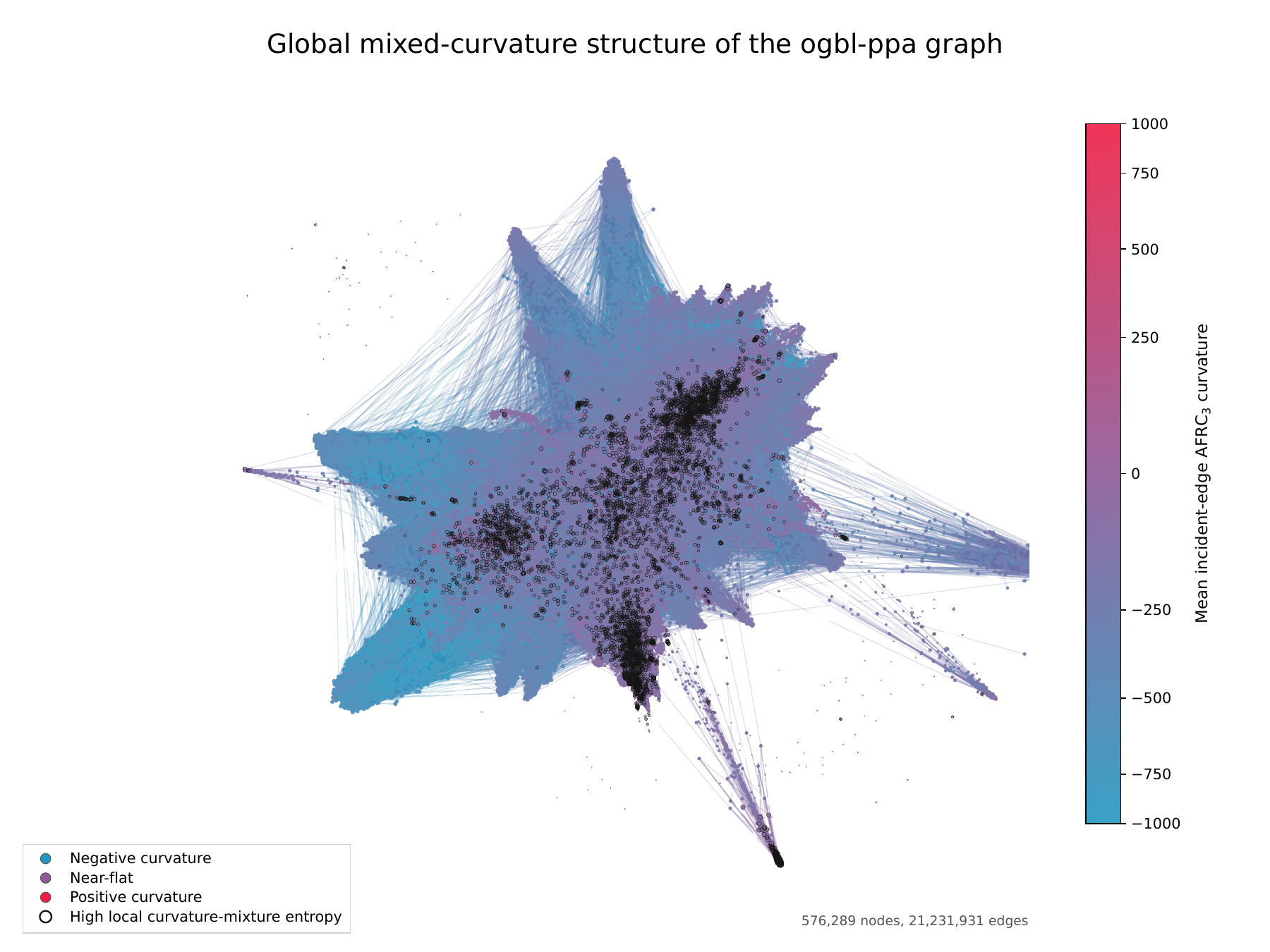}
        \caption{ \begin{tabular}[t]{@{}c@{}} $\text{OGBL:}~|V| = 576{,}289,$\\ $|E| = 21{,}231{,}931$. \end{tabular} }
        \label{fig:subfig_b}
    \end{subfigure}
    \caption{Graph curvature distributions of three
biological networks of increasing scale. All three graphs exhibit mixed-curvature regimes 
with rich interactions structures.}
    \label{fig:two_subfigures}
    \vspace{-14pt}
\end{figure}
\paragraph{Metric reconstruction on KEGG and HumanCyc.}
We evaluate geometric representation capacity by reconstructing the shortest
path metrics of KEGG and HumanCyc. Each node is represented by a learnable
point, while all methods use the same training pairs, data split, regression
objective, and evaluation protocol, differing mainly in the latent geometry
and its pairwise dissimilarity. Given graph distance $d_G(i,j)$ and latent
dissimilarity $D_{\mathcal M}(z_i,z_j)$, all geometries optimize
\(
\mathcal L
=
\frac{1}{|\mathcal B|}
\sum_{(i,j)\in\mathcal B}
[
\log(1+sD_{\mathcal M}(z_i,z_j))
-
\log(1+d_G(i,j))
]^2,
\)
where $s>0$ is a learned global scale and $\mathcal B$ is a training minibatch. Geometric fidelity is measured by
\begin{equation}
\operatorname{Distortion}_{\mathrm{avg}}
=
\frac{1}{|\mathcal T|}
\sum_{(i,j)\in\mathcal T}
\frac{
|sD_{\mathcal M}(z_i,z_j)-d_G(i,j)|
}{
\max\{d_G(i,j),1\}
}.
\end{equation}
Here $\mathcal T$ denotes the test pairs. Distortion measures deformation of the original graph metric. We further report $q50$, $q90$, and $q95$ for typical and tail errors. Local curvature mixture entropy is defined as \(H_\kappa(\mathcal N(v)) =-\sum_{s\in\{-,0,+\}}p_v^\kappa(s)\log p_v^\kappa(s)\), where \(p_v^\kappa(s)= |\{u\in\mathcal N(v):\kappa_{vu}\in s\}|/|\mathcal N(v)|\). Higher entropy indicates stronger local mixing of curvature regimes. High-related measures pairs involving high entropy nodes, Low-Low focuses on pairs between low entropy nodes, and Worst-group reports the largest mean distortion across entropy groups. We first conduct a compact comparison using $\mathbb{SL}(4)$ and commonly used geometries, followed by a higher dimensional comparison. Dimensions are matched as closely as possible; for matrix manifolds, we align matrix size rather than intrinsic dimension. Product manifolds permit mixed-curvature coexistence but have zero coupling, while Heisenberg has both capacities equal to $1$. We find that $\mathbb{SL}(n)$ not only improves all reported metrics, but achieves its largest gains in high entropy regions, particularly over mixed-curvature product manifolds and compared with Low-Low pairs. Since product manifolds allow mixed-curvature coexistence without coupling, this concentration of gains in strongly mixed regions suggests that capturing coupled curvature may be important for representation learning.
\vspace{-5pt}
\begin{table}[h]
    \centering
    \caption{
    KEGG metric reconstruction results for geometric baselines. (Lower is better)
     }
    \label{tab:kegg_constant_curvature}

    \resizebox{0.95\linewidth}{!}{%
    \begin{tabular}{lrrrrrrrr}
    \toprule
    Space
    & \textcolor{DeepKleinBlue}{\textbf{Distortion$_{\mathrm{avg.}}$}} $\downarrow$
    & q50 $\downarrow$
    & q90 $\downarrow$
    & q95 $\downarrow$
    & High-related $\downarrow$
    & Low--Low $\downarrow$
    & Worst-group $\downarrow$
    & MAE $\downarrow$ \\
    \midrule

    \rowcolor{black!2}
    \textcolor{lightMagenta}{$\mathbb{S}^{15}$}
    & \heatcell{2}{$0.0992 {\scriptstyle \pm 0.0006}$}
    & \heatcell{2}{$0.0636$}
    & \heatcell{2}{$0.2166$}
    & \heatcell{2}{$0.3173$}
    & \heatcell{2}{$0.1125$}
    & \heatcell{2}{$0.0989$}
    & \heatcell{3}{$0.1850$}
    & \heatcell{2}{$0.2522$} \\

    \rowcolor{black!2}
    \textcolor{KleinBlue}{$\mathbb{H}^{15}$}
    & \heatcell{4}{$0.0905 {\scriptstyle \pm 0.0002}$}
    & \heatcell{4}{$0.0589$}
    & \heatcell{4}{$0.2008$}
    & \heatcell{6}{$0.2798$}
    & \heatcell{4}{$0.1113$}
    & \heatcell{6}{$0.0755$}
    & \heatcell{2}{$0.1857$}
    & \heatcell{4}{$0.2341$} \\

    \rowcolor{black!2}
    $\mathbb{E}^{15}$
    & \heatcell{3}{$0.0953 {\scriptstyle \pm 0.0002}$}
    & \heatcell{3}{$0.0625$}
    & \heatcell{3}{$0.2107$}
    & \heatcell{3}{$0.3005$}
    & \heatcell{3}{$0.1114$}
    & \heatcell{3}{$0.0893$}
    & \heatcell{4}{$0.1837$}
    & \heatcell{3}{$0.2435$} \\

    \rowcolor{black!2}
    $\textcolor{lightMagenta}{\mathbb{S}^{5}}
    \times
    \textcolor{KleinBlue}{\mathbb{H}^{5}}
    \times
    \mathbb{E}^{5}$
    & \heatcell{13}{$0.0724 {\scriptstyle \pm 0.0003}$}
    & \heatcell{13}{$0.0427$}
    & \heatcell{13}{$0.1597$}
    & \heatcell{12}{$0.2403$}
    & \heatcell{14}{$0.0933$}
    & \heatcell{13}{$0.0549$}
    & \heatcell{13}{$0.1711$}
    & \heatcell{13}{$0.1849$} \\

    \rowcolor{black!2}
    $\textcolor{lightMagenta}{\mathbb{S}^{5}_{\kappa_1}}
    \times
    \textcolor{KleinBlue}{\mathbb{H}^{5}_{\kappa_2}}
    \times
    \mathbb{E}^{5}$
    & \heatcell{16}{$0.0613 {\scriptstyle \pm 0.0005}$}
    & \heatcell{16}{$0.0338$}
    & \heatcell{16}{$0.1404$}
    & \heatcell{17}{$0.2070$}
    & \heatcell{16}{$0.0890$}
    & \heatcell{19}{$0.0325$}
    & \heatcell{17}{$0.1642$}
    & \heatcell{16}{$0.1558$} \\

    \rowcolor{black!2}
    \rainbowmath{\mathbb{SL}(4)}
    & \heatcell{20}{$0.0586 {\scriptstyle \pm 0.0016}$}
    & \heatcell{19}{$0.0288$}
    & \heatcell{20}{$0.1238$}
    & \heatcell{20}{$0.1947$}
    & \heatcell{20}{$0.0667$}
    & \heatcell{12}{$0.0556$}
    & \heatcell{20}{$0.1338$}
    & \heatcell{19}{$0.1485$} \\

    \midrule

    \rowcolor{black!2}
    $\textcolor{lightMagenta}{\mathbb{S}^{32}}
    \times
    \textcolor{KleinBlue}{\mathbb{H}^{32}}$
    & \heatcell{12}{$0.0744 {\scriptstyle \pm 0.0001}$}
    & \heatcell{12}{$0.0454$}
    & \heatcell{12}{$0.1616$}
    & \heatcell{14}{$0.2330$}
    & \heatcell{12}{$0.0986$}
    & \heatcell{14}{$0.0522$}
    & \heatcell{12}{$0.1759$}
    & \heatcell{12}{$0.1887$} \\

    \rowcolor{black!2}
    $\textcolor{lightMagenta}{\mathbb{S}^{32}}
    \times
    \mathbb{E}^{32}$
    & \heatcell{7}{$0.0842 {\scriptstyle \pm 0.0002}$}
    & \heatcell{10}{$0.0505$}
    & \heatcell{10}{$0.1730$}
    & \heatcell{7}{$0.2775$}
    & \heatcell{10}{$0.1028$}
    & \heatcell{7}{$0.0725$}
    & \heatcell{10}{$0.1765$}
    & \heatcell{9}{$0.2108$} \\

    \rowcolor{black!2}
    $\textcolor{KleinBlue}{\mathbb{H}^{32}}
    \times
    \mathbb{E}^{32}$
    & \heatcell{14}{$0.0706 {\scriptstyle \pm 0.0000}$}
    & \heatcell{14}{$0.0420$}
    & \heatcell{14}{$0.1550$}
    & \heatcell{13}{$0.2332$}
    & \heatcell{13}{$0.0966$}
    & \heatcell{16}{$0.0464$}
    & \heatcell{14}{$0.1689$}
    & \heatcell{14}{$0.1788$} \\

    \rowcolor{black!2}
    $\textcolor{lightMagenta}{\mathbb{S}^{21}_{\kappa_1}}
    \times
    \textcolor{KleinBlue}{\mathbb{H}^{21}_{\kappa_2}}
    \times
    \mathbb{E}^{21}$
    & \heatcell{19}{$\underline{0.0591 {\scriptstyle \pm 0.0002}}$}
    & \heatcell{20}{$\underline{0.0286}$}
    & \heatcell{17}{$0.1371$}
    & \heatcell{19}{$\underline{0.2013}$}
    & \heatcell{17}{$0.0862$}
    & \heatcell{20}{$\underline{0.0295}$}
    & \heatcell{16}{$0.1683$}
    & \heatcell{17}{$0.1502$} \\

    \rowcolor{black!2}
    Heisenberg-$H^{31}$
    & \heatcell{10}{$0.0814 {\scriptstyle \pm 0.0001}$}
    & \heatcell{7}{$0.0526$}
    & \heatcell{9}{$0.1764$}
    & \heatcell{10}{$0.2432$}
    & \heatcell{7}{$0.1088$}
    & \heatcell{10}{$0.0572$}
    & \heatcell{9}{$0.1804$}
    & \heatcell{10}{$0.2073$} \\

    \rowcolor{black!2}
    $\textcolor{deepPurple}{\mathbf{Gr}}(7,16)$
    & \heatcell{6}{$0.0879 {\scriptstyle \pm 0.0010}$}
    & \heatcell{9}{$0.0516$}
    & \heatcell{7}{$0.1787$}
    & \heatcell{4}{$0.2926$}
    & \heatcell{9}{$0.1088$}
    & \heatcell{4}{$0.0763$}
    & \heatcell{6}{$0.1836$}
    & \heatcell{6}{$0.2190$} \\

    \rowcolor{black!2}
    $\textcolor{lightKleinBlue}{\mathbf{SPD}(8)}$
    & \heatcell{9}{$0.0842 {\scriptstyle \pm 0.0004}$}
    & \heatcell{6}{$0.0550$}
    & \heatcell{6}{$0.1821$}
    & \heatcell{9}{$0.2526$}
    & \heatcell{6}{$0.1095$}
    & \heatcell{9}{$0.0620$}
    & \heatcell{7}{$0.1805$}
    & \heatcell{7}{$0.2160$} \\

    \rowcolor{black!2}
    $\textcolor{DeepKleinBlue}{\mathbf{Siegel}(8)}$
    & \heatcell{17}{$0.0601 {\scriptstyle \pm 0.0004}$}
    & \heatcell{17}{$0.0328$}
    & \heatcell{19}{$\underline{0.1339}$}
    & \heatcell{16}{$0.2109$}
    & \heatcell{19}{$\underline{0.0789}$}
    & \heatcell{17}{$0.0454$}
    & \heatcell{19}{$\underline{0.1539}$}
    & \heatcell{20}{$\underline{0.1475}$} \\

    \rowcolor{black!2}
    \rainbowmath{\mathbb{SL}(8)}
    & \heatcell{22}{
      \textcolor{DeepKleinBlue}{
        $\bm{0.0329} {\scriptstyle \bm{\pm 0.0001}}$
      }}
    & \heatcell{22}{\textcolor{DeepKleinBlue}{$\bm{0.0115}$}}
    & \heatcell{22}{\textcolor{DeepKleinBlue}{$\bm{0.0698}$}}
    & \heatcell{22}{\textcolor{DeepKleinBlue}{$\bm{0.1207}$}}
    & \heatcell{22}{\textcolor{DeepKleinBlue}{$\bm{0.0389}$}}
    & \heatcell{22}{\textcolor{DeepKleinBlue}{$\bm{0.0290}$}}
    & \heatcell{22}{\textcolor{DeepKleinBlue}{$\bm{0.0841}$}}
    & \heatcell{22}{\textcolor{DeepKleinBlue}{$\bm{0.0811}$}} \\

    \midrule
    \textbf{Improvement}
    & \textcolor{deepMagenta}{\textbf{\(\uparrow\)44.3\%}}
    & \textbf{\(\uparrow\)59.8\%}
    & \textbf{\(\uparrow\)47.9\%}
    & \textbf{\(\uparrow\)40.0\%}
    & \textbf{\(\uparrow\)50.7\%}
    & \textbf{\(\uparrow\)1.7\%}
    & \textbf{\(\uparrow\)45.4\%}
    & \textbf{\(\uparrow\)45.0\%} \\
    \bottomrule
    \end{tabular}%
    }
  \end{table}

\begin{table}[h]
    \centering
    \caption{
    HumanCyc metric reconstruction results for geometric baselines.
    }
    \label{tab:humancyc_constant_curvature}

    \resizebox{0.95\linewidth}{!}{%
    \begin{tabular}{lrrrrrrrr}
    \toprule
    Space
    & \textcolor{DeepKleinBlue}{\textbf{Distortion$_{\mathrm{avg.}}$}} $\downarrow$
    & q50 $\downarrow$
    & q90 $\downarrow$
    & q95 $\downarrow$
    & High-related $\downarrow$
    & Low--Low $\downarrow$
    & Worst-group $\downarrow$
    & MAE $\downarrow$ \\
    \midrule

    \rowcolor{black!2}
    $\textcolor{lightMagenta}{\mathbb{S}^{72}}\!\times\!\textcolor{KleinBlue}{\mathbb{H}^{72}}$
    & \heatcell{12}{$0.0649 {\scriptstyle \pm 0.0001}$}
    & \heatcell{12}{$0.0450$}
    & \heatcell{12}{$0.1272$}
    & \heatcell{12}{$0.1771$}
    & \heatcell{12}{$0.0689$}
    & \heatcell{10}{$0.0569$}
    & \heatcell{12}{$0.0841$}
    & \heatcell{12}{$0.2379$} \\

    \rowcolor{black!2}
    $\textcolor{lightMagenta}{\mathbb{S}^{72}}\!\times\!\mathbb{E}^{72}$
    & \heatcell{7}{$0.0682 {\scriptstyle \pm 0.0000}$}
    & \heatcell{7}{$0.0480$}
    & \heatcell{7}{$0.1308$}
    & \heatcell{7}{$0.1853$}
    & \heatcell{9}{$0.0701$}
    & \heatcell{5}{$0.0671$}
    & \heatcell{11}{$0.0842$}
    & \heatcell{7}{$0.2514$} \\

    \rowcolor{black!2}
    $\textcolor{KleinBlue}{\mathbb{H}^{72}}\!\times\!\mathbb{E}^{72}$
    & \heatcell{10}{$0.0655 {\scriptstyle \pm 0.0001}$}
    & \heatcell{10}{$0.0453$}
    & \heatcell{9}{$0.1298$}
    & \heatcell{10}{$0.1803$}
    & \heatcell{10}{$0.0697$}
    & \heatcell{12}{$0.0567$}
    & \heatcell{9}{$0.0852$}
    & \heatcell{10}{$0.2421$} \\

    \rowcolor{black!2}
    $\textcolor{lightMagenta}{\mathbb{S}^{48}_{\kappa_1}}
    \times
    \textcolor{KleinBlue}{\mathbb{H}^{48}_{\kappa_2}}
    \times
    \mathbb{E}^{48}$
    & \heatcell{9}{$0.0660 {\scriptstyle \pm 0.0000}$}
    & \heatcell{9}{$0.0464$}
    & \heatcell{10}{$0.1294$}
    & \heatcell{12}{$0.1771$}
    & \heatcell{9}{$0.0702$}
    & \heatcell{14}{$0.0559$}
    & \heatcell{10}{$0.0849$}
    & \heatcell{9}{$0.2433$} \\

    \rowcolor{black!2}
    $\textcolor{deepPurple}{\mathbf{Gr}}(11,24)$
    & \heatcell{6}{$0.0693 {\scriptstyle \pm 0.0001}$}
    & \heatcell{6}{$0.0483$}
    & \heatcell{6}{$0.1334$}
    & \heatcell{5}{$0.1892$}
    & \heatcell{10}{$0.0699$}
    & \heatcell{2}{$0.0746$}
    & \heatcell{14}{$0.0835$}
    & \heatcell{6}{$0.2557$} \\

    \rowcolor{black!2}
    $\textcolor{lightKleinBlue}{\mathbf{SPD}(12)}$
    & \heatcell{5}{$0.0698 {\scriptstyle \pm 0.0001}$}
    & \heatcell{5}{$0.0495$}
    & \heatcell{5}{$0.1347$}
    & \heatcell{5}{$0.1892$}
    & \heatcell{7}{$0.0719$}
    & \heatcell{4}{$0.0679$}
    & \heatcell{7}{$0.0859$}
    & \heatcell{5}{$0.2586$} \\

    \rowcolor{black!2}
    $\textcolor{DeepKleinBlue}{\mathbf{Siegel}(12)}$
    & \heatcell{19}{$\underline{0.0526 {\scriptstyle \pm 0.0001}}$}
    & \heatcell{19}{$\underline{0.0332}$}
    & \heatcell{19}{$\underline{0.1070}$}
    & \heatcell{19}{$\underline{0.1534}$}
    & \heatcell{19}{$\underline{0.0557}$}
    & \heatcell{19}{$\underline{0.0492}$}
    & \heatcell{19}{$\underline{0.0708}$}
    & \heatcell{19}{$\underline{0.1899}$} \\

    \rowcolor{black!2}
    \rainbowmath{\mathbb{SL}(12)}
    & \heatcell{22}{
      \textcolor{DeepKleinBlue}{
        $\bm{0.0313} {\scriptstyle \bm{\pm 0.0003}}$
      }}
    & \heatcell{22}{
      \textcolor{DeepKleinBlue}{$\bm{0.0153}$}}
    & \heatcell{22}{
      \textcolor{DeepKleinBlue}{$\bm{0.0678}$}}
    & \heatcell{22}{
      \textcolor{DeepKleinBlue}{$\bm{0.1019}$}}
    & \heatcell{22}{
      \textcolor{DeepKleinBlue}{$\bm{0.0329}$}}
    & \heatcell{22}{
      \textcolor{DeepKleinBlue}{$\bm{0.0309}$}}
    & \heatcell{22}{
      \textcolor{DeepKleinBlue}{$\bm{0.0360}$}}
    & \heatcell{22}{
      \textcolor{DeepKleinBlue}{$\bm{0.1098}$}} \\

    \midrule
    \textbf{Improvement}
    & \textcolor{deepMagenta}{\textbf{\(\uparrow\)40.5\%}}
    & \textbf{\(\uparrow\)53.9\%}
    & \textbf{\(\uparrow\)36.7\%}
    & \textbf{\(\uparrow\)33.6\%}
    & \textbf{\(\uparrow\)40.9\%}
    & \textbf{\(\uparrow\)37.2\%}
    & \textbf{\(\uparrow\)49.1\%}
    & \textbf{\(\uparrow\)42.2\%} \\
    \bottomrule
    \end{tabular}%
    }

    \vspace{-7pt}
  \end{table}
\paragraph{Large scale link prediction on OGBL-PPA.}
We further test whether the geometric representation advantage transfers to a
downstream task on OGBL-PPA, a substantially larger protein association graph.
All methods also use the same experimental settings, while varying only the latent representation
space. This controlled setting isolates how well each geometry organizes nodes
for recovering unseen links.
Table~\ref{tab:ogbl_ppa_link_prediction} evaluates complementary aspects of
ranking quality. Hits@$K$ measures how often positive edges appear among the
top ranked candidates, MRR summarizes reciprocal rank, and Rank50, Rank90, and
Rank95 characterize the typical and tail ranks of positive edges. AUC and AP
further measure global discrimination between positive and negative pairs.
$\mathbb{SL}(8)$ improves all reported metrics, with particularly consistent
gains across both Hits and rank quantiles. This indicates that its advantage is
not limited to a particular ranking threshold, but extends across the ranking
distribution and positive negative separation. Together with metric
reconstruction, these results suggest that the geometry learned by
$\mathbb{SL}(n)$ supports both faithful graph representation and downstream
relational prediction at substantially larger scale.

\begin{table}[h]
  \centering
  \caption{
  OGBL-PPA link prediction results for geometric latent spaces.
  }
  \label{tab:ogbl_ppa_link_prediction}

  \resizebox{0.99\linewidth}{!}{%
  \begin{tabular}{lrrrrrrrrr}
  \toprule
  Space
  & \textcolor{DeepKleinBlue}{\textbf{Hits@20}} $\uparrow$
  & \textcolor{DeepKleinBlue}{\textbf{Hits@50}} $\uparrow$
  & \textcolor{DeepKleinBlue}{\textbf{Hits@100}} $\uparrow$
  & Rank50 $\downarrow$
  & Rank90 $\downarrow$
  & Rank95 $\downarrow$
  & MRR $\uparrow$
  & AUC $\uparrow$
  & AP $\uparrow$ \\
  \midrule

  \rowcolor{black!2}
  $\textcolor{lightMagenta}{\mathbb{S}^{32}}
   \!\times\!
   \textcolor{KleinBlue}{\mathbb{H}^{32}}$
  & \heatcell{8}{$0.0888 {\scriptstyle \pm 0.0050}$}
  & \heatcell{8}{$\underline{0.1558 {\scriptstyle \pm 0.0143}}$}
  & \heatcell{10}{$\underline{0.2332 {\scriptstyle \pm 0.0144}}$}
  & \heatcell{15}{$\underline{803}$}
  & \heatcell{18}{$\underline{30060}$}
  & \heatcell{19}{$\underline{90715}$}
  & \heatcell{9}{$0.0157$}
  & \heatcell{19}{$\underline{0.9915}$}
  & \heatcell{18}{$\underline{0.9926}$} \\

  \rowcolor{black!2}
  $\textcolor{lightMagenta}{\mathbb{S}^{32}}
   \!\times\!
   \mathbb{E}^{32}$
  & \heatcell{2}{$0.0725 {\scriptstyle \pm 0.0129}$}
  & \heatcell{5}{$0.1454 {\scriptstyle \pm 0.0127}$}
  & \heatcell{7}{$0.2102 {\scriptstyle \pm 0.0096}$}
  & \heatcell{13}{$931$}
  & \heatcell{17}{$37757$}
  & \heatcell{18}{$111444$}
  & \heatcell{4}{$0.0140$}
  & \heatcell{17}{$0.9903$}
  & \heatcell{17}{$0.9915$} \\

  \rowcolor{black!2}
  $\textcolor{KleinBlue}{\mathbb{H}^{32}}
   \!\times\!
   \mathbb{E}^{32}$
  & \heatcell{6}{$0.0841 {\scriptstyle \pm 0.0036}$}
  & \heatcell{2}{$0.1303 {\scriptstyle \pm 0.0022}$}
  & \heatcell{3}{$0.1893 {\scriptstyle \pm 0.0089}$}
  & \heatcell{7}{$1248$}
  & \heatcell{7}{$90588$}
  & \heatcell{9}{$279414$}
  & \heatcell{4}{$0.0140$}
  & \heatcell{8}{$0.9811$}
  & \heatcell{7}{$0.9844$} \\

  \rowcolor{black!2}
  $\textcolor{KleinBlue}{\mathbb{H}^{21}}
   \!\times\!
   \mathbb{E}^{21}
   \!\times\!
   \textcolor{lightMagenta}{\mathbb{S}^{21}}$
  & \heatcell{8}{$\underline{0.0898 {\scriptstyle \pm 0.0171}}$}
  & \heatcell{8}{$0.1556 {\scriptstyle \pm 0.0094}$}
  & \heatcell{8}{$0.2185 {\scriptstyle \pm 0.0101}$}
  & \heatcell{13}{$920$}
  & \heatcell{17}{$40199$}
  & \heatcell{17}{$118889$}
  & \heatcell{9}{$0.0155$}
  & \heatcell{17}{$0.9899$}
  & \heatcell{16}{$0.9912$} \\

  \rowcolor{black!2}
  $\textcolor{KleinBlue}{\mathbb{H}^{21}_{\kappa_1}}
   \!\times\!
   \mathbb{E}^{21}
   \!\times\!
   \textcolor{lightMagenta}{\mathbb{S}^{21}_{\kappa_2}}$
  & \heatcell{5}{$0.0809 {\scriptstyle \pm 0.0094}$}
  & \heatcell{6}{$0.1494 {\scriptstyle \pm 0.0118}$}
  & \heatcell{8}{$0.2217 {\scriptstyle \pm 0.0090}$}
  & \heatcell{12}{$990$}
  & \heatcell{15}{$48604$}
  & \heatcell{16}{$144410$}
  & \heatcell{6}{$0.0145$}
  & \heatcell{15}{$0.9883$}
  & \heatcell{15}{$0.9899$} \\

  \rowcolor{black!2}
  $\textcolor{deepPurple}{\mathbf{Gr}}(7,16)$
  & \heatcell{4}{$0.0775 {\scriptstyle \pm 0.0097}$}
  & \heatcell{3}{$0.1336 {\scriptstyle \pm 0.0115}$}
  & \heatcell{2}{$0.1808 {\scriptstyle \pm 0.0054}$}
  & \heatcell{2}{$1588$}
  & \heatcell{2}{$119504$}
  & \heatcell{2}{$397421$}
  & \heatcell{10}{$\underline{0.0157}$}
  & \heatcell{2}{$0.9748$}
  & \heatcell{2}{$0.9802$} \\

  \rowcolor{black!2}
  $\textcolor{lightKleinBlue}{\mathbf{SPD}(8)}$
  & \heatcell{4}{$0.0784 {\scriptstyle \pm 0.0153}$}
  & \heatcell{4}{$0.1386 {\scriptstyle \pm 0.0122}$}
  & \heatcell{4}{$0.1964 {\scriptstyle \pm 0.0086}$}
  & \heatcell{6}{$1370$}
  & \heatcell{14}{$52211$}
  & \heatcell{16}{$147717$}
  & \heatcell{8}{$0.0151$}
  & \heatcell{15}{$0.9881$}
  & \heatcell{14}{$0.9896$} \\

  \rowcolor{black!2}
  $\textcolor{DeepKleinBlue}{\mathbf{Siegel}(8)}$
  & \heatcell{4}{$0.0769 {\scriptstyle \pm 0.0295}$}
  & \heatcell{6}{$0.1462 {\scriptstyle \pm 0.0140}$}
  & \heatcell{8}{$0.2183 {\scriptstyle \pm 0.0095}$}
  & \heatcell{12}{$998$}
  & \heatcell{12}{$64364$}
  & \heatcell{13}{$198532$}
  & \heatcell{2}{$0.0134$}
  & \heatcell{13}{$0.9855$}
  & \heatcell{12}{$0.9877$} \\

  \rowcolor{black!2}
  \rainbowmath{\mathbb{SL}(8)}
  & \heatcell{22}{
    \textcolor{DeepKleinBlue}{
    $\bm{0.1282} {\scriptstyle \bm{\pm 0.0056}}$}}
  & \heatcell{22}{
    \textcolor{DeepKleinBlue}{
    $\bm{0.2188} {\scriptstyle \bm{\pm 0.0055}}$}}
  & \heatcell{22}{
    \textcolor{DeepKleinBlue}{
    $\bm{0.3093} {\scriptstyle \bm{\pm 0.0119}}$}}
  & \heatcell{22}{\textcolor{DeepKleinBlue}{$\bm{346}$}}
  & \heatcell{22}{\textcolor{DeepKleinBlue}{$\bm{10700}$}}
  & \heatcell{22}{\textcolor{DeepKleinBlue}{$\bm{36145}$}}
  & \heatcell{22}{\textcolor{DeepKleinBlue}{$\bm{0.0195}$}}
  & \heatcell{22}{\textcolor{DeepKleinBlue}{$\bm{0.9950}$}}
  & \heatcell{22}{\textcolor{DeepKleinBlue}{$\bm{0.9957}$}} \\

  \midrule
\textbf{Improvement}
& \textcolor{deepMagenta}{\textbf{\(\uparrow\)42.8\%}}
& \textcolor{deepMagenta}{\textbf{\(\uparrow\)40.4\%}}
& \textcolor{deepMagenta}{\textbf{\(\uparrow\)32.6\%}}
& \textbf{\(\uparrow\)57.0\%}
& \textbf{\(\uparrow\)64.4\%}
& \textbf{\(\uparrow\)60.2\%}
& \textbf{\(\uparrow\)24.1\%}
& \textbf{\(\uparrow\)0.3\%}
& \textbf{\(\uparrow\)0.3\%} \\

  \bottomrule
  \end{tabular}%
  }
  \end{table}

\subsection{Deep Order-Aware Composition}
\label{sec:flickr}

\begin{center}
\begin{minipage}{0.9\linewidth}
  \centering

  \begin{minipage}[c]{0.32\linewidth}
    \centering
    \includegraphics[width=\linewidth]{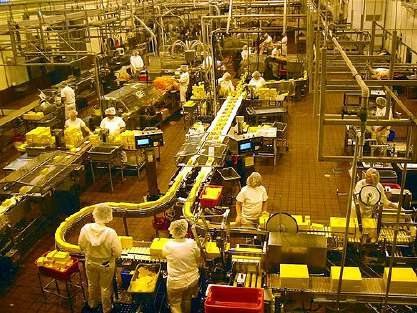}
  \end{minipage}
  \hfill
  \begin{minipage}[c]{0.64\linewidth}
    \scriptsize

    \textcolor{red}{\textbf{Positive 0:}}
    a \textcolor{red}{[dozen workers]} wearing uniforms and sanitation hats are
    \textcolor{red}{[working on an assembly line]} in
    \textcolor{red}{[a factory]}

    \vspace{0.35em}
    \textbf{Negative 1:}
    a \textcolor{KleinBlue}{[dozen workers]} wearing uniforms and sanitation hats are working on an
    \textcolor{KleinBlue}{[line assembly]} in
    \textcolor{KleinBlue}{[a factory]}

    \vspace{0.35em}
    \textbf{Negative 2:}
    a \textcolor{KleinBlue}{[dozen workers]} wearing uniforms and sanitation hats are
    \textcolor{KleinBlue}{[assembly on an working line]} in
    \textcolor{KleinBlue}{[a factory]}

    \vspace{0.35em}
    \textbf{Negative 3:}
    a \textcolor{KleinBlue}{[are workers]} wearing uniforms and sanitation hats
    \textcolor{KleinBlue}{[dozen working]} on an assembly line in
    \textcolor{KleinBlue}{[a factory]}

    \vspace{0.35em}
    \textbf{Negative 4:}
    a \textcolor{KleinBlue}{[dozen workers]} wearing uniforms and sanitation hats are
    \textcolor{KleinBlue}{[working on an assembly line]} in
    \textcolor{KleinBlue}{[factory a]}
  \end{minipage}

  \captionof{figure}{
  Flickr30k-Order: The positive caption preserves the original word order, while the negative captions perturb local phrase order with nearly the same bag of words.
  }
  \label{fig:flickr30k_order_example}
\end{minipage}
\vspace{-2pt}
\end{center}
\paragraph{Order sensitive composition on Flickr30k-Order.}
We finally isolate the compositional property of $\mathbb{SL}(n)$ on Flickr30k-Order,
where performance depends on preserving semantic order. Order Accuracy and
Order Margin measure order discrimination, while Hard Accuracy and MRR evaluate
harder and ranking based cases. For a controlled comparison, we freeze the
same CLIP backbone and train only a lightweight group specific head for each
representation space.
We compare additive models with noncommutative groups of increasing effective
order depth $D_{\rm ord}$. Heisenberg and the unitriangular group
$\mathrm{UT}(n)$ have finite depth, whereas $\mathbb{SL}(4)$ has
$D_{\rm ord}=\infty$. Performance generally improves with larger
$D_{\rm ord}$, and ordered $\mathbb{SL}(4)$ performs best across all metrics,
while removing ordered composition causes a large drop. This supports deep
noncommutative composition as an advantage of $\mathbb{SL}(n)$ beyond its
latent geometry.
\begin{table}[h]
\vspace{-4pt}
\centering
\caption{
Order sensitive composition on ARO Flickr30k-Order. 
}
\label{tab:aro_flickr30k_order}

 \resizebox{0.99\linewidth}{!}{%
  \begin{tabular}{lccc|cccc}
  \toprule
  Model
  & Dim.
  & $D_{\rm ord}$
  & Composition
  & \textcolor{DeepKleinBlue}{\textbf{OrderAcc.}} $\uparrow$
  & OrderMargin $\uparrow$
  & HardAcc. $\uparrow$
  & MRR $\uparrow$ \\
  \midrule

  \rowcolor{black!2}
  BoW
  & -- & $0$
  & Additive
  & \heatcell{0}{$47.82 {\scriptstyle \pm 1.96}$}
  & \heatcell{0}{$-0.0000$}
  & \heatcell{0}{$41.52$}
  & \heatcell{0}{$0.561$} \\

  \rowcolor{black!2}
  CLIP
  & -- & --
  & Implicit
  & \heatcell{2}{$86.08 {\scriptstyle \pm 0.00}$}
  & \heatcell{0}{$0.0176$}
  & \heatcell{1}{$66.07$}
  & \heatcell{2}{$0.805$} \\

  \midrule

  \rowcolor{black!2}
  \rainbowmath{\mathbb{SL}(4)} w/o Ordered Comp.
  & $15$ & $0$
  & Commutative addition
  & \heatcell{0}{$17.97 {\scriptstyle \pm 3.14}$}
  & \heatcell{0}{$-0.0079$}
  & \heatcell{0}{$3.54$}
  & \heatcell{0}{$0.272$} \\

  \rowcolor{black!2}
  Heisenberg-$H^{7}$
  & $15$ & $1$
  & Group matrix multiplication
  & \heatcell{13}{$94.28 {\scriptstyle \pm 0.02}$}
  & \heatcell{10}{$0.6324$}
  & \heatcell{7}{$82.32$}
  & \heatcell{11}{$0.906$} \\

  \rowcolor{black!2}
  $\mathrm{UT}(4)$
  & $6$ & $2$
  & Group matrix multiplication
  & \heatcell{13}{$94.11 {\scriptstyle \pm 0.27}$}
  & \heatcell{9}{$0.5804$}
  & \heatcell{6}{$81.64$}
  & \heatcell{10}{$0.903$} \\

  \rowcolor{black!2}
  $\mathrm{UT}(6)$
  & $15$ & $4$
  & Group matrix multiplication
  & \heatcell{18}{$96.30 {\scriptstyle \pm 0.18}$}
  & \heatcell{15}{$0.8862$}
  & \heatcell{13}{$89.11$}
  & \heatcell{17}{$0.942$} \\

    \rowcolor{black!2}
  $\mathrm{UT}(15)$
  & $105$ & $13^{*}$
  & Group matrix multiplication
  & \heatcell{22}{$\underline{97.25 {\scriptstyle \pm 0.06}}$}
  & \heatcell{23}{$\underline{1.1683}$}
  & \heatcell{21}{$\underline{92.13}$}
  & \heatcell{22}{$\underline{0.957}$} \\

  \rowcolor{black!2}
  \textbf{Full }\rainbowmath{\mathbb{SL}(4)}
  w/ Ordered Comp.
  & $15$ & $\infty$
  & \textbf{Group matrix multiplication}
  & \heatcell{25}{
    \textcolor{DeepKleinBlue}{
      $\bm{97.46} {\scriptstyle \bm{\pm 0.07}}$
    }}
  & \heatcell{25}{
    \textcolor{DeepKleinBlue}{$\bm{1.1934}$}}
  & \heatcell{25}{
    \textcolor{DeepKleinBlue}{$\bm{92.87}$}}
  & \heatcell{25}{
    \textcolor{DeepKleinBlue}{$\bm{0.962}$}} \\

  \bottomrule
  \end{tabular}%
  }
  \vspace{0.5mm}
  \begin{minipage}{0.99\linewidth}
  \scriptsize
  $^{*}$For dimension $15=\dim\mathbb{SL}(4)$, the maximal finite
$D_{\mathrm{ord}}$ of a nilpotent Lie algebra is $13$.
  \end{minipage}

  \end{table}

\vspace{-15pt}
\subsection{Ablation Study and Sensitivity test}
\vspace{-8pt}
\begin{figure}[h]
    \centering
    \begin{subfigure}[t]{0.4\linewidth}
        \centering
        \includegraphics[width=\linewidth]
        {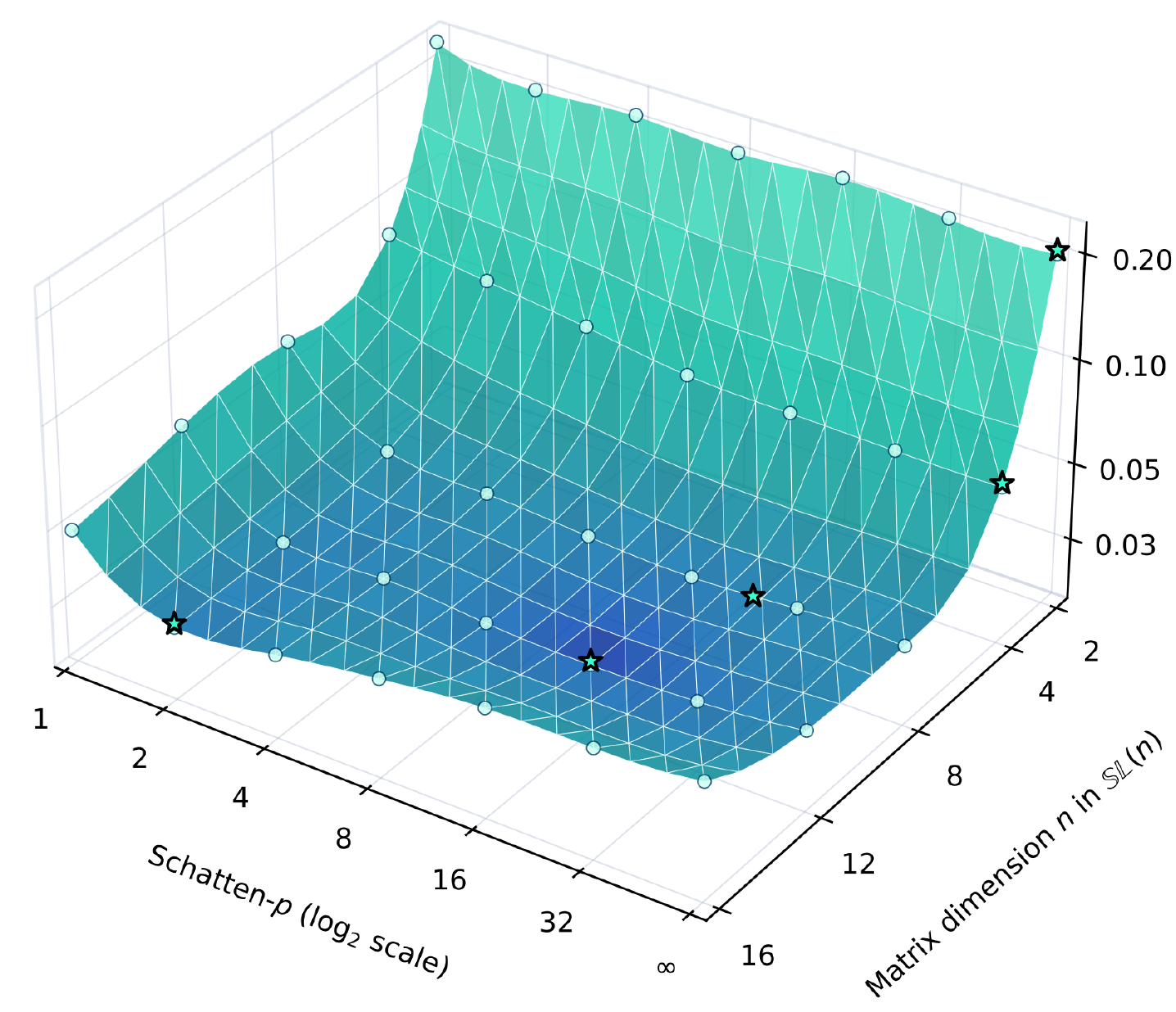}
        \caption{
           Schatten-$p$ sensitivity across $\mathbb{SL}(n)$ dimensions.
        }
        \label{fig:kegg_sln_p_sensitivity}
    \end{subfigure}
    \hfill
    \begin{subfigure}[t]{0.5\linewidth}
        \centering
        \includegraphics[width=\linewidth]
        {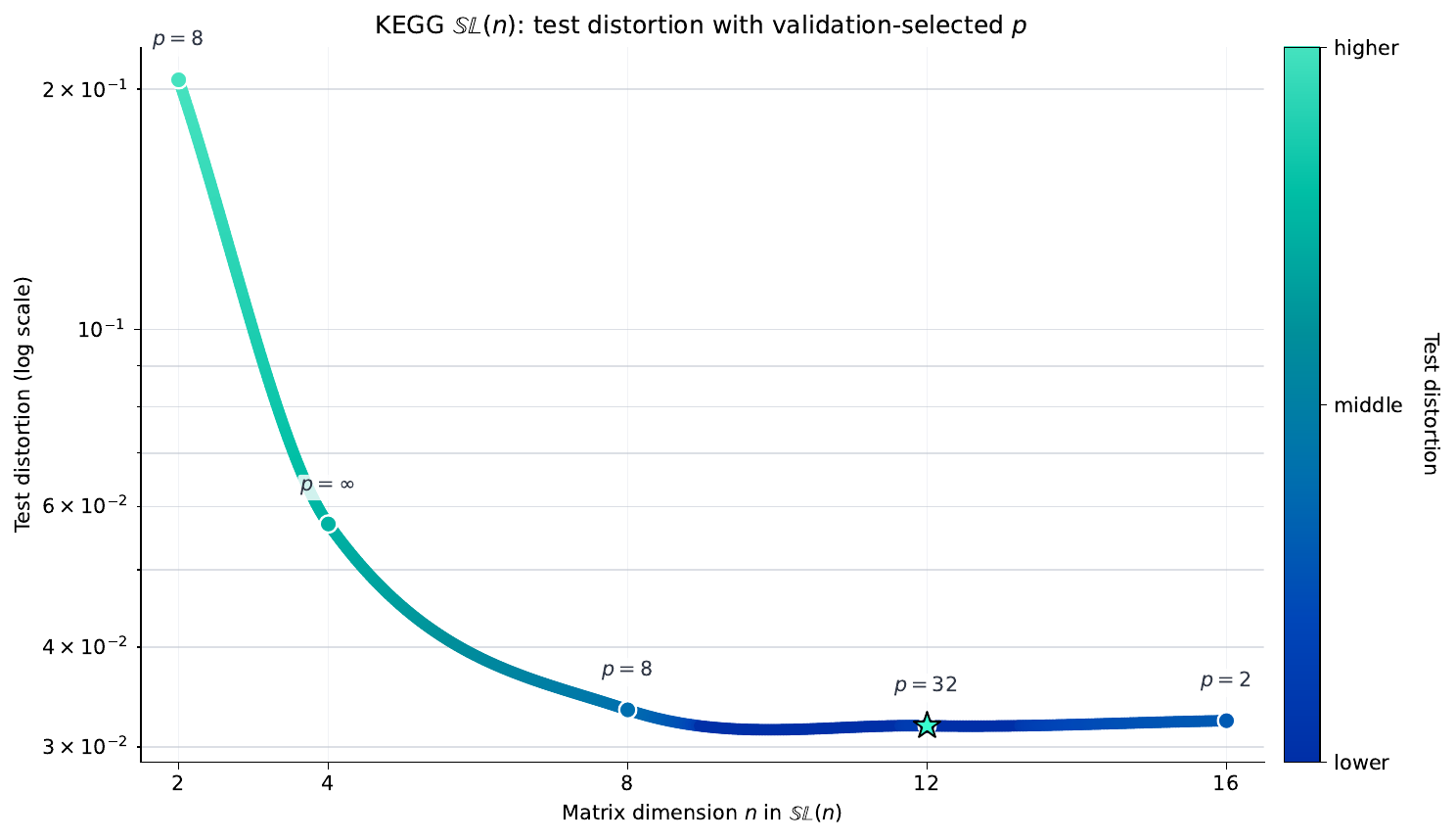}
        \caption{
        Dimension ablation with validation-optimal Schatten-$p$.
        }
        \label{fig:kegg_sln_dimension_ablation}
    \end{subfigure}

    \caption{
        Sensitivity and dimension ablation studies
        for $\mathbb{SL}(n)$ on KEGG.
    }
    \label{fig:kegg_sln_sensitivity_dimension}
\end{figure}
\vspace{-4pt}
\begin{wraptable}{r}{0.50\columnwidth}
    \vspace{-1.1em}
    \centering
    \caption{Representative training loss and test distortion on KEGG.}
    \label{tab:kegg_train_test_gap}
    \vspace{-0.4em}
    \scriptsize
    \setlength{\tabcolsep}{3.0pt}
    \renewcommand{\arraystretch}{1.08}
    \begin{tabular}{c cc cc}
        \toprule
        & \multicolumn{2}{c}{$n=12$}
        & \multicolumn{2}{c}{$n=16$} \\
        \cmidrule(lr){2-3}
        \cmidrule(lr){4-5}
        $p$
        & Train loss
        & Test dist.
        & Train loss
        & Test dist. \\
        \midrule
        $2$
        & $8.05{\times}10^{-7}$
        & $0.03279$
        & $1.89{\times}10^{-7}$
        & $0.03149$ \\
        $16$
        & $1.25{\times}10^{-5}$
        & $0.03026$
        & $1.15{\times}10^{-7}$
        & $0.04040$ \\
        \bottomrule
    \end{tabular}
    \vspace{-1.0em}
\end{wraptable}
Figure~\ref{fig:kegg_sln_sensitivity_dimension} shows a clear interaction
between matrix dimension $n$ and Schatten order $p$. Increasing $n$ enlarges
representation capacity, whereas larger $p$ places greater emphasis on dominant
singular directions. The preferred $p$ varies substantially with $n$, indicating
that these two hyperparameters control different aspects of the geometry,
namely representation capacity and directional sensitivity, and should be tuned
jointly. Test distortion improves rapidly up to $n=8$--$12$ and then saturates.
Table~\ref{tab:kegg_train_test_gap} further shows that at $p=16$, increasing
$n$ from $12$ to $16$ reduces training loss by about $109\times$ while
worsening test distortion by $33.5\%$, showing that additional capacity can
reduce training error without improving test performance.
\vspace{-4pt}
\section{Conclusion}
\vspace{-4pt}
$\mathbb{SL}(n)$ provides a single representation space combining coupled
mixed-curvature with deep order-aware composition. Remarkably, this richness
emerges from a minimal construction consisting only of the $\det(A)=1$
constraint and a simple left invariant Schatten-$p$ tangent norm. Its Finsler
geometry realizes curvature signs $\{-,0,+\}$ with asymptotically maximal
mixed-curvature and curvature-coupling capacities, while its non-nilpotent
Lie algebra structure supports noncommutative interactions at arbitrary depth. Thus,
rich geometric coexistence, intrinsic interaction, and deep composition need
not rely on separate representation components. The consistent gains across
metric reconstruction, large scale link prediction, and ordered composition
show that this structural simplicity preserves expressive power. Together,
these results establish $\mathbb{SL}(n)$ as a general structured latent space
for intrinsically coupled mixed-curvature geometry and deep composition,
illustrating how simple structural constraints can yield unexpectedly rich
representations, with potential applications across geometric, relational,
sequential, multimodal, and scientific representation learning.

\newpage

\subsubsection*{Acknowledgments}
This work was partially supported by JST Moonshot R\&D Grant Number JPMJPS2011
and by the New Energy and Industrial Technology Development Organization (NEDO)
under project JPNP26007. Yusuke Mukuta was supported by JSPS KAKENHI Grant
Number JP25K21268. Xingrun Li was supported by JST SPRING, Grant Number
JPMJSP2108.

\bibliography{iclr2025_conference}
\bibliographystyle{iclr2025_conference}
\newpage
\appendix

\section*{\LARGE Appendix}

\section{Related Work and Discussion}
\label{app:related_work}

\subsection{Related Work}

\paragraph{Mixed-curvature and manifold representation learning.}
Non-Euclidean representation learning uses the geometry of the latent manifold
as an inductive bias for structured data. Constant-curvature manifolds provide
three canonical geometric regimes: Euclidean spaces model approximately flat
structures, hyperbolic spaces naturally accommodate hierarchical and tree-like
structures, while spherical spaces provide compact positively curved geometry
for structures with cyclic or globally constrained relations
\citep{nickel2017poincare,chami2019hyperbolic,bachmann2020constant}.
These geometries have been applied broadly to knowledge graphs, recommendation systems, generative
modeling, graph learning, and visual representation
\citep{chami2020low,skopek2020mixed,yang2025hgformer}.
Since real data often contain structures that cannot be captured by one
curvature regime, product manifolds combine Euclidean, hyperbolic, and
spherical components to form mixed-curvature representation spaces
\citep{gu2019mixed,skopek2020mixed}. Such spaces have been further developed
for graph learning and biological networks
\citep{sun2022self,mcneela2024product}. Beyond constant-curvature factors,
matrix and higher rank manifolds including SPD, Grassmann, and Siegel spaces
provide richer intrinsic geometries for covariance, subspace, and graph
representations
\citep{huang2018grassmann,lopez2021symmetric}.
A complementary line of work adapts geometry through the learning architecture,
for example by learning curvature parameters, weighting multiple manifold
components, or dynamically selecting and combining geometric experts
\citep{bachmann2020constant,sun2022self,nguyen2023weighted,guo2025graphmore}.
Our work instead focuses on the underlying representation manifold.
Compared with existing geometric spaces, $\mathbb{SL}(n)$ provides intrinsic
mixed curvature and curvature coupling within a single manifold, while its
Lie group structure additionally supports noncommutative composition.
Because these properties arise from the representation space itself rather
than from a specialized geometry-learning architecture, $\mathbb{SL}(n)$ can
serve as a general latent space beyond curvature-adaptive models. Accordingly,
our controlled experiments vary the representation manifold while keeping the
surrounding learning framework fixed.

\paragraph{Finsler geometry and Lie structures in machine learning.}
Finsler geometry extends Riemannian geometry by allowing the tangent norm to
depend on direction, and has been used in representation learning to enrich the
geometry available beyond standard Riemannian metrics. In particular, Finsler
metrics on symmetric spaces have been studied for graph embeddings
\citep{lopez2021symmetric}, while more recent work has explored Finsler
geometry for asymmetric embedding and graph learning
\citep{dages2025finsler,roddenberry2026finsler}. Lie groups and Lie algebras
have been widely used in machine learning for geometric representation,
continuous symmetries, and equivariant architectures
\citep{huang2017lie,finzi2020generalizing,hutchinson2021lietransformer,
dehmamy2021automatic}. More recent work has extended this direction to
noncompact and semisimple groups, including architectures involving
$\mathrm{SL}(2,\mathbb R)$, $\mathrm{SL}(n,\mathbb R)$, and their Lie
algebras
\citep{lawrence2024sl2,mironenco2024lie,lin2024lie,kim2026equivariant}.
Many of these approaches exploit Lie groups and Lie algebras as symmetry
groups, transformation domains, or algebraic structures for constructing
equivariant networks. In contrast, we do not use $\mathrm{SL}(n)$ only as a
symmetry acting on external features. We make $\mathbb{SL}(n)$ itself the
latent representation space, where the left invariant Schatten-$p$ Finsler
structure determines intrinsic mixed curvature and curvature coupling, while
the group product and Lie algebra provide noncommutative and higher order
composition. Thus, our use of Lie structure extends beyond symmetry and
equivariance, with geometry and composition jointly defining the underlying
representation space.
\newpage

\subsection{Discussion and Limitations}
\label{app:limitations}

\subsubsection{Discussion about applications}

A particularly appealing aspect of $\mathbb{SL}(n)$ is that a remarkably
simple structural constraint gives rise to a surprisingly rich representation
space. Defined only by the determinant-one constraint and equipped with a
left-invariant Schatten-$p$ geometry, $\mathbb{SL}(n)$ simultaneously exhibits
intrinsic mixed curvature, strong curvature coupling, noncommutative group
composition, and arbitrarily deep nested Lie-bracket interactions. These
properties arise from a single underlying structure rather than from separately
designed geometric or compositional components. Also, $\mathbb{SL}(n)$
can be used as a plug-and-play embedding space. Existing encoders can simply
output matrix-valued latent representations in $\mathbb{SL}(n)$, allowing rich
geometry and composition to be introduced without redesigning the surrounding
architecture.

One important direction is \textbf{relational and sequential learning}. In
knowledge graphs, relations are naturally compositional and often asymmetric,
while the underlying relational structure may simultaneously contain
hierarchical, cyclic, and densely interconnected patterns. An
$\mathbb{SL}(n)$ embedding can therefore use its mixed-curvature geometry to
capture heterogeneous structural organization, while its noncommutative group
operation provides a native mechanism for ordered relation composition and
multi hop reasoning. Similar considerations arise in recommender systems and
behavioral modeling, where the meaning of an interaction sequence often depends
not only on which events occur, but also on their order and accumulated effect.
The ability of $\mathbb{SL}(n)$ to support deep order-sensitive composition may
therefore provide a useful inductive bias for long-range relational and
sequential dependencies.

Another promising direction is \textbf{multimodal, temporal, and dynamical
representation learning}. Different modalities may exhibit distinct local
geometries while still requiring structured alignment within a shared latent
space, while temporal observations can often be viewed as transformations
accumulated over time. Because $\mathbb{SL}(n)$ acts primarily as an embedding
space rather than a task-specific architecture, it can be incorporated with
minimal modification into existing GNNs, Transformers, state space models, and
multimodal encoders by replacing or augmenting their latent representation
layer. This plug-and-play nature makes it possible to introduce intrinsic
geometric heterogeneity and structured composition into a broad range of
architectures without designing separate mechanisms for each application.

\subsubsection{Limitations and Future Directions}

Our experiments are designed to isolate the effect of the underlying
representation space and therefore do not explore architectures specifically
optimized for $\mathbb{SL}(n)$. The results show that the space works
effectively within controlled frameworks, while dedicated $\mathbb{SL}(n)$
layers may better exploit its geometric and compositional structure. Another
limitation concerns distance computation. The closed form semidistance used in
our scalable experiments is restricted to the principal logarithm domain,
whereas exact intrinsic path distances are more expensive. We partially
address this through path approximations, while globally robust distance
constructions remain an important direction.

Matrix valued representations also incur higher computational cost than vector
embeddings due to matrix multiplication, matrix logarithms, and Schatten norm
evaluations. This is a general challenge for matrix manifold methods rather
than one specific to $\mathbb{SL}(n)$, while group multiplication and inversion
remain standard matrix operations. Future work may reduce this cost through low rank Lie algebra updates,
approximate logarithms, and efficient structured parameterizations.

Integrating $\mathbb{SL}(n)$ with larger architectures is a natural next step.
Dedicated $\mathbb{SL}(n)$ layers may allow models to exploit mixed curvature,
intrinsic coupling, and noncommutative composition jointly rather than using
the space only as an embedding domain. Extending this perspective to other
matrix Lie groups may reveal how geometry and algebra match structural priors.
Promising applications include foundation models, knowledge graphs,
multimodal and sequential learning, and scientific representation learning.
\newpage

\section{Distance and Optimization  Analysis}
\label{app:kegg,h}
\subsection{Schatten Semidistance}
\label{app:semidistance_geodesic}

\subsubsection{Relation between the Schatten Semidistance and Intrinsic Distance}

The Schatten semidistance $D_{\mathbb{SL}}$ is used for pairwise comparison
in $\mathbb{SL}_p(n)$, while the Finsler metric $F_p$ induces the intrinsic
geodesic distance $d_p$. We first establish their local relation.
\begin{lemma}[Second-order tightness of the Schatten semidistance]
\label{lem:semidistance_local}
For every $1<p<\infty$, there exist local constants $\delta,C>0$ such that,
for any $A,B\in\mathbb{SL}_p(n)$ in the principal-logarithm domain with
$D_{\mathbb{SL}}(A,B)<\delta$,
\begin{equation}
0
\leq
D_{\mathbb{SL}}(A,B)-d_p(A,B)
\leq
C\,D_{\mathbb{SL}}(A,B)^2.
\end{equation}
Consequently,
$D_{\mathbb{SL}}(A,B)=d_p(A,B)+O(D_{\mathbb{SL}}(A,B)^2)$, and equivalently
$D_{\mathbb{SL}}(A,B)=d_p(A,B)+O(d_p(A,B)^2)$ as $B\to A$.
\end{lemma}

\begin{proof}
Let $X=\log(A^{-1}B)$. In the principal-logarithm domain,
$\log(B^{-1}A)=-X$, so
$D_{\mathbb{SL}}(A,B)=\|X\|_{S_p}$.
The exponential path $\gamma(t)=A\exp(tX)$ has constant
left-trivialized velocity $X$ and length $\|X\|_{S_p}$. Hence
\[
d_p(A,B)\leq D_{\mathbb{SL}}(A,B).
\]

Conversely, let $\gamma$ be any sufficiently short piecewise-smooth
curve from $A$ to $B$, and write
$L=L_p(\gamma)=\int_0^1
\|\gamma(t)^{-1}\dot{\gamma}(t)\|_{S_p}\,dt$.
The Magnus expansion of the endpoint $A^{-1}B$ gives
\[
X
=
\int_0^1\gamma(t)^{-1}\dot{\gamma}(t)\,dt
+
\frac12\int_0^1\int_0^{t_1}
\big[
\gamma(t_2)^{-1}\dot{\gamma}(t_2),
\gamma(t_1)^{-1}\dot{\gamma}(t_1)
\big]\,dt_2\,dt_1
+ O(L^3) .
\]
The first term is linear in the path velocity, whereas every remaining
term contains at least two such factors. Since
$\|[U,V]\|_{S_p}\leq2\|U\|_{S_p}\|V\|_{S_p}$, the second term is
$O(L^2)$, and local convergence of the Magnus series gives, for
sufficiently small $L$,
\[
\|X\|_{S_p}\leq L+CL^2.
\]

Choose $\delta>0$ small enough that this estimate is valid whenever
$L<2\delta$. If $D_{\mathbb{SL}}(A,B)<\delta$, then the first inequality
gives $d_p(A,B)<\delta$. Therefore, a minimizing sequence of curves has
$L<2\delta$ for all sufficiently late terms. Applying the preceding
estimate and letting $L\to d_p(A,B)$ yields
\[
D_{\mathbb{SL}}(A,B)
\leq d_p(A,B)+C\,d_p(A,B)^2.
\]
Combining this with $d_p(A,B)\leq D_{\mathbb{SL}}(A,B)$ gives
\[
0\leq D_{\mathbb{SL}}(A,B)-d_p(A,B)
\leq C\,d_p(A,B)^2
\leq C\,D_{\mathbb{SL}}(A,B)^2.
\]
Thus
$D_{\mathbb{SL}}(A,B)
=d_p(A,B)+O(D_{\mathbb{SL}}(A,B)^2)$.
Since the two distances are locally equivalent, the remainder may
equivalently be written as $O(d_p(A,B)^2)$.
\end{proof}

\paragraph{Empirical correlation with numerical geodesic distance.}
We further examine whether this local agreement extends to representations
encountered in practice. On KEGG, we randomly sample $100$ learned
representation pairs and compute a high-accuracy numerical reference
$\widehat d_p$ by geodesic path optimization under $F_p$. We compare
$D_{\mathbb{SL}}$ with $\widehat d_p$ using their mean relative discrepancy,
Pearson correlation, and Spearman rank correlation. We additionally record
the computation time of the numerical reference. $100$ pairs are randomly sampled on KEGG. The last column reports the slowdown of numerical geodesic computation
relative to $D_{\mathbb{SL}}$.

\begin{table}[h]
\centering
\caption{
Correlation between the Schatten semidistance $D_{\mathbb{SL}}$ and
numerical geodesic distances.
}
\label{tab:semidistance_geodesic}
\small
\setlength{\tabcolsep}{5pt}
\resizebox{0.7\linewidth}{!}{%
\begin{tabular}{c c c c c c}
\toprule
$p$
& Rel. Diff. $\downarrow$
& Pearson $\uparrow$
& Spearman $\uparrow$
& Ref. Time / Pair
& Slowdown \\
\midrule
$2$
& $27.11\%$
& $0.9407$
& $0.9401$
& $473$--$488$ s
& $(1.95$--$2.01)\times10^{5}$ \\
$4$
& $18.58\%$
& $0.9553$
& $0.9526$
& $389$--$464$ s
& $(1.61$--$1.92)\times10^{5}$ \\
$8$
& $15.08\%$
& $0.9856$
& $0.9766$
& $478$--$516$ s
& $(1.97$--$2.13)\times10^{5}$ \\
$16$
& $13.61\%$
& $0.9886$
& $0.9777$
& $593$ s
& $2.45\times10^{5}$ \\
$32$
& $11.80\%$
& $0.9831$
& $0.9454$
& $380$--$565$ s
& $(1.57$--$2.33)\times10^{5}$ \\
\bottomrule
\end{tabular}
}
\end{table}

Across all $p$, $D_{\mathbb{SL}}$ remains strongly correlated with the
numerical geodesic reference, with Pearson correlations of
$0.94$--$0.99$ and Spearman correlations of $0.94$--$0.98$.
The mean relative discrepancy decreases from $27.11\%$ at $p=2$ to
$11.80\%$ at $p=32$, while the numerical geodesic computation requires
hundreds of seconds per pair and is approximately
$1.6\times10^5$--$2.5\times10^5$ times slower than
$D_{\mathbb{SL}}$. These results show that the closed-form Schatten
semidistance preserves both the magnitude and ranking structure of the
intrinsic geometry at a fraction of the computational cost.


\subsubsection{Geodesic Path Ablation and Logarithm Robustness}
\label{app:geodesic_path_ablation}

The Schatten semidistance $D_{\mathbb{SL}}$ admits a direct interpretation
under the same Schatten-$p$ length structure used to define the geometry.
For
\(
X=\log(A^{-1}B),
\)
the canonical exponential path
$\gamma(t)=A\exp(tX)$ satisfies
\begin{equation}
L_p(\gamma)
=
\int_0^1
\|\gamma(t)^{-1}\dot{\gamma}(t)\|_{S_p}\,dt
=
\|X\|_{S_p}
=
D_{\mathbb{SL}}(A,B),
\end{equation}
whenever the principal logarithm is well defined. Hence,
$D_{\mathbb{SL}}$ is exactly the $F_p$-length of a canonical admissible
path rather than an unrelated pairwise objective, and consequently
\(
d_p(A,B)\leq D_{\mathbb{SL}}(A,B).
\)
Lemma~\ref{lem:semidistance_local} further shows that this upper bound is
second-order tight locally.
To examine whether the empirical performance depends specifically on this
single exponential path, we introduce the $K$-segment piecewise-exponential
path approximation
\begin{equation}
\widehat d_p^{(K)}(A,B)
=
\inf_{\substack{
G_0=A,\;G_K=B\\
G_1,\ldots,G_{K-1}\in\mathbb{SL}_p(n)
}}
\sum_{k=0}^{K-1}
D_{\mathbb{SL}}(G_k,G_{k+1}).
\end{equation}
For $K=1$, this reduces exactly to the original Schatten semidistance,
\begin{equation}
\widehat d_p^{(1)}(A,B)=D_{\mathbb{SL}}(A,B),
\end{equation}
whereas increasing $K$ allows increasingly flexible piecewise-exponential
paths and therefore provides progressively tighter numerical
approximations to the intrinsic path distance.

We evaluate this effect on a connected $64$-node subgraph of KEGG using
$\mathbb{SL}_2(4)$ and three random seeds. We perform end-to-end training
with $K=1$ and $K=2$, while $K=4$ and $K=8$ are used for numerical path
refinement on frozen learned embeddings. All other model, optimization,
initialization, and data-split settings are held fixed between $K=1$ and
$K=2$ for each seed. In the following, ``Gap to $K{=}8$'' denotes the mean
relative discrepancy to the numerically stabilized $K=8$ reference
evaluated on the same learned embedding.

\begin{table}[h]
\centering
\caption{
End-to-end path-objective ablation on KEGG using
$\mathbb{SL}_2(4)$.
}
\label{tab:kegg_small_geodesic_ablation}
\small
\setlength{\tabcolsep}{7pt}
\begin{tabular}{c c c c}
\toprule
$K$
& Gap to $K{=}8$ $\downarrow$
& Test Distortion $\downarrow$
& Relative Runtime $\downarrow$ \\
\midrule
$1$
& $0.02146 \pm 0.00068$
& $0.17752 \pm 0.00293$
& $1.00 \pm 0.00$ \\
$2$
& $\mathbf{0.00502 \pm 0.00016}$
& $\mathbf{0.17671 \pm 0.00190}$
& $5.48 \pm 0.20$ \\
\bottomrule
\end{tabular}
\end{table}

As shown in Table~\ref{tab:kegg_small_geodesic_ablation}, replacing the
single exponential path by a two-segment path substantially tightens the
numerical path approximation. The discrepancy to the $K=8$ reference
decreases from $0.02146$ to $0.00502$, corresponding to a $76.6\%$
reduction. In contrast, the resulting representation performance changes
only marginally: test distortion improves from $0.17752$ to $0.17671$, a
relative improvement of approximately $0.46\%$, while training becomes
approximately $5.48\times$ more expensive. Thus, substantially refining
the pairwise path geometry produces only a minor change in the learned
representation quality, while incurring a considerably larger
computational cost. Results are reported as mean $\pm$ standard deviation. Runtime is normalized by the $K=1$ setting.

\paragraph{Path refinement on frozen embeddings.}
We further evaluate $K\in\{1,2,4,8\}$ on the six learned embeddings
obtained from the $K=1$ and $K=2$ training runs. This yields $2{,}424$
held-out pairwise observations and isolates the numerical effect of path
refinement from changes in the learned representation. The gap is measured relative to the $K=8$ reference for the same pair.

\begin{table}[h]
\centering
\caption{
Piecewise-path refinement on frozen KEGG embeddings.
}
\label{tab:kegg_small_path_refinement}
\small
\setlength{\tabcolsep}{7pt}
\begin{tabular}{c c c c}
\toprule
$K$
& Mean Gap $\downarrow$
& p95 Gap $\downarrow$
& Max Gap $\downarrow$ \\
\midrule
$1$
& $0.02182$
& $0.04626$
& $0.09072$ \\
$2$
& $0.00494$
& $0.01049$
& $0.01909$ \\
$4$
& $0.00098$
& $0.00207$
& $0.00373$ \\
$8$
& $0$
& $0$
& $0$ \\
\bottomrule
\end{tabular}
\end{table}

The refinement is fully consistent with the expected path hierarchy:
across all $2{,}424$ observations, we obtain
\begin{equation}
\widehat d_p^{(8)}
\leq
\widehat d_p^{(4)}
\leq
\widehat d_p^{(2)}
\leq
\widehat d_p^{(1)},
\end{equation}
with no observed violations. Moreover, the discrepancy to the $K=8$
reference decreases rapidly, from $0.02182$ at $K=1$ to $0.00494$ at
$K=2$ and $0.00098$ at $K=4$.

We additionally verify the numerical stability of the $K=8$ reference.
Increasing the optimization budget from $1{,}000$ to $2{,}000$ iterations
changes the final objective by only $1.82\times10^{-12}$ on average, and
independent deterministic restarts exhibit an average relative spread of
$1.83\times10^{-10}$. We therefore use $K=8$ as a high-accuracy numerical
piecewise-path reference rather than as an exact closed-form geodesic
distance.

\paragraph{Principal-logarithm domain.}
The definition of $D_{\mathbb{SL}}$ requires the relative matrices to lie
in the principal-logarithm domain. We therefore monitor the spectrum of
relative matrices
\(
R=A^{-1}B
\)
throughout training and evaluation. For an eigenvalue
$\lambda=|\lambda|e^{i\theta}$, we define its angular margin to the
negative-real branch cut by
\begin{equation}
m(R)
=
\min_{\lambda\in\sigma(R)}
\bigl(\pi-|\arg\lambda|\bigr).
\end{equation}
A positive margin ensures the spectrum avoids the principal-log branch cut.

\begin{table}[h]
\centering
\caption{
Principal-logarithm domain statistics on KEGG
}
\label{tab:log_stability}
\small
\setlength{\tabcolsep}{7pt}
\begin{tabular}{l r}
\toprule
Statistic & Value \\
\midrule
Relative matrices evaluated
& $15{,}488$ \\
Principal-logarithm domain rate
& $100\%$ \\
Branch margin $<10^{-2}$
& $0\%$ \\
Branch margin $<10^{-3}$
& $0\%$ \\
Minimum branch margin
& $2.3295$ \\
NaN / Inf rate
& $0\%$ \\
\bottomrule
\end{tabular}
\end{table}

All $15{,}488$ relative matrices encountered during training and evaluation
remain inside the principal-logarithm domain. Moreover, the minimum observed
branch margin is $2.3295$ radians, and no sample approaches either the
$10^{-2}$ or $10^{-3}$ branch-margin thresholds. We observe no NaN or Inf
values. Thus, in this controlled training regime, the logarithmic objective
operates well inside its regular principal domain rather than merely
avoiding the branch cut by a small numerical margin.

Taken together, these experiments clarify the relation between the practical
Schatten semidistance and the intrinsic path geometry. First,
$D_{\mathbb{SL}}$ is exactly the $F_p$-length of the canonical exponential
path and locally upper-bounds the intrinsic distance with second-order
error. Second, allowing additional path segments substantially reduces the
discrepancy to the refined path reference, yet changes end-to-end
reconstruction performance by only $0.46\%$, while increasing training
time by approximately $5.48\times$. Third, all relative matrices observed
in this experiment remain well inside the principal-logarithm domain, and
the symmetric and one-sided formulations coincide up to machine precision.

These results show that the practical performance is robust to substantial
refinement of the underlying path objective, while the closed-form
$D_{\mathbb{SL}}$ retains a clear computational advantage. We therefore
use $D_{\mathbb{SL}}$ in the large-scale experiments and regard the
$K$-segment construction as a controlled numerical approximation for
examining its relation to the intrinsic $\mathbb{SL}$ path geometry.

\newpage


\subsection{Optimization Parameterization and Riemannian Control}
\label{sec:sl_optimization}

The representation geometry and the optimization geometry need not coincide.
In our main experiments, representations remain in $\mathbb{SL}(n)$ and the
Schatten-$p$ geometry enters through the pairwise objective, while a common
Euclidean parameterization is optimized for every $p$. This isolates changes
in the Schatten order from changes in the optimizer. We additionally use a
left trivialized Riemannian AdamW method as a manifold aware control.

\paragraph{Exponential parameterized AdamW.}
We introduce unconstrained $X\in\mathbb{R}^{n\times n}$ and represent
$A(X)=\exp(\Pi_{\mathfrak{sl}}(X))\in\mathbb{SL}(n)$, where $\Pi_{\mathfrak{sl}}(X)
=
X-\operatorname{tr}(X)I/n
\in\mathfrak{sl}(n).$ Then $\det A(X)=1$. AdamW is applied to
$X$, with gradients propagated through the projection and matrix exponential.
We call this \emph{Exponential Parameterized AdamW} (\textsc{Exp-AdamW}).

\begin{algorithm}[h]
\caption{Exponential-Parameterized AdamW on $\mathbb{SL}(n)$}
\label{alg:exp_adamw_sl}
\begin{algorithmic}[1]
\Require Raw matrix $X_0\in\mathbb{R}^{n\times n}$, learning rates
$\{\eta_t\}_{t=1}^{T}$, $\beta_1,\beta_2\in[0,1)$,
$\epsilon>0$, and weight decay $\lambda\geq0$
\State $M_0\gets0$, $V_0\gets0$
\For{$t=1,\ldots,T$}
    \State $\Theta_{t-1}\gets
    \Pi_{\mathfrak{sl}}(X_{t-1})$
    \State $A_{t-1}\gets\exp(\Theta_{t-1})$
    \State $G_t\gets\nabla_X f(A(X))|_{X=X_{t-1}}$
    \State $M_t\gets
    \beta_1M_{t-1}+(1-\beta_1)G_t$
    \State $V_t\gets
    \beta_2V_{t-1}+(1-\beta_2)(G_t\odot G_t)$
    \State $\widehat{M}_t\gets M_t/(1-\beta_1^t)$,
    $\widehat{V}_t\gets V_t/(1-\beta_2^t)$
    \State $X_t\gets
    (1-\eta_t\lambda)X_{t-1}
    -\eta_t\widehat{M}_t/
    (\sqrt{\widehat{V}_t}+\epsilon)$
\EndFor
\State \Return $A_T=\exp(\Pi_{\mathfrak{sl}}(X_T))$
\end{algorithmic}
\end{algorithm}

Here, $\odot$, the square root, and division are applied elementwise.

\paragraph{Left trivialized Riemannian AdamW control.} As a manifold aware control, we optimize $A$ directly on $\mathbb{SL}(n)$ with the left invariant Frobenius metric. Let $G_t=\nabla_A^E f(A_t)$. For a tangent direction $A_tZ$, $df_{A_t}(A_tZ)=\langle A_t^\top G_t,Z\rangle_F$, so the left trivialized Riemannian gradient is $\Xi_t=\Pi_{\mathfrak{sl}}(A_t^\top G_t)$ and the corresponding tangent vector is $A_t\Xi_t$. Left invariance identifies tangent spaces with $\mathfrak{sl}(n)$, allowing the first moment to remain in the Lie algebra and the squared Frobenius norm to serve as a scalar second moment without explicit vector transport. For a smooth regularizer $r$, we use $\Omega_t=\Pi_{\mathfrak{sl}}(A_t^\top\nabla_A^E r(A_t))$ and apply the decay direction outside the adaptive moments, following decoupled weight decay.

\begin{algorithm}[h]
\caption{Left-Trivialized Riemannian AdamW on $\mathbb{SL}(n)$}
\label{alg:riemannian_adamw_sl}
\begin{algorithmic}[1]
\Require $A_0\in\mathbb{SL}(n)$, learning rates
$\{\eta_t\}_{t=1}^{T}$, $\beta_1,\beta_2\in[0,1)$,
$\epsilon>0$, weight decay $\lambda\geq0$, and regularizer $r$
\State $M_0\gets0\in\mathfrak{sl}(n)$, $v_0\gets0$
\For{$t=1,\ldots,T$}
    \State $G_t\gets\nabla_A^E f(A_{t-1})$
    \State $\Xi_t\gets
    \Pi_{\mathfrak{sl}}(A_{t-1}^{\top}G_t)$
    \State $M_t\gets
    \beta_1M_{t-1}+(1-\beta_1)\Xi_t$
    \State $v_t\gets
    \beta_2v_{t-1}+(1-\beta_2)\|\Xi_t\|_F^2$
    \State $\widehat{M}_t\gets M_t/(1-\beta_1^t)$,
    $\widehat{v}_t\gets v_t/(1-\beta_2^t)$
    \State $D_t\gets
    \Pi_{\mathfrak{sl}}\!\left(
    \widehat{M}_t/(\sqrt{\widehat{v}_t}+\epsilon)\right)$
    \State $\Omega_t\gets
    \Pi_{\mathfrak{sl}}\!\left(
    A_{t-1}^{\top}\nabla_A^E r(A_{t-1})\right)$
    \State $A_t\gets
    A_{t-1}\exp\!\left[-\eta_t(D_t+\lambda\Omega_t)\right]$
\EndFor
\State \Return $A_T$
\end{algorithmic}
\end{algorithm}

Every update direction in Algorithm~\ref{alg:riemannian_adamw_sl} is trace
free. Consequently,
$\det(\exp[-\eta_t(D_t+\lambda\Omega_t)])=1$, so the iterates remain in
$\mathrm{SL}(n)$ up to numerical precision. We use
$r(A)=\frac12\|A\|_F^2$, giving
$\Omega_t=\Pi_{\mathfrak{sl}}(A_t^\top A_t)$.
Unlike direct Euclidean shrinkage of $A$, this decay preserves the
determinant constraint. When $\lambda=0$, the method reduces to a left
trivialized Riemannian Adam optimizer.

For $p=2$, Riem-AdamW follows the same left invariant Frobenius geometry as
the representation objective. For $p\neq2$, it serves as a manifold aware
control rather than an intrinsic Schatten-$p$ Finsler optimizer. Exp-AdamW,
in contrast, optimizes an unconstrained parameterization and is not an
intrinsic Finsler gradient method. In both cases, the Schatten-$p$ geometry
enters through the same representation objective. This separation allows us
to test whether the effect of changing $p$ persists independently of the
optimization geometry.

\begin{figure}[h]
    \centering

    \begin{subfigure}[t]{0.45\linewidth}
        \centering
        \includegraphics[width=\linewidth]{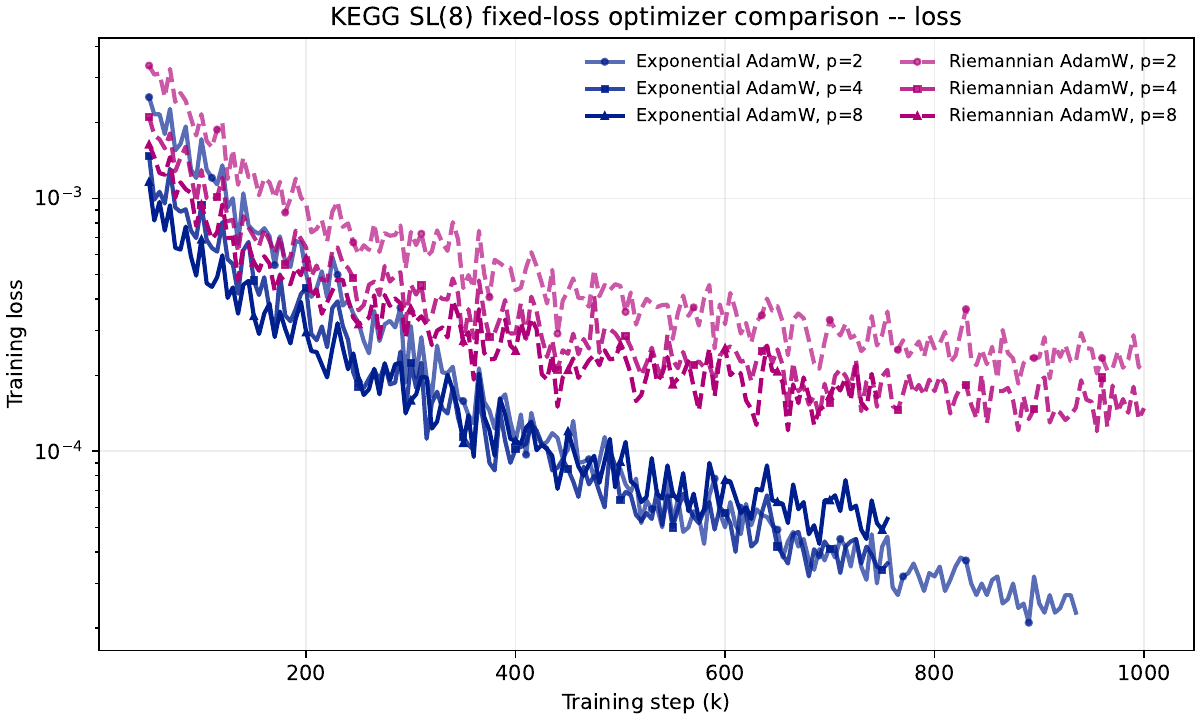}
        \caption{
        \begin{tabular}[t]{@{}c@{}}
        Training loss curves for the optimizer\\
        comparison on $\mathbb{SL}_p(8)$.
        \end{tabular}
        }
        \label{fig:kegg_sl8_fixed_loss}
    \end{subfigure}
    \hfill
    \begin{subfigure}[t]{0.45\linewidth}
        \centering
        \includegraphics[width=\linewidth]{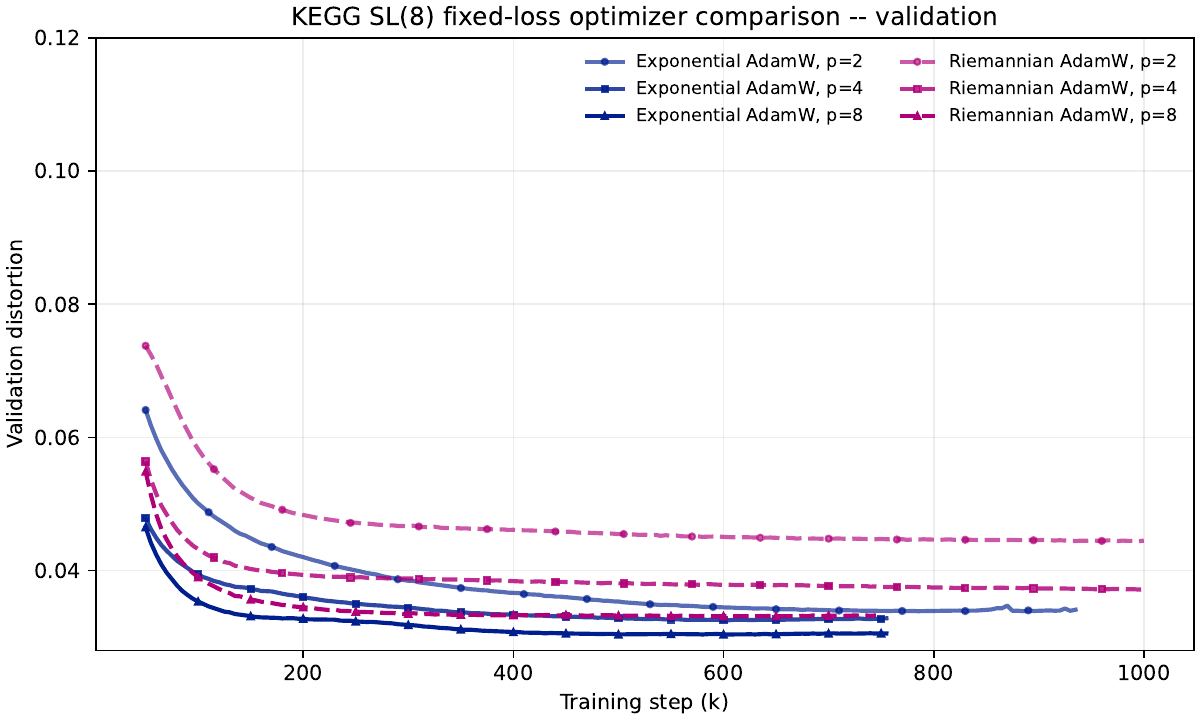}
        \caption{
        \begin{tabular}[t]{@{}c@{}}
        Validation distortion curves for the optimizer\\
        comparison on $\mathbb{SL}_p(8)$.
        \end{tabular}
        }
        \label{fig:kegg_sl8_fixed_validation}
    \end{subfigure}

    \caption{
    KEGG $\mathbb{SL}(8)$ optimizer comparison between
EXP-AdamW and Riemannian AdamW 
under matched Schatten-$p$ objectives.
    }
    \label{fig:kegg_sl8_fixed_optimizer_curves}
\end{figure}

\begin{table}[h]
    \centering
    \caption{
    KEGG $\mathbb{SL}(8)$ optimizer comparison between
EXP-AdamW and Riemannian AdamW
under matched Schatten-$p$ objectives.
    }
    \label{tab:kegg_sl8_fixed_optimizer}

    \resizebox{0.95\linewidth}{!}{%
    \begin{tabular}{llrrrrrrrr}
    \toprule
    $p$
    & Optimizer
    & \textcolor{DeepKleinBlue}{\textbf{Distortion$_{\mathrm{avg.}}$}} $\downarrow$
    & q50 $\downarrow$
    & q90 $\downarrow$
    & q95 $\downarrow$
    & High-related $\downarrow$
    & Low--Low $\downarrow$
    & Worst-group $\downarrow$
    & Best Val. $\downarrow$ \\
    \midrule

    $2$
    & Riem-AdamW
    & $0.04537$
    & $0.01967$
    & $0.10111$
    & $0.17079$
    & $0.05182$
    & $0.04446$
    & $0.10547$
    & $0.04442$ \\

    $2$
    & Exp-AdamW
    & $0.03658$
    & \textcolor{DeepKleinBlue}{\textbf{0.00766}}
    & \underline{0.06910}
    & $0.14796$
    & $0.04301$
    & $0.03629$
    & \underline{0.08452}
    & $0.03391$ \\
    \midrule

    $4$
    & Riem-AdamW
    & $0.03786$
    & $0.01562$
    & $0.08882$
    & $0.14575$
    & $0.04422$
    & $0.03547$
    & $0.08828$
    & $0.03715$ \\

    $4$
    & Exp-AdamW
    & $0.03490$
    & \underline{0.00928}
    & $0.07009$
    & $0.13330$
    & \underline{0.03930}
    & $0.03458$
    & $0.08470$
    & \underline{0.03255} \\
    \midrule

    $8$
    & Riem-AdamW
    & \underline{0.03463}
    & $0.01488$
    & $0.07717$
    & \underline{0.12715}
    & $0.04123$
    & \textcolor{DeepKleinBlue}{\textbf{0.02994}}
    & $0.08971$
    & $0.03313$ \\

    $8$
    & Exp-AdamW
    & \textcolor{DeepKleinBlue}{\textbf{0.03338}}
    & $0.01006$
    & \textcolor{DeepKleinBlue}{\textbf{0.06727}}
    & \textcolor{DeepKleinBlue}{\textbf{0.12430}}
    & \textcolor{DeepKleinBlue}{\textbf{0.03869}}
    & \underline{0.03180}
    & \textcolor{DeepKleinBlue}{\textbf{0.08231}}
    & \textcolor{DeepKleinBlue}{\textbf{0.03041}} \\

    \bottomrule
    \end{tabular}%
    }
\end{table}

\paragraph{Geometry versus optimization.}
Figure~\ref{fig:kegg_sl8_fixed_optimizer_curves} and
Table~\ref{tab:kegg_sl8_fixed_optimizer} show that Exp-AdamW achieves lower
average distortion than Riem-AdamW for $p=2,4,8$, with relative reductions
of $19.4\%$, $7.8\%$, and $3.6\%$, respectively. All runs use seed 0 and
batch size 512.

More importantly, increasing $p$ from $2$ to $8$ improves average distortion
under both optimizers, from $0.04537$ to $0.03463$ under Riem-AdamW and from
$0.03658$ to $0.03338$ under Exp-AdamW. Since the optimizer is fixed within
each comparison, this common trend supports an effect of the Schatten order
on the learned geometric bias rather than an optimizer artifact. At the same
time, the strong performance of Exp-AdamW shows that these gains do not
require an intrinsic manifold optimizer. We therefore use Exp-AdamW in the
main experiments for its simplicity and stronger empirical performance.

\newpage

\section{Experimental Details}
\label{app:experimental_details}

\subsection{Common Experimental Protocol}

We evaluate the representation spaces on KEGG, HumanCyc, OGBL-PPA, and
Flickr30k-Order, covering metric reconstruction, large-scale link prediction,
and multimodal order modeling. Unless stated otherwise, final results are
computed over three independent model initialization seeds. Hyperparameters
and checkpoints are selected using validation data only, and the test split is
accessed only after model selection. For $\mathbb{SL}(n)$, the Schatten order $p$ is selected using validation data:
we use $p=8$ for $\mathbb{SL}(4)$ and $p=24$ for $\mathbb{SL}(8)$ on KEGG,
$p=4$ for $\mathbb{SL}(12)$ on HumanCyc, $p=8$ for $\mathbb{SL}(8)$ on
OGBL-PPA, and $p=2$ for $\mathbb{SL}(4)$ on Flickr30k-Order.

Experiments are run across NVIDIA A100-SXM4-40GB,
NVIDIA A100-SXM4-80GB, and NVIDIA H100 PCIe GPUs. Multiple independent runs
may share one GPU, while maintaining independent model parameters, optimizer
states, random-number states, and checkpoints. Hardware allocation affects
wall-clock time only and does not change the data split, training budget, or
model-selection protocol.

Within each task, all methods share the same data, supervision, training
objective, and evaluation implementation. Only the latent representation,
geometry-specific parameterization, and pairwise dissimilarity or score
function are changed. Method-specific learning rates and batch sizes are
predeclared to accommodate differences in numerical scale, memory footprint,
and computational cost. Training budgets are therefore specified primarily
in optimizer updates or epochs rather than wall-clock time.

\subsection{Datasets and Task Protocols}

\paragraph{KEGG.}
We use the largest connected component of KEGG pathway 24, treated as an
undirected graph with self-loops removed. The resulting graph contains
$377$ nodes and $1{,}545$ edges. Target distances are unweighted shortest-path
distances. All
\(
\binom{377}{2}=70{,}876
\)
unordered node pairs are deterministically partitioned into
$70\%/10\%/20\%$ training, validation, and test sets. Training pairs are
sampled uniformly with replacement.

\paragraph{HumanCyc.}
We use the largest connected component of the HumanCyc-0 pathway graph after
removing self-loops. The resulting graph contains $2{,}682$ nodes and
$28{,}177$ edges. All
\(
\binom{2682}{2}=3{,}595{,}221
\)
unordered node pairs are partitioned using the same deterministic
train/validation/test protocol as KEGG.
For both reconstruction datasets, each node is represented directly by a
trainable point $z_i\in\mathcal M$. No node features, graph encoder, or
message-passing network is used, allowing the experiments to isolate the
representation capacity of the latent geometry.

\paragraph{OGBL-PPA.}
We use the official \texttt{ogbl-ppa} split without modification. The graph
contains $576{,}289$ nodes and $21{,}231{,}931$ training positive edges.
The validation split contains $6{,}062{,}562$ positive and $3{,}000{,}000$
global negative edges, while the test split contains $3{,}031{,}780$
positive and $3{,}000{,}000$ global negative edges. We do not use the
provided node features or message passing.

Training positives are sampled uniformly with replacement from the official
training edges. For every positive edge $(u,v)$, one negative destination
$v^{-}$ is sampled uniformly from all nodes while retaining the source $u$.
Self-loops and edges present in the official training graph are rejected.
Validation and test positives are not consulted by the training sampler.

\paragraph{Flickr30k-Order.}
We use OpenCLIP ViT-B/32 pretrained with the \texttt{openai} weights and keep
the backbone frozen. Training uses the Flickr30k Karpathy training split with
$29{,}000$ images, five captions per image, and $145{,}000$ training cases.
For every training caption, four self-swap negatives are generated by randomly
exchanging two word positions.
Validation and test use the official ARO Flickr30k-Order splits. The
validation set contains $1{,}014$ images, $5{,}070$ caption cases, and
$19{,}160$ valid positive-negative comparisons. The test set contains
$1{,}000$ images, $4{,}995$ retained caption cases, and $18{,}859$
positive-negative comparisons. Corruptions that become identical to the
positive caption after preprocessing are removed.

\subsection{Representation and Baseline Configurations}

For metric reconstruction, all methods share the same node-level learning
interface and reconstruction objective and differ only in the underlying
space and pairwise dissimilarity.

\begin{table}[h]
\centering
\caption{
Geometric baselines used for metric reconstruction.
}
\label{tab:baseline_specifications}

\scriptsize
\setlength{\tabcolsep}{3.5pt}
\renewcommand{\arraystretch}{1.5}
\setlength{\extrarowheight}{1pt}

\resizebox{\linewidth}{!}{%
\begin{tabular}{llll}
\toprule
\textbf{Space}
& \textbf{Intrinsic dimension}
& \textbf{Definition}
& \textbf{Pairwise dissimilarity} \\
\midrule

\textcolor{lightMagenta}{$\mathbb{S}^{d}$}
&
$d$
&
$\{x\in\mathbb R^{d+1}:\|x\|_2=1\}$
&
$\displaystyle
d_{\mathbb S}(x,y)
=
\arccos\!\left(\langle x,y\rangle\right)$
\\

\textcolor{KleinBlue}{$\mathbb{H}^{d}$}
&
$d$
&
$\{x\in\mathbb R^d:\|x\|_2<1\}$
&
$\displaystyle
d_{\mathbb H}(x,y)
=
\operatorname{arcosh}\!\left(
1+
\frac{2\|x-y\|_2^2}
{(1-\|x\|_2^2)(1-\|y\|_2^2)}
\right)$
\\

$\mathbb{E}^{d}$
&
$d$
&
$\mathbb R^d$
&
$\displaystyle
d_{\mathbb E}(x,y)=\|x-y\|_2$
\\

\addlinespace[1mm]

$\textcolor{lightMagenta}{\mathbb{S}^{d}}
\times
\textcolor{KleinBlue}{\mathbb{H}^{d}}$
&
$2d$
&
$\mathbb S^d\times\mathbb H^d$
&
$\displaystyle
d=
\left(
d_{\mathbb S}^{2}
+d_{\mathbb H}^{2}
\right)^{1/2}$
\\

$\textcolor{lightMagenta}{\mathbb{S}^{d}}
\times
\mathbb{E}^{d}$
&
$2d$
&
$\mathbb S^d\times\mathbb E^d$
&
$\displaystyle
d=
\left(
d_{\mathbb S}^{2}
+d_{\mathbb E}^{2}
\right)^{1/2}$
\\

$\textcolor{KleinBlue}{\mathbb{H}^{d}}
\times
\mathbb{E}^{d}$
&
$2d$
&
$\mathbb H^d\times\mathbb E^d$
&
$\displaystyle
d=
\left(
d_{\mathbb H}^{2}
+d_{\mathbb E}^{2}
\right)^{1/2}$
\\

\addlinespace[1mm]

$\textcolor{lightMagenta}{\mathbb{S}^{d}}
\times
\textcolor{KleinBlue}{\mathbb{H}^{d}}
\times
\mathbb{E}^{d}$
&
$3d$
&
$\mathbb S^d\times\mathbb H^d\times\mathbb E^d$
&
$\displaystyle
d=
\left(
d_{\mathbb S}^{2}
+d_{\mathbb H}^{2}
+d_{\mathbb E}^{2}
\right)^{1/2}$
\\

$\textcolor{lightMagenta}{\mathbb{S}^{d}_{\kappa_1}}
\times
\textcolor{KleinBlue}{\mathbb{H}^{d}_{\kappa_2}}
\times
\mathbb{E}^{d}$
&
$3d$
&
$\substack{c_{\mathbb S},c_{\mathbb H}>0,\\
w_{\mathbb S},w_{\mathbb H},w_{\mathbb E}\ge0}$
&
$\displaystyle
d=
\left(
w_{\mathbb S}d_{\mathbb S_{c_{\mathbb S}}}^{2}
+
w_{\mathbb H}d_{\mathbb H_{-c_{\mathbb H}}}^{2}
+
w_{\mathbb E}d_{\mathbb E}^{2}
\right)^{1/2}$
\\

\addlinespace[1mm]

Heisenberg-$H^{2d+1}$
&
$2d+1$
&
$(x,y,t)\in\mathbb R^d\times\mathbb R^d\times\mathbb R$
&
$\displaystyle
d_H(P,Q)
=
\inf_{\substack{\gamma(0)=P\\ \gamma(1)=Q}}
\int_0^1
\sqrt{
\|\dot x\|_2^2+\|\dot y\|_2^2+
\left(
\dot t+\frac12(y^\top\dot x-x^\top\dot y)
\right)^2
}\,ds$
\\

\addlinespace[1mm]

$\textcolor{deepPurple}{\mathbf{Grassmann}}(k,n)$
&
$k(n-k)$
&
$\{\operatorname{span}(Q):Q^\top Q=I_k\}$
&
$\displaystyle
d_{\mathbf{Grassmann}}(Q_i,Q_j)
=
\left\|
\arccos
\sigma(Q_i^\top Q_j)
\right\|_2$
\\

$\textcolor{lightKleinBlue}{\mathbf{SPD}(n)}$
&
$\dfrac{n(n+1)}{2}$
&
$\{P=P^\top\succ0\}$
&
$\displaystyle
d_{\mathbf{SPD}}(P_i,P_j)
=
\left\|
\log\!\left(
P_i^{-1/2}P_jP_i^{-1/2}
\right)
\right\|_F$
\\

$\textcolor{DeepKleinBlue}{\mathbf{Siegel}(n)}$
&
$n(n+1)$
&
$\{X+\sqrt{-1}Y:X=X^\top,\ Y\succ0\}$
&
$\displaystyle
d_{\mathbf{Siegel}}(i,j)
=
\sum_{\ell=1}^{n}
a_\ell
\log
\frac{1+\sigma_\ell(W_{ij})}
     {1-\sigma_\ell(W_{ij})},
\quad
a_\ell\ge0,\ 
\sum_{\ell}a_\ell=n$
\\

\addlinespace[1mm]

\rainbowmath{\mathbb{SL}(n)}
&
$n^2-1$
&
$\{A\in\mathbb R^{n\times n}:\det(A)=1\}$
&
$\displaystyle
D_{\mathbb{SL}}(A,B)
=
\frac{1}{2}
\left(
\left\|\log(A^{-1}B)\right\|_{S_p}
+
\left\|\log(B^{-1}A)\right\|_{S_p}
\right)$
\\

\bottomrule
\end{tabular}%
}
\end{table}

For the learnable-curvature product model, the spherical and hyperbolic
curvature magnitudes and the nonnegative factor weights are learned jointly.
The factor weights are normalized to have mean one. For the Siegel baseline,
we use
\(
Z_{ij}=Y_i^{-1/2}(Z_j-X_i)Y_i^{-1/2}
\)
and
\(
W_{ij}
=
(Z_{ij}-\sqrt{-1}I)(Z_{ij}+\sqrt{-1}I)^{-1},
\)
where $\sigma_\ell(W_{ij})$ denotes the corresponding Takagi singular values.

The Heisenberg group used in our experiments admits the matrix realization
\begin{equation}
H(x,y,t)
=
\begin{pmatrix}
1
&
x^\top
&
t+\frac12 x^\top y
\\
0
&
I_d
&
y
\\
0
&
0
&
1
\end{pmatrix},
\qquad
x,y\in\mathbb R^d,\quad t\in\mathbb R.
\end{equation}

The unitriangular group is
\(
\mathrm{UT}(m)
=
\left\{
U\in\mathbb R^{m\times m}
:
U_{ii}=1,\;
U_{ij}=0\ \text{for }i>j
\right\},
\)
with the schematic form
\begin{equation}
U=
\begin{pmatrix}
1 & * & * & \cdots & *\\
0 & 1 & * & \cdots & *\\
0 & 0 & 1 & \cdots & *\\
\vdots & \vdots & \ddots & \ddots & \vdots\\
0 & 0 & \cdots & 0 & 1
\end{pmatrix}.
\end{equation}

For non-matrix representation spaces, we match intrinsic dimension to that of
the corresponding $\mathbb{SL}(n)$ model as closely as possible. For matrix
manifolds, we use comparable matrix-scale configurations whenever a natural
matrix correspondence is available. Intrinsic dimensions and trainable
parameter counts are reported explicitly for all methods.

\paragraph{OGBL-PPA representation.}
For OGBL-PPA, we use $\mathbb{SL}_{8}(8)$. Each node stores
$63$ sparse Lie-algebra coordinates $x_i$, which are mapped to the group as
\begin{equation}
A_i
=
\exp\!\left(
0.05\,\Pi_{\mathfrak{sl}}(x_i)
\right).
\end{equation}
The link score is
\begin{equation}
s(u,v)
=
b-
\exp\!\bigl(\operatorname{clip}(\rho,-5,5)\bigr)
D_{\mathcal G}(z_u,z_v),
\end{equation}
where $b$ and $\rho$ are learned scalar parameters.

\paragraph{Flickr30k-Order composition models.}
Frozen CLIP token and image features are passed through separate two-layer
projection heads with hidden width $512$. The output coordinates are mapped
to the corresponding Lie algebra and then to the group using
\begin{equation}
g_t
=
\operatorname{exp}_{\mathcal G}(\alpha X_t),
\qquad
g_I
=
\operatorname{exp}_{\mathcal G}(\alpha_I X_I),
\qquad
\alpha=\alpha_I=0.1.
\end{equation}
Caption tokens are composed in their original left-to-right order by
\begin{equation}
G_t=g_tG_{t-1},
\qquad
G_{1:T}=g_Tg_{T-1}\cdots g_1.
\end{equation}
We compare Heisenberg-$H^7$, $\mathrm{UT}(4)$, $\mathrm{UT}(6)$,
$\mathrm{UT}(15)$, full $\mathrm{SL}(4)$, and a commutative control. The
commutative control retains the same $\mathfrak{sl}(4)$ token
parameterization but replaces ordered multiplication by
\(
G^{\mathrm{comm}}_{1:T}
=
\exp\!\left(
\alpha\sum_{t=1}^{T}X_t
\right).
\)
The image-caption score is
\begin{equation}
s(I,C)
=
-\beta
D_{\mathcal G}(g_I,G_{1:T}),
\qquad
\beta=\exp(\tau),
\end{equation}
where $\beta$ is learned, initialized to $10$, and upper-bounded by $100$.

\subsection{Training and Optimization}

\paragraph{Metric reconstruction.}
KEGG and HumanCyc share the reconstruction objective
\begin{equation}
\mathcal L
=
\frac{1}{|\mathcal B|}
\sum_{(i,j)\in\mathcal B}
\left[
\log\!\left(
1+sD_{\mathcal M}(z_i,z_j)
\right)
-
\log\!\left(
1+d_G(i,j)
\right)
\right]^2.
\end{equation}
The logarithmic transformation prevents distant graph pairs from dominating
the objective.

For KEGG, batch sizes are $8192$ for product models, $2048$ for Grassmann,
$128$ for Siegel, and $512$ otherwise. Learning rates are
$2\times10^{-3}$ for products, $2\times10^{-4}$ for Siegel, and
$5\times10^{-4}$ otherwise. We use weight decay $10^{-6}$ throughout and
gradient clipping at $5$ for Heisenberg and $10$ otherwise.
All KEGG high-capacity methods use at most $800{,}000$ optimizer updates.
Validation is performed every $5{,}000$ updates, with early stopping after
$150{,}000$ updates without improvement.

HumanCyc uses a common maximum budget of $2{,}000{,}000$ updates, batch size
$64$, evaluation batch size $128$, learning rate $2\times10^{-4}$, weight
decay $10^{-6}$, and gradient-norm clipping at $10$. Validation is performed
every $10{,}000$ updates, with early stopping after $50$ consecutive
non-improving validation events. Neither reconstruction experiment uses
learning-rate warmup or a learning-rate scheduler.

\paragraph{OGBL-PPA.}
Training minimizes the pairwise BPR objective
\begin{equation}
\mathcal L_{\mathrm{BPR}}
=
-\frac{1}{|\mathcal B|}
\sum_{(u,v)\in\mathcal B}
\log\sigma
\left(
s(u,v)-s(u,v^-)
\right).
\end{equation}
The effective positive batch size is $4{,}194{,}304$, evaluated through
$32$ gradient-accumulation microbatches of $131{,}072$ edges each.

Sparse node-coordinate rows are optimized using
\emph{exponential-parameterized SparseAdam} with learning rate
$3\times10^{-3}$. The decoder scale and bias are optimized separately using
Adam with learning rate $6\times10^{-3}$. No weight decay, learning-rate
warmup, or learning-rate scheduler is used. Coordinates are clipped to
$[-0.75,0.75]$, sparse gradients are clipped elementwise to
$[-0.05,0.05]$, and dense gradients are clipped to global norm $1.0$.

Training uses at most $17{,}000$ optimizer updates. Validation is performed
every $100$ updates, with early stopping after $20$ consecutive
non-improving validation evaluations, corresponding to $2{,}000$ optimizer
updates.

\paragraph{Flickr30k-Order.}
Training minimizes the pairwise margin-ranking objective
\begin{equation}
\mathcal L_{\mathrm{ord}}
=
\frac{1}{\sum_i K_i}
\sum_i
\sum_{j=1}^{K_i}
\left[
0.1
-
s(I_i,C_i^+)
+
s(I_i,C_{ij}^-)
\right]_+.
\end{equation}
All composition models use AdamW with learning rate $10^{-3}$, weight decay
$10^{-4}$, batch size $192$, evaluation batch size $192$, and global
gradient-norm clipping at $1.0$. Training lasts $12$ epochs and uses neither
warmup nor a learning-rate scheduler. Validation is performed once per epoch,
and no early stopping is used.

\subsection{Evaluation Metrics}

For metric reconstruction, define the relative distortion of a held-out pair
as
\begin{equation}
\delta_{ij}
=
\frac{
\left|
sD_{\mathcal M}(z_i,z_j)-d_G(i,j)
\right|
}{
\max\{d_G(i,j),1\}
}.
\end{equation}

The reported metrics across the three experimental settings are summarized
below.

\begin{table}[h]
\centering
\caption{Evaluation metrics used across the experiments.}
\label{tab:evaluation_metrics}
\scriptsize
\setlength{\tabcolsep}{4pt}
\renewcommand{\arraystretch}{1.25}
\resizebox{\linewidth}{!}{%
\begin{tabular}{lll}
\toprule
\textbf{Task}
& \textbf{Metric}
& \textbf{Definition / interpretation}\\
\midrule

Metric reconstruction
& Distortion$_{\mathrm{avg.}}$
& Mean relative distortion $\delta_{ij}$ over all test pairs\\

Metric reconstruction
& q50 / q90 / q95
& $50$th, $90$th, and $95$th percentiles of $\delta_{ij}$\\

Metric reconstruction
& MAE
& Mean absolute error between scaled latent and graph distances\\

Metric reconstruction
& High-related
& Mean distortion for pairs with at least one high-mixture endpoint\\

Metric reconstruction
& Low--Low
& Mean distortion for pairs whose endpoints are both in the low-mixture group\\

Metric reconstruction
& Worst-group
& Largest mean distortion among endpoint-group combinations\\

\addlinespace[1mm]

OGBL-PPA
& Hits@20 / Hits@50 / Hits@100
& Positive-edge ranking against the common global negative pool\\

OGBL-PPA
& MRR
& Mean reciprocal rank of positive edges\\

OGBL-PPA
& Rank50 / Rank90 / Rank95
& Corresponding quantiles of the positive-edge rank distribution\\

OGBL-PPA
& AUC / AP
& Classification metrics over the complete positive and negative score sets\\

\addlinespace[1mm]

Flickr30k-Order
& OrderAcc
& Fraction of positive captions scoring above individual order corruptions\\

Flickr30k-Order
& OrderMargin
& Mean positive-minus-negative score margin\\

Flickr30k-Order
& HardAcc
& Fraction of cases where the positive caption exceeds every corruption\\

Flickr30k-Order
& MRR
& Mean reciprocal rank of the positive caption within each candidate set\\

\bottomrule
\end{tabular}%
}
\end{table}

For reconstruction, nodes are grouped by local curvature-mixture entropy into
low, medium, and high groups using thresholds $0.3$ and $0.7$. Group-based
metrics are diagnostic only and are not used for checkpoint selection.

For OGBL-PPA, Hits@100 is computed using the official
\texttt{Evaluator(name="ogbl-ppa")}. Hits@20, Hits@50, MRR, and rank
quantiles use the same global negative pool. For a positive edge $e^+$,
\begin{equation}
\operatorname{rank}(e^+)
=
1+
\sum_{e^-\in\mathcal E^-}
\mathbf{1}\!\left[
s(e^-)\geq s(e^+)
\right].
\end{equation}
AUC and AP are computed from the complete positive and negative score sets.

\subsection{Fairness and Model Selection}

Within each task, all baselines use the same data split, supervision,
objective, validation protocol, and test evaluation. Test data are never used
for choosing geometry parameters, learning rates, stopping points, or
checkpoints.
KEGG and HumanCyc select the checkpoint with the lowest validation mean
distortion. All remaining reconstruction metrics are computed from this same
checkpoint.

For OGBL-PPA, model selection uses one fixed validation subset containing
$50{,}000$ official validation positives and $200{,}000$ official validation
negatives. The subset is sampled once using a fixed random seed and shared
across all methods and model seeds. Checkpoints are selected by Hits@100 on
this subset. After model selection, final metrics are recomputed on the
complete official validation and test splits.

Flickr30k-Order evaluates the full validation set after each epoch and selects
the checkpoint with the highest OrderAcc, using OrderMargin only to break ties.

Final results use three independent model initialization seeds and report the
mean and standard deviation when applicable. Method-specific optimization
hyperparameters are fixed before final test evaluation.

\newpage

\section{Proofs of the Theoretical Results}
\label{app:proofs}

\subsection{Preliminaries}
\label{app:preliminaries}

We provide the Finsler geometric background needed for the curvature analysis
below. The main text introduced the fundamental tensor and flag curvature
compactly. Here we develop these objects from the underlying Finsler norm,
explain their geometric meaning, and describe how the Chern curvature and
Jacobi operators used in our proofs are computed.

\paragraph{Finsler metric and directional geometry.}
Let $M$ be a smooth manifold and $TM$ its tangent bundle. A Riemannian metric
assigns an inner product $g_x$ to each tangent space $T_xM$, so the length of a
tangent vector $y\in T_xM$ is determined by
\(\sqrt{g_x(y,y)}\). In particular, the local quadratic geometry at $x$ does
not depend on the direction along which it is examined.

A Finsler metric generalizes this construction by directly assigning a norm
\[
F:TM\rightarrow[0,\infty)
\]
to tangent vectors~\citep{bao2000riemannfinsler}. For every $x\in M$, the
restriction $F(x,\cdot)$ is positively homogeneous,
\(F(x,\lambda y)=\lambda F(x,y)\) for $\lambda>0$, and its squared norm is
strongly convex in the tangent direction. The essential distinction from
Riemannian geometry is that $F^2(x,y)$ need not be quadratic in $y$.
Consequently, the local geometry obtained by differentiating $F$ may depend
on the reference direction $y$ itself.

Classically, the differential Finsler structure is considered on the slit
tangent bundle \(TM\setminus\{0\}\), since a positively homogeneous norm need
not be differentiable at the zero vector. In our Schatten setting there is an
additional regularity distinction. The norm
\(
F_p(A,V)=\|A^{-1}V\|_{S_p}
\)
is defined for every tangent vector and hence defines the global length
structure used throughout the paper. For $p\neq2$, however, the smooth
differential quantities involved in curvature are considered on full rank
tangent directions. We call such directions \emph{regular}. Thus the
regularity restriction concerns differential curvature analysis, not the
definition of the tangent norm or the induced path length.

\paragraph{Fundamental tensor.}
The first local geometric object derived from a Finsler metric is its
\emph{fundamental tensor}. For a regular nonzero reference direction
$y\in T_xM$, define
\begin{equation}
g_y(u,v)
:=
\frac{1}{2}
\left.
\frac{\partial^2}{\partial s\,\partial t}
F^2(x,y+su+tv)
\right|_{s=t=0}.
\label{eq:app_fundamental_tensor}
\end{equation}
Equivalently, if the energy function
\(\mathcal E(x,y)=\frac12F^2(x,y)\), then
\(g_y=D_y^2\mathcal E(x,y)\).

The fundamental tensor can be understood as the local quadratic
approximation of the squared Finsler norm around the direction $y$. Hence,
although $F$ itself may be nonquadratic, $g_y$ provides an inner product with
which infinitesimal lengths, angles, orthogonality, and curvature can be
measured around that particular direction. 
For a Riemannian metric,
\(F^2(x,y)=g_x(y,y)\), so differentiating twice simply recovers $g_x$ and
the result is independent of $y$. Finsler geometry therefore contains
Riemannian geometry as the special case in which this directional dependence
disappears.

By the two homogeneity of \(F^2\), the fundamental tensor satisfies
\(
g_y(y,y)=F^2(x,y)
\)
and
\(
g_y(y,u)=
\frac12D_yF^2(x,y)[u].
\)
These identities will be repeatedly used below.

\paragraph{Cartan tensor and departure from Riemannian geometry.}
The variation of the fundamental tensor with respect to the reference
direction is measured by the \emph{Cartan tensor},
\begin{equation}
C_y(u,v,w)
:=
\frac{1}{2}
\left.
\frac{d}{dt}
g_{y+tw}(u,v)
\right|_{t=0}
=
\frac{1}{4}
D_y^3F^2(x,y)[u,v,w].
\label{eq:app_cartan_tensor}
\end{equation}
Intuitively, $g_y$ describes the local quadratic geometry seen from $y$,
whereas $C_y$ measures how this quadratic geometry changes when the viewing
direction changes. For a Riemannian metric, $g_y$ is independent of $y$ and
hence $C_y=0$. The nonzero Cartan tensor is therefore one of the fundamental
sources of genuinely Finsler behavior.
Although we do not explicitly manipulate $C_y$ in the main curvature proof,
its effect is implicitly contained in the Chern connection introduced next.

\paragraph{Length, geodesics, and the geodesic spray.}
For a piecewise smooth curve $\gamma(t)$, its Finsler length is
\(
L_F(\gamma)=\int F(\gamma(t),\dot\gamma(t))\,dt.
\)
A geodesic is locally a critical curve of the corresponding energy functional.
As in Riemannian geometry, geodesics describe locally straight motion, but the
direction dependence of $F$ changes their equations.

To make this explicit, take local coordinates
$(x^1,\ldots,x^d)$ on $M$ and write a tangent vector as
\(y=y^i\partial_{x^i}\)
(with $\partial_{x^i}$ playing the role of the Euclidean coordinate basis
vector $e_i$). Let \(g_{ij}(x,y)\) denote the matrix of the
fundamental tensor and \(g^{ij}(x,y)\) its inverse. The geodesic spray
coefficients are
\begin{equation}
G^i(x,y)
=
\frac14 g^{i\ell}(x,y)
\left(
\frac{\partial^2 F^2}{\partial x^k\partial y^\ell}y^k
-
\frac{\partial F^2}{\partial x^\ell}
\right).
\label{eq:app_spray}
\end{equation}
A constant speed Finsler geodesic satisfies
\(
\ddot x^i+2G^i(x,\dot x)=0.
\label{eq:app_geodesic}
\)
Thus the spray $G$ plays the role of the Christoffel symbols contracted with
velocity in Riemannian geometry. Starting from $F$, one can therefore obtain
the geodesic dynamics entirely by differentiation.
The derivatives
\begin{equation}
N^i{}_j(x,y):=\partial G^i(x,y)/\partial y^j
\end{equation}
define the associated nonlinear connection. They separate changes in the
base point from changes in the tangent direction and introduce the horizontal
derivatives
\(
\delta_j
=
\partial_{x^j}
-
N^m{}_j\partial_{y^m}.
\)
These quantities provide the intermediate step from the Finsler norm to its
canonical connection and curvature.

\paragraph{Chern connection.}
Curvature requires comparing tangent vectors at nearby points, which in turn
requires a connection. In Riemannian geometry this role is played by the
Levi--Civita connection. Because the Finsler inner product $g_y$ also depends
on the direction $y$, there is in general no ordinary Levi--Civita connection
depending only on the base point. The standard replacement is the
\emph{Chern connection}~\citep{bao2000riemannfinsler}.
The Chern connection is the canonical torsion free connection that is
compatible with the direction dependent fundamental tensor in the Finsler
sense. In local coordinates its coefficients can be obtained from $g_y$ and
the horizontal derivatives as
\begin{equation}
\Gamma^i{}_{jk}(x,y)
=
\frac12 g^{i\ell}
\left(
\delta_j g_{\ell k}
+
\delta_k g_{\ell j}
-
\delta_\ell g_{jk}
\right).
\label{eq:app_chern_coeff}
\end{equation}
All quantities in this expression depend on both the point $x$ and the
reference direction $y$. This is the main difference from the Riemannian
Christoffel symbols, which depend only on $x$.
Equations~(\ref{eq:app_fundamental_tensor}),
(\ref{eq:app_spray}), and~(\ref{eq:app_chern_coeff}) give a direct
computational chain
\[
F
\longrightarrow
g_y
\longrightarrow
G
\longrightarrow
\Gamma(x,y).
\]

\paragraph{Chern curvature.}
Curvature measures the failure of parallel transport defined by the
connection to commute around infinitesimal loops. For the Chern connection,
we denote the corresponding curvature operator at reference direction $y$ by
\(
\boldsymbol{\mathcal R}^{y}(u,v)w.
\)
Under the curvature convention used in this paper, it is the Finsler
counterpart of
\(
\nabla_u\nabla_vw-\nabla_v\nabla_uw-\nabla_{[u,v]}w
\)
from Riemannian geometry.

In local coordinates, the horizontal Chern curvature coefficients are
obtained by differentiating the connection coefficients,
\begin{equation}
R^i{}_{jkl}
=
\delta_k\Gamma^i{}_{jl}
-
\delta_l\Gamma^i{}_{jk}
+
\Gamma^m{}_{jl}\Gamma^i{}_{mk}
-
\Gamma^m{}_{jk}\Gamma^i{}_{ml}.
\label{eq:app_chern_curvature}
\end{equation}
Consequently,
\(
\boldsymbol{\mathcal R}^{y}(u,v)w
\)
is obtained by contracting these coefficients with $u$, $v$, and $w$.
Together, the computational dependence is
\[
F
\longrightarrow
g_y
\longrightarrow
G
\longrightarrow
\Gamma
\longrightarrow
\boldsymbol{\mathcal R}^{y}.
\]

Because the Chern connection itself depends on the reference direction,
varying that direction may introduce additional anisotropic terms compared
with the curvature of an ordinary affine connection. These terms can be
described through the vertical variation of the Chern connection
\citep{javaloyes2014chern,javaloyes2019anisotropic}. They vanish in several
important situations used below, allowing the Finsler curvature to be
computed through an associated affine connection. We make this
specialization explicit when it is used.

\paragraph{Jacobi operator.}
The full Chern curvature
\(\boldsymbol{\mathcal R}^{y}(u,v)w\) depends on three tangent directions.
For sectional or flag curvature, two of these directions are fixed by the
reference direction $y$. This motivates the \emph{Jacobi operator},
\begin{equation}
R_yu
:=
\boldsymbol{\mathcal R}^{y}(u,y)y.
\label{eq:app_jacobi}
\end{equation}
Thus $R_y$ is a linear operator on the tangent space obtained by inserting
the flagpole $y$ twice into the full curvature tensor.
Geometrically, $R_yu$ describes how a nearby geodesic initially separated
from the reference geodesic in direction $u$ accelerates relative to it.
Positive and negative values of its quadratic form therefore correspond to
qualitatively different local bending of nearby geodesics.

The Jacobi operator satisfies \(R_yy=0\) and is self adjoint with respect to
the fundamental tensor,
\begin{equation}
g_y(R_yu,v)=g_y(u,R_yv).
\label{eq:app_jacobi_self_adjoint}
\end{equation}
Hence its eigenvalues are real. On the $g_y$ orthogonal complement of $y$,
we denote its positive and negative eigenspaces by
\(
E_+(y)
\)
and
\(
E_-(y)
\), respectively. These eigenspaces provide the positive and negative
curvature modes used in the curvature coupling definition of the main text.

\paragraph{Flag curvature.}
\begin{wrapfigure}{r}{0.4\textwidth}
\vspace{-4pt}
  \centering
  \includegraphics[width=0.4\textwidth]{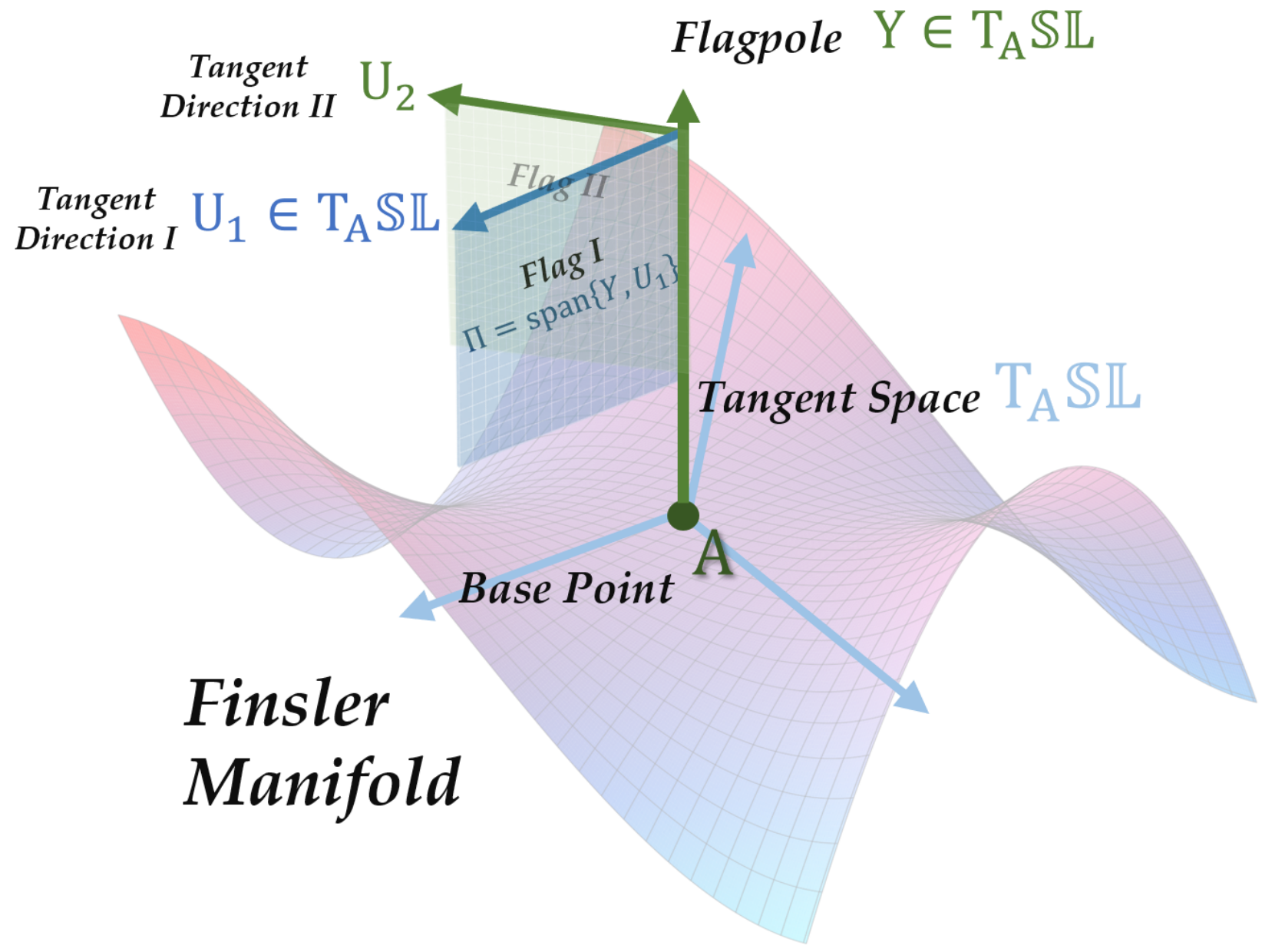}
  \caption{A Finsler flag consists of a tangent plane $\Pi$ together with a
  distinguished tangent direction $y$. }
  \label{fig:flag}
  \vspace{-6pt}
\end{wrapfigure}

In Riemannian geometry, sectional curvature depends only on a two dimensional
plane. In Finsler geometry the local metric depends additionally on the
direction used to inspect that plane. A \emph{flag} is therefore a pair
$(y,\Pi)$ consisting of a nonzero reference direction $y$, called the
\emph{flagpole}, and a two dimensional tangent plane
\(\Pi=\operatorname{span}\{y,u\}\) containing it.

The corresponding \emph{flag curvature} is
\begin{equation}
K_F(y,\Pi)
=
\frac{
g_y(R_yu,u)
}{
g_y(y,y)g_y(u,u)-g_y(y,u)^2
}.
\label{eq:app_flag_curvature}
\end{equation}
The denominator is the Gram determinant of $y$ and $u$ under $g_y$ and is
positive whenever they are linearly independent. As shown in Fig.~\ref{fig:flag}, flag curvature measures the signed curvature
of $\Pi$ as viewed from the distinguished direction $y$. For visual clarity,
the figure uses $Y$ and $U$ for $y$ and $u$ in the text.

As only the component of $u$ transverse to $y$ determines the plane, we may
choose $u$ such that \(g_y(y,u)=0\). Then
\begin{equation}
K_F(y,\operatorname{span}\{y,u\})
=
\frac{g_y(R_yu,u)}
{g_y(y,y)g_y(u,u)}.
\label{eq:flag_sign}
\end{equation}
The denominator is positive, so
\(
\operatorname{sign} K_F
=
\operatorname{sign} g_y(R_yu,u).
\label{eq:app_curvature_sign}
\)
This is why the curvature analysis below focuses on the quadratic form
$g_y(R_yu,u)$ rather than repeatedly evaluating the complete fraction in
Eq.~(\ref{eq:app_flag_curvature}).

\paragraph{Specialization to the Schatten $p$ geometry.}
We now specialize these general constructions to
$\mathbb{SL}_p(n)$. Left invariance identifies the geometry at every point
with the geometry on the Lie algebra. For a regular full rank
$X\in\mathfrak{sl}(n)$, define the energy function
\begin{equation}
\mathcal E_p(X)
=
\frac12\|X\|_{S_p}^2
=
\frac12
\left[
\operatorname{tr}((X^\top X)^{p/2})
\right]^{2/p}.
\label{eq:app_schatten_energy}
\end{equation}
The fundamental tensor at the identity is simply
\begin{equation}
g_X(U,V)
=
D^2\mathcal E_p(X)[U,V].
\label{eq:app_schatten_fundamental}
\end{equation}
At an arbitrary $A\in\mathrm{SL}(n)$, setting
$X=A^{-1}Y$, $\widetilde U=A^{-1}U$, and
$\widetilde V=A^{-1}V$ gives
\(
g_Y(U,V)=g_X(\widetilde U,\widetilde V).
\)
Thus it is sufficient to compute the differential geometry at the identity
and transport the result by left multiplication.
This reduction is especially useful for curvature. A reference direction
$X\in\mathfrak{sl}(n)$ is called a \emph{geodesic vector} when the trajectory
generated by $X$ is a geodesic. For a left invariant Finsler metric this is
equivalent to
\begin{equation}
g_X(X,[X,Z])=0
\qquad
\text{for all }Z\in\mathfrak{sl}(n).
\label{eq:app_geodesic_vector}
\end{equation}
At such a direction, the Chern connection can be represented algebraically
by the connection operator $N_X$~\citep{xu2017homogeneousfinslerspacesflagwise}. 
\begin{equation}
2g_X(N_X(V),W)
=
g_X([W,V],X)
+
g_X([W,X],V)
+
g_X([V,X],W).
\label{eq:homogeneous_connection_short}
\end{equation}
This formula is the homogeneous counterpart of the coordinate Chern
connection in Eq.~(\ref{eq:app_chern_coeff}). Once the fundamental tensor is
known, the right hand side contains only inner products and matrix
commutators, so $N_X(V)$ can be solved directly.
At a geodesic reference direction, the Jacobi operator then takes the form
\begin{equation}
R_X(V)
=
-N_X(N_X(V))
+
N_X([X,V])
-
[X,N_X(V)].
\label{eq:jacobi_N_short}
\end{equation}
Hence the curvature calculation used in our proofs follows the concrete
sequence
\begin{equation}
\mathcal E_p
\ \longrightarrow\
g_X
\ \longrightarrow\
N_X
\ \longrightarrow\
R_X
\ \longrightarrow\
K_F.
\label{eq:app_curvature_pipeline}
\end{equation}
This avoids solving the geodesic equations or evaluating the full coordinate
Chern curvature tensor directly.

\paragraph{Pairwise matrix directions and spectral derivatives.}
The remaining calculations exploit the spectral structure of the Schatten
norm. We choose a diagonal regular reference direction
\(
X=\operatorname{diag}(\lambda_1,\ldots,\lambda_n)
\)
with \(\sum_i\lambda_i=0\). Let \(E_{ij}\in\mathbb{R}^{n\times n}\) denote the
standard matrix unit with a single \(1\) in the \((i,j)\) entry and zeros
elsewhere. For every pair \(i<j\), we define the symmetric and skew-symmetric
directions
\(
S_{ij}:=(E_{ij}+E_{ji})/\sqrt{2}
\)
and
\(
A_{ij}:=(E_{ij}-E_{ji})/\sqrt{2},
\)
respectively. Thus, \(S_{ij}\) and \(A_{ij}\) are the symmetric and
skew-symmetric combinations of the same pair of matrix units
\(E_{ij}\) and \(E_{ji}\), which are
\begin{equation}
S_{ij}
=
\frac{1}{\sqrt2}
\left[
\begin{array}{c|ccccc}
  & \cdots & i & \cdots & j & \cdots \\
\hline
\vdots & \ddots & \vdots &  & \vdots &  \\
i      & \cdots & 0 & \cdots & \textcolor{KleinBlue}{1} & \cdots \\
\vdots &        & \vdots & \ddots & \vdots &  \\
j      & \cdots & \textcolor{KleinBlue}{1} & \cdots & 0 & \cdots \\
\vdots &        & \vdots &  & \vdots & \ddots
\end{array}
\right],
\quad
A_{ij}
=
\frac{1}{\sqrt2}
\left[
\begin{array}{c|ccccc}
  & \cdots & i & \cdots & j & \cdots \\
\hline
\vdots & \ddots & \vdots &  & \vdots &  \\
i      & \cdots & 0 & \cdots & \textcolor{lightMagenta}{1} & \cdots \\
\vdots &        & \vdots & \ddots & \vdots &  \\
j      & \cdots & \textcolor{lightMagenta}{-1} & \cdots & 0 & \cdots \\
\vdots &        & \vdots &  & \vdots & \ddots
\end{array}
\right].
\label{eq:app_root_directions}
\end{equation}
These directions perturb only the $(i,j)$ coordinates of the matrix. They
therefore reduce the high dimensional matrix calculation to simple
two dimensional blocks.
For diagonal $X$, their interaction with the reference direction is
controlled by the spectral difference
\(d_{ij}:=\lambda_i-\lambda_j\), with
\begin{equation}
[X,S_{ij}]
=
d_{ij}A_{ij},
\qquad
[X,A_{ij}]
=
d_{ij}S_{ij},
\qquad
[A_{ij},S_{ij}]
=
E_{ii}-E_{jj}.
\label{eq:root_brackets_short}
\end{equation}
Thus the two directions remain inside the same small matrix block under the
operations entering the connection and curvature formulas.

Differentiating a matrix spectral function such as
\(\operatorname{tr}((X^\top X)^{p/2})\) naturally produces
\emph{divided differences}. Define
\(
\phi_p(t)=t|t|^{p-2}.
\)
For scalars $a\neq b$, its divided difference is
\(
(\phi_p(a)-\phi_p(b))/(a-b)
\),
with the derivative used as the continuous extension when $a=b$.
Divided differences are the matrix analogue of ordinary derivatives when a
perturbation mixes two spectral coordinates. Since $\phi_p$ is strictly
increasing for $p>1$, these divided differences are positive.

This observation explains the quantities $s_{ij}$ and $a_{ij}$ introduced
below: they are precisely the fundamental tensor weights of the symmetric and
skew symmetric directions,
\(
s_{ij}=g_X(S_{ij},S_{ij})
\)
and
\(
a_{ij}=g_X(A_{ij},A_{ij})
\).
Their ratio
\(
m_{ij}(X)=a_{ij}/s_{ij}
\)
summarizes how the Schatten $p$ geometry weights the two directions inside
the same pairwise block. The next proposition computes these quantities and
shows how they enter the Chern curvature.

\paragraph{Regularity and analytic dependence.}
On the full rank matrix locus, $X^\top X$ is positive definite and
\(
X\mapsto
[\operatorname{tr}((X^\top X)^{p/2})]^{2/p}
\)
is real analytic~\citep{higham2008functions}. Consequently, the fundamental
tensor and the resulting Chern curvature coefficients vary analytically with
a regular reference direction wherever the fundamental tensor is
nondegenerate. We later use the standard fact that a nonzero real analytic
scalar function on a connected open set has an open dense nonzero locus.
This allows finitely many nonvanishing curvature interactions to be realized
simultaneously without requiring a specially tuned reference direction.


\subsection{Rootwise Curvature and Proof of Mixed Flag Curvature}
\label{app:rootwise_curvature}

\begin{proposition}[Rootwise effective Cartan curvature]
\label{prop:effective_cartan}
Let
$X=\operatorname{diag}(\lambda_1,\ldots,\lambda_n)
\in\mathfrak{sl}(n)$
be full rank and let
\(
S_{ij}
=
\frac{E_{ij}+E_{ji}}{\sqrt2},
A_{ij}
=
\frac{E_{ij}-E_{ji}}{\sqrt2}, i<j.
\)
Define $\phi_p(t)=t|t|^{p-2}$ and
\begin{equation}
\begin{aligned}
s_{ij}
&:=
g_X(S_{ij},S_{ij})
=
\|X\|_{S_p}^{\,2-p}
\frac{\phi_p(\lambda_i)-\phi_p(\lambda_j)}
{\lambda_i-\lambda_j},
\\
a_{ij}
&:=
g_X(A_{ij},A_{ij})
=
\|X\|_{S_p}^{\,2-p}
\frac{\phi_p(\lambda_i)+\phi_p(\lambda_j)}
{\lambda_i+\lambda_j},
\end{aligned}
\label{eq:sa_def}
\end{equation}
where the divided differences are understood by continuous
extension when a denominator vanishes. Then
$s_{ij}>0$ and $a_{ij}>0$. Define the effective Cartan parameter
\(
m_{ij}(X)
:=
\frac{a_{ij}}{s_{ij}}
>0.
\)
Whenever $\lambda_i\neq\lambda_j$, the two root directions are
eigenvectors of the Jacobi operator and satisfy
\begin{equation}
R_X(A_{ij})
=
\frac{(\lambda_i-\lambda_j)^2}{4}
m_{ij}(X)A_{ij},
\quad
R_X(S_{ij})
=
-
\frac{(\lambda_i-\lambda_j)^2}{4}
\bigl(4+3m_{ij}(X)\bigr)S_{ij}.
\label{eq:effective_cartan_curvature}
\end{equation}
Consequently,
$g_X(R_XA_{ij},A_{ij})>0$ and
$g_X(R_XS_{ij},S_{ij})<0$.
\end{proposition}

\begin{proof}
We first compute the fundamental tensor on the two-dimensional
$(i,j)$ root block. Since
$\mathcal E_p(Z)
=
\frac12(\operatorname{tr}(Z^\top Z)^{p/2})^{2/p}$,
standard second-order spectral calculus at a diagonal full-rank
matrix gives
\begin{equation}
\begin{aligned}
D^2\operatorname{tr}((X^\top X)^{p/2})
[S_{ij},S_{ij}]
&=
p\,
\frac{\phi_p(\lambda_i)-\phi_p(\lambda_j)}
{\lambda_i-\lambda_j},
\\
D^2\operatorname{tr}((X^\top X)^{p/2})
[A_{ij},A_{ij}]
&=
p\,
\frac{\phi_p(\lambda_i)+\phi_p(\lambda_j)}
{\lambda_i+\lambda_j}.
\end{aligned}
\label{eq:spectral_second_variation}
\end{equation}
Every off-diagonal perturbation has zero first variation at
diagonal $X$. Applying the scalar chain rule to
$\mathcal E_p$ therefore multiplies both expressions in
\eqref{eq:spectral_second_variation} by the common positive factor
$\|X\|_{S_p}^{\,2-p}/p$, which gives \eqref{eq:sa_def}.

The function $\phi_p(t)=t|t|^{p-2}$ is strictly increasing for
$p>1$. Hence the first quotient in \eqref{eq:sa_def} is a positive
divided difference of $\phi_p$. For the second quotient, use the
oddness of $\phi_p$ to write
$\phi_p(\lambda_i)+\phi_p(\lambda_j)
=
\phi_p(\lambda_i)-\phi_p(-\lambda_j)$ and
$\lambda_i+\lambda_j
=
\lambda_i-(-\lambda_j)$.
It is therefore again a positive divided difference of the same
strictly increasing function. At a vanishing denominator, the
continuous extension is
$(p-1)|\lambda_i|^{p-2}>0$ because $X$ is full rank.
Thus $s_{ij}>0$, $a_{ij}>0$, and consequently
$m_{ij}(X)>0$.

Transpose invariance of $\mathcal E_p$ gives
$g_X(S_{ij},A_{ij})=0$, while diagonal sign conjugations imply
orthogonality between distinct root blocks. Since $X$ is diagonal,
both $S_{ij}$ and $A_{ij}$ are also $g_X$-orthogonal to $X$.
We next verify that the diagonal reference direction $X$ is
geodesic. By two-homogeneity,
$g_X(X,V)=D\mathcal E_p(X)[V]$. Hence, for every
$Z\in\mathfrak{sl}(n)$,
\begin{equation}
\begin{aligned}
g_X(X,[X,Z])
&=
\|X\|_{S_p}^{\,2-p}
\operatorname{tr}
\left(
X|X|^{p-2}[X,Z]
\right)
\\
&=
\|X\|_{S_p}^{\,2-p}
\operatorname{tr}
\left(
[X|X|^{p-2},X]Z
\right)
=
0,
\end{aligned}
\label{eq:diagonal_geodesic_short}
\end{equation}
because both $X$ and $X|X|^{p-2}$ are diagonal. Thus $X$ is a
geodesic vector.

Let $N_X$ denote the homogeneous Chern connection operator at this
geodesic reference direction. It satisfies
\begin{equation}
2g_X(N_X(V),W)
=
g_X([W,V],X)
+
g_X([W,X],V)
+
g_X([V,X],W).
\label{eq:homogeneous_connection_short}
\end{equation}
The only Lie brackets needed on the $(i,j)$ root block are
\begin{equation}
[X,S_{ij}]
=
(\lambda_i-\lambda_j)A_{ij},
\qquad
[X,A_{ij}]
=
(\lambda_i-\lambda_j)S_{ij},
\qquad
[A_{ij},S_{ij}]
=
E_{ii}-E_{jj}.
\label{eq:root_brackets_short}
\end{equation}
Moreover, two-homogeneity and \eqref{eq:sa_def} give
\begin{equation}
g_X(E_{ii}-E_{jj},X)
=
(\lambda_i-\lambda_j)s_{ij}.
\label{eq:diag_root_pairing}
\end{equation}

We now compute the connection on this root block. Testing
\eqref{eq:homogeneous_connection_short} against $S_{ij}$, against
diagonal directions, and against every distinct root block shows
that $N_X(S_{ij})$ has only an $A_{ij}$ component. Pairing with
$A_{ij}$ and using
\eqref{eq:root_brackets_short}--\eqref{eq:diag_root_pairing} gives
\begin{equation}
\begin{aligned}
2g_X(N_X(S_{ij}),A_{ij})
&=
g_X([A_{ij},S_{ij}],X)
+
g_X([A_{ij},X],S_{ij})
+
g_X([S_{ij},X],A_{ij})
\\
&=
(\lambda_i-\lambda_j)s_{ij}
-
(\lambda_i-\lambda_j)s_{ij}
-
(\lambda_i-\lambda_j)a_{ij}
\\
&=
-(\lambda_i-\lambda_j)a_{ij}.
\end{aligned}
\label{eq:N_S_short}
\end{equation}
Since $g_X(A_{ij},A_{ij})=a_{ij}$, this yields
$N_X(S_{ij})
=
-\frac{\lambda_i-\lambda_j}{2}A_{ij}$.

Similarly, testing
\eqref{eq:homogeneous_connection_short} shows that
$N_X(A_{ij})$ has only an $S_{ij}$ component. Pairing with
$S_{ij}$ gives
\begin{equation}
\begin{aligned}
2g_X(N_X(A_{ij}),S_{ij})
&=
g_X([S_{ij},A_{ij}],X)
+
g_X([S_{ij},X],A_{ij})
+
g_X([A_{ij},X],S_{ij})
\\
&=
-(\lambda_i-\lambda_j)s_{ij}
-
(\lambda_i-\lambda_j)a_{ij}
-
(\lambda_i-\lambda_j)s_{ij}
\\
&=
-(\lambda_i-\lambda_j)(2s_{ij}+a_{ij}).
\end{aligned}
\label{eq:N_A_short}
\end{equation}
Using $g_X(S_{ij},S_{ij})=s_{ij}$ and
$m_{ij}(X)=a_{ij}/s_{ij}$, we therefore obtain the compact
rootwise connection formulas
\begin{equation}
N_X(S_{ij})
=
-\frac{\lambda_i-\lambda_j}{2}A_{ij},
\qquad
N_X(A_{ij})
=
-\frac{\lambda_i-\lambda_j}{2}
\bigl(2+m_{ij}(X)\bigr)S_{ij}.
\label{eq:effective_connection}
\end{equation}

The significance of \eqref{eq:effective_connection} is that the
entire dependence of the Schatten-$p$ fundamental tensor on this
root block is compressed into the single positive scalar
$m_{ij}(X)$. The connection has the similar rootwise
algebraic form as the classical Cartan metric
$\nu_m=-mB|_{\mathfrak k}+B|_{\mathfrak p}$, with $m$ replaced by
the effective parameter $m_{ij}(X)$.

For a left-invariant Finsler metric at a geodesic reference
direction, the Jacobi operator satisfies
\begin{equation}
R_X(V)
=
-N_X(N_X(V))
+
N_X([X,V])
-
[X,N_X(V)].
\label{eq:jacobi_N_short}
\end{equation}
We first apply \eqref{eq:jacobi_N_short} to $A_{ij}$. Using
\eqref{eq:root_brackets_short} and
\eqref{eq:effective_connection}, its three terms are respectively
\begin{equation}
\begin{aligned}
-N_X(N_X(A_{ij}))
&=
-\frac{(\lambda_i-\lambda_j)^2}{4}
\bigl(2+m_{ij}(X)\bigr)A_{ij},
\\
N_X([X,A_{ij}])
&=
-\frac{(\lambda_i-\lambda_j)^2}{2}A_{ij},
\\
-[X,N_X(A_{ij})]
&=
\frac{(\lambda_i-\lambda_j)^2}{2}
\bigl(2+m_{ij}(X)\bigr)A_{ij}.
\end{aligned}
\label{eq:curvature_A_three_terms}
\end{equation}
Adding the three coefficients leaves only
$m_{ij}(X)/4$, and therefore
\begin{equation}
R_X(A_{ij})
=
\frac{(\lambda_i-\lambda_j)^2}{4}
m_{ij}(X)A_{ij}.
\label{eq:curvature_A_effective}
\end{equation}

For $S_{ij}$, the three terms are
\begin{equation}
\begin{aligned}
-N_X(N_X(S_{ij}))
&=
-\frac{(\lambda_i-\lambda_j)^2}{4}
\bigl(2+m_{ij}(X)\bigr)S_{ij},
\\
N_X([X,S_{ij}])
&=
-\frac{(\lambda_i-\lambda_j)^2}{2}
\bigl(2+m_{ij}(X)\bigr)S_{ij},
\\
-[X,N_X(S_{ij})]
&=
\frac{(\lambda_i-\lambda_j)^2}{2}S_{ij}.
\end{aligned}
\label{eq:curvature_S_three_terms}
\end{equation}
Their sum is
\begin{equation}
R_X(S_{ij})
=
-
\frac{(\lambda_i-\lambda_j)^2}{4}
\bigl(4+3m_{ij}(X)\bigr)S_{ij}.
\label{eq:curvature_S_effective}
\end{equation}
Equations
\eqref{eq:curvature_A_effective} and
\eqref{eq:curvature_S_effective} prove
\eqref{eq:effective_cartan_curvature}.

Finally, $a_{ij}>0$, $s_{ij}>0$, and
$m_{ij}(X)>0$. Hence, whenever $\lambda_i\neq\lambda_j$,
\begin{equation}
g_X(R_XA_{ij},A_{ij})
=
\frac{(\lambda_i-\lambda_j)^2}{4}
m_{ij}(X)a_{ij}
>0,
\end{equation}
\begin{equation}
g_X(R_XS_{ij},S_{ij})
=
-
\frac{(\lambda_i-\lambda_j)^2}{4}
\bigl(4+3m_{ij}(X)\bigr)s_{ij}
<0.
\label{eq:root_signs_effective}
\end{equation}
This proves the proposition.
\end{proof}

\begin{proof}[Proof of Theorem~\ref{thm:sl_mixed}]
We first work at the identity. Choose
\begin{equation}
X
=
\operatorname{diag}
\left(
1,2,\ldots,n-1,-\frac{n(n-1)}2
\right).
\label{eq:mixed_flagpole}
\end{equation}
This matrix is trace free, full rank, and has pairwise distinct
diagonal entries. Proposition~\ref{prop:effective_cartan} applied
to the $(1,2)$ root block therefore gives
$g_X(R_XA_{12},A_{12})>0$ and
$g_X(R_XS_{12},S_{12})<0$.

To obtain zero curvature around the same flagpole, define
$U(t)=\cos t\,A_{12}+\sin t\,S_{12}$ for
$t\in[0,\pi/2]$. Both root directions are $g_X$-orthogonal to
$X$, so every $U(t)$ is transverse to $X$. Moreover,
$A_{12}$ and $S_{12}$ are mutually $g_X$-orthogonal and are
eigenvectors of $R_X$. Hence
\begin{equation}
\begin{aligned}
g_X(R_XU(t),U(t))
&=
\frac{(\lambda_1-\lambda_2)^2}{4}
m_{12}(X)a_{12}\cos^2 t
\\
&\quad
-
\frac{(\lambda_1-\lambda_2)^2}{4}
\bigl(4+3m_{12}(X)\bigr)s_{12}\sin^2 t.
\end{aligned}
\label{eq:zero_interpolation}
\end{equation}
The expression is positive at $t=0$ and negative at
$t=\pi/2$. By continuity, there exists
$t_0\in(0,\pi/2)$ for which
$g_X(R_XU(t_0),U(t_0))=0$.

Thus the three transverse directions
$A_{12}$, $U(t_0)$, and $S_{12}$ have respectively positive,
zero, and negative flag curvature around the same regular
full-rank flagpole $X$ by \eqref{eq:flag_sign}.

Finally, let $P\in\mathbb{SL}_p(n)$ be arbitrary. Left
translation by $P$ is an isometry because, for every
$B\in\mathbb{SL}_p(n)$ and tangent vector $V$ at $B$,
\begin{equation}
F_p(PB,PV)
=
\|(PB)^{-1}PV\|_{S_p}
=
\|B^{-1}V\|_{S_p}
=
F_p(B,V).
\label{eq:left_isometry_short}
\end{equation}
Therefore flag curvature is preserved by left translation.
The flagpole $PX\in T_P\mathbb{SL}_p(n)$ remains regular and
full rank, while the translated directions
$PA_{12}$, $PU(t_0)$, and $PS_{12}$ have respectively positive,
zero, and negative flag curvature. Since $P$ was arbitrary,
the three curvature regimes coexist at every point around a
common regular flagpole.
\end{proof}

\newpage

\subsection{Proof of Asymptotically Maximal Mixed Curvature and Coupling}
\label{proof:2.3}

The preceding Jacobi curvature calculation identifies positive and negative
curvature eigenspaces and hence their mixed-curvature capacity. To establish
intrinsic coupling according to Definition~\ref{def:coupling_capacity}, we
additionally need a nonzero interaction under the full Chern curvature operator
$\boldsymbol{\mathcal R}^{X}$.

\begin{lemma}[Nonvanishing Chern interaction on a root block]
\label{lem:root_chern_anchor}
Let $1<p<\infty$ and let
$X=\operatorname{diag}(\lambda_1,\ldots,\lambda_n)\in\mathfrak{sl}(n)$
be full rank. Fix $i<j$ and suppose
$\lambda_i=\lambda_j=c\neq0$.
Then
\begin{equation}
\boldsymbol{\mathcal R}^{X}(A_{ij},S_{ij})S_{ij}
=
\frac{a_{ij}}{2s_{ij}}A_{ij}.
\end{equation}
By Proposition~\ref{prop:effective_cartan},
At this equal-eigenvalue point,
When $\lambda_i=\lambda_j\neq 0$, the continuous extensions give
\(
a_{ij}
=
\|X\|_{S_p}^{\,2-p}|\lambda_i|^{p-2}
>0,
~
s_{ij}
=
\|X\|_{S_p}^{\,2-p}(p-1)|\lambda_i|^{p-2}
>0.
\) and therefore
\begin{equation}
g_X
\left(
\boldsymbol{\mathcal R}^{X}(A_{ij},S_{ij})S_{ij},
A_{ij}
\right)
=
\frac{a_{ij}^2}{2s_{ij}}
>0.
\end{equation}
\end{lemma}

\begin{proof}
Since $\lambda_i=\lambda_j$, we have
$[X,A_{ij}]=[X,S_{ij}]=0$.
Proposition~\ref{prop:effective_cartan} therefore gives
$N_X(A_{ij})=N_X(S_{ij})=0$.
For $H_{ij}=E_{ii}-E_{jj}$, the same connection identity gives
$N_X(H_{ij})=0$.

Let $\nabla^X$ denote the affine Chern connection associated with the
left invariant reference field determined by $X$.
The geodesic specialization of the homogeneous connection identity agrees
with the Chern Koszul formula
{\color{citegray}\cite[Proposition~2.3]{javaloyes2014chern}}, so
$(\nabla^X_UX)_I=N_X(U)$.
Consequently,
$\nabla^X_{A_{ij}}X=
\nabla^X_{S_{ij}}X=
\nabla^X_{H_{ij}}X=0$.
The anisotropic curvature formula of
{\color{citegray}\cite[Lemma~2.11]{javaloyes2019anisotropic}} then implies that, for the
directions used below, the anisotropic correction terms vanish. Hence
\begin{equation}
\boldsymbol{\mathcal R}^{X}(A_{ij},S_{ij})S_{ij}
=
R^{\nabla^X}(A_{ij},S_{ij})S_{ij}.
\end{equation}

At the equal eigenvalue point,
$g_X(H_{ij},H_{ij})=2s_{ij}$ and
$A_{ij},S_{ij},H_{ij}$ are mutually $g_X$ orthogonal.
For an arbitrary diagonal
$D=\operatorname{diag}(d_1,\ldots,d_n)$,
we also have
$g_X(H_{ij},D)=s_{ij}(d_i-d_j)$.
The relevant brackets are
$[A_{ij},S_{ij}]=H_{ij}$,
$[H_{ij},S_{ij}]=2A_{ij}$, and
$[H_{ij},A_{ij}]=2S_{ij}$.

Conjugation by diagonal orthogonal sign matrices fixes $X$ and shows that
$\nabla^X_{A_{ij}}S_{ij}$ and
$\nabla^X_{S_{ij}}S_{ij}$ are diagonal.
The Chern Koszul formula therefore gives, for every diagonal $D$,
\begin{equation}
\begin{aligned}
2g_X(\nabla^X_{A_{ij}}S_{ij},D)
&=
g_X([A_{ij},S_{ij}],D)
-
g_X([S_{ij},D],A_{ij})
+
g_X([D,A_{ij}],S_{ij})
\\
&=
(2s_{ij}+a_{ij})(d_i-d_j).
\end{aligned}
\end{equation}
On the other hand,
\begin{equation}
2g_X
\left(
\left(1+\frac{a_{ij}}{2s_{ij}}\right)H_{ij},
D
\right)
=
(2s_{ij}+a_{ij})(d_i-d_j).
\end{equation}
Both vectors are diagonal, and $g_X$ is nondegenerate there. Therefore
$\nabla^X_{A_{ij}}S_{ij}
=
(1+a_{ij}/(2s_{ij}))H_{ij}$.
Applying the same Koszul identity to
$\nabla^X_{S_{ij}}S_{ij}$ and
$\nabla^X_{H_{ij}}S_{ij}$ gives
$\nabla^X_{S_{ij}}S_{ij}=0$ and
$\nabla^X_{H_{ij}}S_{ij}=A_{ij}$.
Since the Chern connection is torsion free,
\begin{equation}
\nabla^X_{S_{ij}}H_{ij}
=
\nabla^X_{H_{ij}}S_{ij}
-
[H_{ij},S_{ij}]
=
-A_{ij}.
\end{equation}

Using
$R^{\nabla^X}(U,V)W
=
\nabla^X_U\nabla^X_VW
-
\nabla^X_V\nabla^X_UW
-
\nabla^X_{[U,V]}W$,
we obtain
\begin{equation}
\begin{aligned}
\boldsymbol{\mathcal R}^{X}(A_{ij},S_{ij})S_{ij}
&=
R^{\nabla^X}(A_{ij},S_{ij})S_{ij}
\\
&=
-\nabla^X_{S_{ij}}
\left(
\left(1+\frac{a_{ij}}{2s_{ij}}\right)H_{ij}
\right)
-
\nabla^X_{H_{ij}}S_{ij}
\\
&=
\left(1+\frac{a_{ij}}{2s_{ij}}\right)A_{ij}
-
A_{ij}
\\
&=
\frac{a_{ij}}{2s_{ij}}A_{ij}.
\end{aligned}
\end{equation}
Taking the $g_X$ inner product with $A_{ij}$ proves the final claim.
\end{proof}

\begin{proof}[Proof of Corollary~\ref{thm:asymptotic_coupling}]
For $n=2$, the lower bound
$\binom{n-1}{2}=0$ is immediate, so we assume $n\geq3$.

For fixed $p\in(1,\infty)$,
$\mathcal E_p(X)=\frac12[\operatorname{tr}((X^\top X)^{p/2})]^{2/p}$ is real
analytic on the full rank matrix locus. Indeed, $X^\top X$ is positive
definite there, and standard analytic matrix function calculus
\citep{higham2008functions}  applies. Hence the fundamental tensor,
the Chern connection, and the Chern curvature operator are real analytic in
the reference direction wherever the fundamental tensor is nondegenerate.

First, we consider the connected diagonal chamber
\begin{equation}
\Omega
=
\left\{
\operatorname{diag}(\lambda_1,\ldots,\lambda_n):
\lambda_1,\ldots,\lambda_{n-1}>0,\ 
\lambda_n=-\sum_{r=1}^{n-1}\lambda_r
\right\}.
\end{equation}
For each $1\leq i<j\leq n-1$, let
$\kappa_{ij}(X)
=
g_X(
\boldsymbol{\mathcal R}^{X}(A_{ij},S_{ij})S_{ij},
A_{ij})$.
This is a real analytic scalar function on $\Omega$.
For every fixed pair $(i,j)$, $\Omega$ contains a point with
$\lambda_i=\lambda_j>0$, and
Lemma~\ref{lem:root_chern_anchor} gives
$\kappa_{ij}(X)>0$ there.
Thus $\kappa_{ij}$ is not identically zero.
Its nonzero set is therefore open and dense in $\Omega$.

There are only finitely many pairs $1\leq i<j\leq n-1$.
Intersecting all of these open dense sets with the open dense subset on which
the diagonal entries are pairwise distinct gives a regular full rank
$X\in\Omega$ satisfying
\begin{equation}
\kappa_{ij}(X)\neq0,
\qquad
1\leq i<j\leq n-1.
\end{equation}

For any diagonal sign matrix
$D=\operatorname{diag}(\varepsilon_1,\ldots,\varepsilon_n)$ with
$\varepsilon_r\in\{\pm1\}$, conjugation
$\Phi_D(B)=DBD^{-1}$ fixes every diagonal $X$ and is an isometry because
\begin{equation}
F_p(\Phi_D(B),d\Phi_D(V))
=
\|DB^{-1}VD^{-1}\|_{S_p}
=
\|B^{-1}V\|_{S_p}.
\end{equation}
Moreover,
$d\Phi_D(A_{ij})=(\varepsilon_i\varepsilon_j)A_{ij}$ and
$d\Phi_D(S_{ij})=(\varepsilon_i\varepsilon_j)S_{ij}$.

Naturality of Chern curvature under isometries now gives the required block
separation. For two unordered pairs $(i,j)$ and $(k,l)$,
$\boldsymbol{\mathcal R}^{X}(A_{kl},S_{ij})S_{ij}$ transforms under every
$\Phi_D$ with the same sign
$\varepsilon_k\varepsilon_l$ as the $(k,l)$ root block, because the two
copies of $S_{ij}$ contribute the square
$(\varepsilon_i\varepsilon_j)^2=1$.
Distinct root blocks have distinct sign patterns under all such
conjugations. Therefore
\begin{equation}
\boldsymbol{\mathcal R}^{X}(A_{kl},S_{ij})S_{ij}
\in
\operatorname{span}\{A_{kl},S_{kl}\}.
\end{equation}
The same argument, now with $A_{ij}$ and $S_{ij}$ contributing a cancelling
square, gives
\begin{equation}
\boldsymbol{\mathcal R}^{X}(A_{ij},S_{kl})S_{ij}
\in
\operatorname{span}\{A_{kl},S_{kl}\}.
\end{equation}

For the generic $X$ fixed above, consider
\begin{equation}
W_+
=
\operatorname{span}
\{A_{ij}:1\leq i<j\leq n-1\},
\qquad
W_-
=
\operatorname{span}
\{S_{ij}:1\leq i<j\leq n-1\}.
\end{equation}
Proposition~\ref{prop:effective_cartan} gives
$W_+\subseteq E_+(X)$ and
$W_-\subseteq E_-(X)$, with
\begin{equation}
\dim W_+
=
\dim W_-
=
\binom{n-1}{2}.
\end{equation}
It follows immediately that
\begin{equation}
\mathcal C_{\mathrm{mix}}(X)
=
\min\{\dim E_+(X),\dim E_-(X)\}
\geq
\binom{n-1}{2}.
\end{equation}

We next show that the same lower bound remains after excluding uncoupled
modes. First,
$\mathcal N_+(X)\cap W_+=\{0\}$.
Suppose otherwise and write a nonzero vector in the intersection as
$U=\sum_{k<l}u_{kl}A_{kl}$.
Choose $(i,j)$ with $u_{ij}\neq0$.
Since $S_{ij}\in E_-(X)$ and
$U\in\mathcal N_+(X)$, the definition of
$\mathcal N_+(X)$ gives
$\boldsymbol{\mathcal R}^{X}(U,S_{ij})=0$
as an endomorphism.
Applying it to $S_{ij}$ yields
\begin{equation}
0
=
\sum_{k<l}
u_{kl}
\boldsymbol{\mathcal R}^{X}(A_{kl},S_{ij})S_{ij}.
\end{equation}
By the root block separation proved above, the term indexed by $(k,l)$ lies
in $\operatorname{span}\{A_{kl},S_{kl}\}$.
These root blocks are linearly independent, so the component in
$\operatorname{span}\{A_{ij},S_{ij}\}$ must vanish:
\begin{equation}
u_{ij}
\boldsymbol{\mathcal R}^{X}(A_{ij},S_{ij})S_{ij}
=
0.
\end{equation}
Taking the $g_X$ inner product with $A_{ij}$ gives
$u_{ij}\kappa_{ij}(X)=0$, contradicting both
$u_{ij}\neq0$ and $\kappa_{ij}(X)\neq0$.
Hence
$\mathcal N_+(X)\cap W_+=\{0\}$.

Both $\mathcal N_+(X)$ and $W_+$ are subspaces of $E_+(X)$.
Their trivial intersection therefore implies
\begin{equation}
\dim E_+(X)-\dim\mathcal N_+(X)
\geq
\dim W_+
=
\binom{n-1}{2}.
\end{equation}

The negative side is analogous, but we spell it out.
Suppose
$0\neq V=\sum_{k<l}v_{kl}S_{kl}
\in\mathcal N_-(X)\cap W_-$
and choose $(i,j)$ with $v_{ij}\neq0$.
Because $A_{ij}\in E_+(X)$ and
$V\in\mathcal N_-(X)$,
we have
$\boldsymbol{\mathcal R}^{X}(A_{ij},V)=0$.
Applying this endomorphism to $S_{ij}$ gives
\begin{equation}
0
=
\sum_{k<l}
v_{kl}
\boldsymbol{\mathcal R}^{X}(A_{ij},S_{kl})S_{ij}.
\end{equation}
The second root block separation property places each summand in its
$(k,l)$ root block. Projecting onto
$\operatorname{span}\{A_{ij},S_{ij}\}$ gives
\begin{equation}
v_{ij}
\boldsymbol{\mathcal R}^{X}(A_{ij},S_{ij})S_{ij}
=
0,
\end{equation}
which again contradicts
$v_{ij}\kappa_{ij}(X)\neq0$.
Thus
$\mathcal N_-(X)\cap W_-=\{0\}$, and the same dimension argument gives
\begin{equation}
\dim E_-(X)-\dim\mathcal N_-(X)
\geq
\binom{n-1}{2}.
\end{equation}
Therefore
\begin{equation}
\mathcal C_{\mathrm{cpl}}(X)
\geq
\binom{n-1}{2}.
\end{equation}

Let $A\in\mathbb{SL}_p(n)$ be arbitrary and consider the left translated
flagpole $AX\in T_A\mathbb{SL}_p(n)$.
Left translation by $A$ is an isometry of the Schatten $p$ structure.
Hence its differential preserves the fundamental tensor and the Chern
curvature:
\begin{equation}
\begin{aligned}
g_{AX}(AU,AV)
&=
g_X(U,V),
\\
R_{AX}(AU)
&=
A\,R_XU,
\\
\boldsymbol{\mathcal R}^{AX}(AU,AV)(AW)
&=
A\,
\boldsymbol{\mathcal R}^{X}(U,V)W.
\end{aligned}
\end{equation}
The second identity implies
$E_\pm(AX)=AE_\pm(X)$, while the third gives
$\mathcal N_\pm(AX)=A\mathcal N_\pm(X)$.
Since left multiplication is invertible, all corresponding spaces have the
same dimensions. Hence
\begin{equation}
\mathcal C_{\mathrm{mix}}(AX)
=
\mathcal C_{\mathrm{mix}}(X),
\qquad
\mathcal C_{\mathrm{cpl}}(AX)
=
\mathcal C_{\mathrm{cpl}}(X).
\end{equation}
Since $AX$ is a regular full rank flagpole at the arbitrary point $A$,
the pointwise definitions yield
\begin{equation}
\mathcal C_{\mathrm{mix}}(A)
\geq
\binom{n-1}{2},
\qquad
\mathcal C_{\mathrm{cpl}}(A)
\geq
\binom{n-1}{2}.
\end{equation}
Taking the minimum over $A\in\mathrm{SL}(n)$ therefore gives
\begin{equation}
\mathcal C_{\mathbb{SL}}^{\mathrm{mix}}(n,p)
\geq
\binom{n-1}{2},
\qquad
\mathcal C_{\mathbb{SL}}^{\mathrm{cpl}}(n,p)
\geq
\binom{n-1}{2}.
\end{equation}

Then at any $A\in\mathbb{SL}_p(n)$ and any regular flagpole
$Y\in T_A\mathbb{SL}_p(n)$, the tangent space has dimension $n^2-1$, so the
$g_Y$ transverse space has dimension $n^2-2$. Because $R_Y$ is $g_Y$ self
adjoint, its positive, zero, and negative eigenspaces form a direct orthogonal
decomposition of this transverse space. Therefore,
\begin{equation}
\begin{aligned}
\mathcal C_{\mathrm{cpl}}(Y)
&\leq
\mathcal C_{\mathrm{mix}}(Y)
=
\min\{\dim E_+(Y),\dim E_-(Y)\}
\\
&\leq
\left\lfloor
\frac{\dim E_+(Y)+\dim E_-(Y)}{2}
\right\rfloor
\leq
\left\lfloor
\frac{n^2-2}{2}
\right\rfloor.
\end{aligned}
\end{equation}
Define the maximal balanced curvature capacity permitted by the transverse
space as
\(
\mathcal C_{\max}(n):=\lfloor(n^2-2)/2\rfloor.
\)
Since the bound holds for every regular flagpole at every point, taking the
pointwise maximum and then the minimum over the manifold gives
\begin{equation}
\mathcal C_{\mathbb{SL}}^{\mathrm{cpl}}(n,p)
\leq
\mathcal C_{\mathbb{SL}}^{\mathrm{mix}}(n,p)
\leq
\mathcal C_{\max}(n).
\end{equation}

Combining this with the lower bound
$\mathcal C_{\mathbb{SL}}^{\mathrm{cpl}}(n,p)\geq\binom{n-1}{2}$ yields
\begin{equation}
\frac{\binom{n-1}{2}}{\mathcal C_{\max}(n)}
\leq
\frac{\mathcal C_{\mathbb{SL}}^{\mathrm{cpl}}(n,p)}
{\mathcal C_{\max}(n)}
\leq
\frac{\mathcal C_{\mathbb{SL}}^{\mathrm{mix}}(n,p)}
{\mathcal C_{\max}(n)}
\leq
1.
\end{equation}
The leftmost term converges to $1$ as $n\to\infty$. Hence, by squeezing,
\begin{equation}
\frac{\mathcal C_{\mathbb{SL}}^{\mathrm{cpl}}(n,p)}
{\mathcal C_{\max}(n)}
\longrightarrow1,
\qquad
\frac{\mathcal C_{\mathbb{SL}}^{\mathrm{mix}}(n,p)}
{\mathcal C_{\max}(n)}
\longrightarrow1.
\end{equation}
\end{proof}

\subsection{Proof of Infinite Order Depth}
\label{proof:2.4}

\begin{proof}[Proof of Lemma~\ref{thm:sl_order_depth}]
For every $n\geq2$, the Lie algebra $\mathfrak{sl}(n)$ contains the
upper left $2\times2$ copy of $\mathfrak{sl}(2)$. Let
$H=E_{11}-E_{22}$ and $E=E_{12}$.
Then $[H,E]=2E$, and consequently, for every integer $k\geq1$,
\begin{equation}
\operatorname{ad}_H^k(E)
=
\underbrace{
[H,[H,\ldots,[H}_{k\text{ times}},E]\ldots]]
=
2^kE
\neq0.
\end{equation}
Choosing $X_0=E$ and
$X_1=\cdots=X_k=H$ in the definition of order depth therefore gives a
nonzero nested Lie bracket at every finite depth $k$.
Since
$T_I\mathbb{SL}_p(n)=\mathfrak{sl}(n)$
for every $p\in(1,\infty)$, this construction is independent of $p$.
Hence
\begin{equation}
D_{\mathrm{ord}}(\mathbb{SL}_p(n))
=
\infty.
\end{equation}
\end{proof}

\subsection{Counting Product Geometry Candidates}
\label{app:product_count}

We detail how the number $11{,}555{,}651{,}398$ in the introduction is
calculated. Consider
\begin{equation}
\mathcal P
=
\mathbb E^{d_0}
\times
\prod_{i=1}^{m_-}\mathbb H^{d_i}
\times
\prod_{j=1}^{m_+}\mathbb S^{d_j},
\end{equation}
with fixed total dimension
\(d_0+\sum_i d_i+\sum_j d_j=64\).
We fix the curvature magnitudes and count only distinct choices of factor
dimensions. Factors of the same type are treated as unordered, since
permuting, for example, $\mathbb H^{d_1}$ and $\mathbb H^{d_2}$ does not
produce a different product geometry.
Let $p(n)$ denote the integer partition number. The partition numbers are
computed recursively from
\begin{equation}
p(n)
=
p(n-1)+p(n-2)-p(n-5)-p(n-7)
+p(n-12)+p(n-15)-\cdots,
\end{equation}
where the offsets $1,2,5,7,12,15,\ldots$ are the generalized pentagonal
numbers, with $p(0)=1$ and $p(n)=0$ for $n<0$.
Hence, if the total dimension assigned to the hyperbolic factors is $h$, the
possible decompositions $h=d_1+\cdots+d_{m_-}$ are counted by $p(h)$. For
example, for $h=5$, the partitions $5$, $4+1$, $3+2$, $3+1+1$, $2+2+1$,
$2+1+1+1$, and $1+1+1+1+1$ correspond to seven distinct choices of
hyperbolic factors. The spherical factors are counted independently in the
same way.
For a fixed Euclidean dimension $d_0$, let $h=\sum_i d_i$ be the total
hyperbolic dimension. The remaining spherical dimension is then
$s=64-d_0-h$. There are therefore $p(h)$ possible hyperbolic decompositions
and $p(s)$ possible spherical decompositions. Since the two choices are
independent, a fixed pair $(d_0,h)$ contributes
$p(h)p(64-d_0-h)$ candidates. Summing over all admissible dimension
allocations gives
\begin{equation}
N_{64}
=
\sum_{d_0=0}^{64}
\sum_{h=0}^{64-d_0}
p(h)p(64-d_0-h)
=
11{,}555{,}651{,}398.
\end{equation}

\end{document}